%% file: main.tex
\documentclass[11pt,reqno]{amsart}
\usepackage{geometry}
\usepackage{graphicx}
\usepackage{amssymb}
\usepackage{caption}
\usepackage{subcaption}
\usepackage{enumitem}
\usepackage{mathtools}
\usepackage[all]{xy}
\usepackage{xcolor}
\usepackage{soul}
\usepackage[hidelinks,colorlinks=true,linkcolor=blue,urlcolor=blue,citecolor=blue]{hyperref}

\usepackage{microtype}

\usepackage{bbold}

\usepackage{MnSymbol}
\usepackage[linesnumbered,ruled,vlined]{algorithm2e}

\SetCommentSty{mycommfont}
\SetKwInput{KwInput}{Input}               
\SetKwInput{KwOutput}{Output}

\usepackage{graphicx}

\usepackage{tikz}

\usepackage[numbers,sort&compress]{natbib}
\bibpunct{[}{]}{;}{n}{,}{,}

\graphicspath{{figures/}}

\numberwithin{equation}{section}

\newtheorem{theorem}{Theorem}[section]

\newtheorem*{remark*}{Remark}

\DeclarePairedDelimiterX{\inner}[2]{\langle}{\rangle}{#1, #2}

\allowdisplaybreaks

\title[An Operator Learning Framework for Spatiotemporal Super-Resolution]{An Operator Learning Framework for \\ Spatiotemporal Super-Resolution of Scientific Simulations}
\author{Valentin Duruisseaux and Amit Chakraborty}

\begin{document}

	\maketitle
	
	\begin{abstract}  
		
		In numerous contexts, high-resolution solutions to partial differential equations are required to capture faithfully essential dynamics which occur at small spatiotemporal scales, but these solutions can be very difficult and slow to obtain using traditional methods due to limited computational resources. A recent direction to circumvent these computational limitations is to use machine learning techniques for super-resolution, to reconstruct high-resolution numerical solutions from low-resolution simulations which can be obtained more efficiently. The proposed approach, the Super Resolution Operator Network (SROpNet), frames super-resolution as an operator learning problem and draws inspiration from existing architectures to learn continuous representations of solutions to parametric differential equations from low-resolution approximations, which can then be evaluated at any desired location. In addition, no restrictions are imposed on the locations of (the fixed number of) spatiotemporal sensors at which the low-resolution approximations are provided, thereby enabling the consideration of a broader spectrum of problems arising in practice, for which many existing super-resolution approaches are not well-suited.
	\end{abstract}

	\hfill  
	
	\setcounter{tocdepth}{3}
	\tableofcontents

	\newpage

	\section{Preliminaries}

	\subsection{Operator Learning}
	
	\subsubsection{Motivation}

	\hfill \\

	Many phenomena and dynamical systems of interest in science and engineering can be modeled as nonlinear partial differential equations (PDEs). Since closed-form solutions to these PDEs are typically not known, simulating the associated dynamics requires the use of numerical integration methods which quickly become very expensive computationally as the desired simulation resolution increases. Unfortunately, in numerous contexts, high-resolution solutions are required to capture faithfully essential dynamics which occur at small spatiotemporal scales, making most traditional solvers prohibitively slow and expensive. The epitomes of dynamical systems and PDEs where a high resolution is vital are turbulent fluid flows and the Navier--Stokes equation, where small eddies can be responsible for significant energy and heat transfers.  \\
	
	As an attempt to circumvent this issue, Physics-Informed Machine Learning~\cite{Karniadakis2021} has emerged as a new research direction consisting in the use of machine learning techniques to design or improve computational methods for scientific applications. The idea is to encode the physics underlying the systems of interest in the design of the neural networks or in the learning process. Available physics prior knowledge can be used to construct physics-constrained neural networks with improved design and efficiency and a better generalization capacity, which take advantage of the function approximation power of neural networks to deal with incomplete knowledge. 
	
	In the context of solving nonlinear PDEs, a Physics-Informed Neural Network (PINN)~\cite{Raissi2017Part1,Raissi2017Part2,Raissi2019} is a neural network representation of the solution of a PDE, whose parameters are learnt by minimizing both the distance to the reference PDE solution in the dataset and deviations from the physics laws, conservation laws, symmetries and structural properties, and from the governing differential equations. The use of PINNs is theoretically justified by well-known universal function approximation theorems for neural networks~\cite{Cybenko1989,Hornik1989,Pinkus1999,Kidger2020,Yarotsky2020}, and further error estimates have been obtained for PINNs~\cite{Mishra2022}. Many modified and extended versions of PINNs have also been proposed and used to solve PDEs in numerous contexts~\cite{Lu2021deepxde,Pang2019,Zhang2020,Meng2020,Jagtap2020_1,Jagtap2020_2,Yang2021,Yu2022,Shukla2021,Kashinath2021,Kharazmi2021,Cai2021,Cai2022}.

	There are however several limitations to function learning approaches such as PINNs for numerically solving differential equations. First, PINNs only learn a solution representation for a single instance of the parametric family of PDEs of interest, that is for a single specific choice of boundary and initial conditions, forcing terms, source of loading terms, parameter values, etc... As a result, the learnt model does not generalize well to other instances without additional computationally expensive retraining, and PINNs are thus not particularly effective in practice when we need to predict rapidly the dynamics of a system under different operating conditions, although transfer learning approaches can alleviate this limitation to some extent~\cite{Goswami2022_3,Goswami2022_4}. Empirically, it has also been demonstrated that PINNs often struggle to solve challenging PDEs when the solution exhibits high-frequency behaviour or multi-scale structure~\cite{Fuks2020,Wang2021,Raissi_2020,Wang2022_3}. Furthermore, PINNs can also suffer from the curse-of-dimensionality during training since the target is typically the discrete solution of the PDE, possibly evaluated on a very large spatiotemporal mesh.  \\

This is where operator learning comes in. It can be used to obtain a continuous representation of the solution operator to a family of parametric PDEs, that can then be evaluated on any mesh under a wide range of operating conditions without requiring additional expensive training at inference time. For the training data, the target solutions also do not need to be specified on the same grid for every sample, providing greater flexibility. In addition, the trained parametrized representation sometimes even generalizes well to operating conditions outside of the distribution it experienced during training.  \\

	\newpage

	\subsubsection{Operator Learning} \label{subsubsec: Neural Operators}
	
	\hfill   \\

	We are interested in learning a representation for a nonlinear mapping which takes a function as input and returns another function. More precisely, let $\mathcal{U}$ be a space of real-valued functions with domain $D_u \subset \mathbb{R}^{d_u}$, and let $\mathcal{V}$ be a space of real-valued functions with domain $D_v \subset \mathbb{R}^{d_v}$. We wish to learn a representation $\mathcal{G}_\theta$ (with tunable parameters $\theta$) of a mapping $\mathcal{G}$ which takes an input function $u \in \mathcal{U}$ and returns an output function $v = \mathcal{G}(u)(\cdot ) \in \mathcal{V}$, such that
	\begin{equation}
		\mathcal{G}_\theta(u)(y)  \approx  \mathcal{G}(u)(y) \quad \ \forall u \in \mathcal{U}, \  \forall y\in D_v  .
	\end{equation}

	Unlike function learning where the learnt model only takes an output location $y$ as input, the learnt model also takes the input function $u$ as input in operator learning. In practice however, the input function is often represented in a discrete way, which is typically done through function evaluations $\left[ u(x_1), \ldots, u(x_s) \right]$ at a finite set of sensor locations $\{ x_1, \ldots, x_s \} \subset D_u$.     \\

	A very popular and general deep learning architecture for operator learning is \textbf{DeepONet}~\citep{DeepONet}, which consists of a \textbf{Branch Network} encoding the input function values $\left[ u(x_1), \ldots, u(x_s) \right]$ at the sensor locations $\{ x_1, \ldots, x_s \}\in D_u $, and a \textbf{Trunk Network} encoding the location $y \in D_v$ where the output function has to be evaluated. The output of the operator representation is then given by 
	\begin{equation}
		\mathcal{G}_\theta (u)(y) = \sum_{k=1}^{K}{b_k(u) t_k(y)},
	\end{equation}
	where $\{b_1,\ldots, b_K\}$ and $\{t_1,\ldots, t_K\}$ are the outputs of the branch  and trunk networks in some $K$-dimensional latent space, respectively. Here is a depiction of the DeepONet architecture:

	\hfill

	\input{diagrams/General2Branch}

	\hfill

	The trunk network is usually chosen to be a Multilayer Perceptron (MLP) since the dimension $d_v$ of $y$ is typically small. The choice of branch network depends on the dimensionality and structure of the discretization of the input function $u$. Common choices include Multilayer Perceptrons (MLPs), Convolutional Neural Networks (CNNs), Recurrent Neural Networks (RNNs), Graph Neural Networks (GNNs), and combinations and modified versions of these architectures.   \\
	
	The use of DeepONets and most operator learning approaches is theoretically justified by universal approximation theorems for nonlinear operators and error estimates results~\cite{ChenChen1993,ChenChen1995_2,ChenChen1995,DeepONet,Mhaskar1997,Lanthaler2022,deHoop2022,Yu2021}. In particular, DeepONets can approximate continuous operators uniformly over compact
	subsets:
	
	\begin{theorem}
		Suppose $D_u \subset \mathbb{R}^{d_u}$ and $D_v \subset \mathbb{R}^{d_v}$ are compact sets. Let $\mathcal{G}$ be a nonlinear continuous operator mapping a subset of $C(D_u)$ into $C(D_v)$. Then, given $\varepsilon>0$, there exists a DeepONet $\mathcal{G}_\theta$ such that $	|\mathcal{G}(u)(y) - \mathcal{G}_\theta(u)(y) | < \varepsilon $ for all $u\in \mathcal{U}, \ y \in D_v$.
	\end{theorem}

\noindent	This result was generalized in~\cite{Lanthaler2022} to remove the continuity assumption on $\mathcal{G}$ and the compactness assumption, which makes it more easily applicable in the context of approximating nonlinear solution operators to differential equations:
	
	\begin{theorem}
	Let $\mu$ be a probability measure on $C(D_u)$, and $\mathcal{G}$ a Borel measurable mapping on $C(D_u)$ with $\mathcal{G} \in  L^2(\mu)$. Then for every $\varepsilon>0$, there is a DeepONet $\mathcal{G}_\theta$ such that $	\| \mathcal{G} - \mathcal{G}_\theta \|_{L^2(\mu)} < \varepsilon$.
	\end{theorem}

\noindent Some error estimates are obtained in~\cite{Lanthaler2022} by decomposing DeepONets as $\mathcal{G}_\theta = \mathcal{R} \circ \mathcal{A} \circ \mathcal{E}$ where $\mathcal{E}$ encodes the input function space as discretizations $\left[ u(x_1), \ldots, u(x_s) \right]$ at the sensor locations, the approximator $\mathcal{A}$ takes the encoded input function and returns the coefficients $b_k$ of the DeepONet which are then combined with the outputs of the trunk network by the reconstructor $\mathcal{R}$. After defining the decoder $\mathcal{D}$ and projector $\mathcal{P}$ such that $\mathcal{E} \circ \mathcal{D} = \text{Id}, \ \mathcal{D} \circ \mathcal{E} \approx \text{Id}$ and 	$\mathcal{P} \circ \mathcal{R} = \text{Id},  \ \mathcal{R} \circ \mathcal{P} \approx \text{Id}$, the error when approximating an $\alpha$-H\"older continuous operator $\mathcal{G}$ can be decomposed (under a certain set of assumptions) in terms of encoding, approximation, and reconstruction errors
\begin{equation*}
	E_e =  \| \mathcal{D}  \circ  \mathcal{E} - \text{Id}  \|_{L^2(\mu)} , \quad \ \   E_a =  \| \mathcal{A} - \mathcal{P}  \circ  \mathcal{G}  \circ  \mathcal{D}  \|_{L^2(\mathcal{E}_{\#} \mu)} ,  \quad \ \     E_r = \| \mathcal{R}  \circ  \mathcal{P} - \text{Id}  \|_{L^2(\mathcal{G}_{\#} \mu)} ,
\end{equation*}
as 
\begin{equation*}
	 \sqrt{\int_{C(D_u)}{\int_{D_v} {|\mathcal{G}(u)(y) - \mathcal{G}_\theta(u)(y) |^2 dy  } d\mu(u)}}  \   \leq  \   \text{Lip}_{\alpha}(\mathcal{G} )  \text{Lip}_{1}(\mathcal{R}  \circ   \mathcal{P})   E_e^\alpha  +  \   \text{Lip}_{1}(\mathcal{R} )    E_a     \ + \    E_r,
\end{equation*}
where $\text{Lip}_{\beta}$ denotes the $\beta$-H\"older continuity constant, and $\mathcal{E}_{\#} \mu$, $\mathcal{G}_{\#} \mu$ denote push-forwards of the measure $\mu$. Results about the generalization error of DeepONets are also derived in~\cite{Lanthaler2022}.

	\hfill

	Many modified and more sophisticated versions of DeepONet have been introduced to adapt to various scenarios~\cite{Lu2022} and have been used in numerous contexts~\citep{Leoni2023,Lu2022,Oommen2022,Lanthaler2022_2,Lin2023,DeepONet,Lin2021,Leoni2021,Goswami2022_2,Yin2021}. DeepONets can easily be generalized to take multiple input functions in the branch network and output multiple functions. Feature expansions in the subnetworks can be used to encode additional prior knowledge about the solution operator. See~\cite{Leoni2023} for an example where a harmonic feature expansion is used on the prediction location $y$ in the trunk network to deal with the highly oscillating nature of the data. Instead of using a trunk network to learn a basis for the output function, this basis can also be specified or computed prior to learning. For instance, inspired by~\cite{Bhattacharya2020}, POD-DeepONet~\cite{Lu2022} uses a Proper Orthogonal Decomposition (POD) basis computed on the training data, and the trunk network and its outputs are replaced by the POD projection of the prediction location. DeepONet can also be trained in the latent space of an autoencoder~\cite{Oommen2022}.   \\

	In our framework, we use a new modification to the DeepONet architecture, inspired by BelNet~\cite{BelNet}. We add a third subnetwork, the \textbf{Sensor Network}, which takes the sensor locations $\{ x_1, \ldots, x_s \}\subset D_u $ as input, to relax constraints on the sensor locations. Instead of requiring fixed sensor locations across the dataset, every data sample can have different sensor locations as long as the number $s$ of sensors remains the same. The resulting 3-Subnetwork architecture is depicted below:   \\

\input{diagrams/General3Branch}
	
\vspace{3mm}

The extra sensor network makes the resulting modified DeepONet mesh-free: there are no restrictions on the discretization, domains, and grids, for the input and output functions (aside from the fixed number $s$ of sensors). This allows in particular to consider situations where information about the solution is collected from non-stationary sensors (e.g. sensors moving in a fluid, weather stations for meteorological forecasting, or guided probing devices), or situations where there is randomness and non-uniformity in the times at which measurements are collected (either voluntarily based on some prior knowledge of the system's dynamics, or involuntarily due to precision and accuracy limitations of sensing equipment), or situations where adaptive meshes could be particularly useful (e.g. adding more points in regions with faster dynamics while reducing the number of points and computations where dynamics vary slowly, which is frequently done in traditional solvers for PDEs simulation via domain decomposition and adaptive mesh refinement techniques).  \\

	In addition to the freedom of choosing the architecture for each subnetwork, there are many possible variations starting from $u$, $x$ and $y$ as inputs by combining judiciously pairs of subnetworks earlier in the architecture. Possibly helpful examples include stacking the discretized input function $u$ and the sensor locations $x$ before passing it through a common branch network, or computing some appropriate weighted distance tensor between the sensor locations $x$ and prediction location $y$ before passing it through a common trunk network (if $d_u=d_v$):   \\
	
	 \vspace{0.4mm}

	\input{diagrams/General3BranchOtherExamples}
	
	\hfill  
	
	\vspace{-1.8mm}
	
	Note that several operator learning frameworks different or more specific than DeepONet have been developed. In particular, different versions of Fourier Neural Operators (FNOs)~\cite{Kovachki2021,FNO}, based on Fourier transforms and on parametrizing a convolution directly in Fourier space, have been proposed and used successfully in numerous contexts~\cite{Wen2022,Li2023_2,Bonev2023,Wen2023,Li2022,Guibas2021,Kurth2022,PINO,Kossaifi2023,Gopakumar2023,Zhao2023,White2023}, while enjoying a universal operator approximation property~\cite{Kovachki2021_2}.  \\

	In the context of solving families of parametric PDEs, operator learning is typically used to learn the solution operator mapping the initial and boundary conditions to the corresponding solution. Analogously to PINNs in the function learning setting, we can obtain physics-informed operator learning frameworks by minimizing deviations from physics laws and governing differential equations in addition to minimizing the distance to reference solutions. By constraining the representation to satisfy physics laws, one can typically train the model with less data and obtain a surrogate operator with a better generalization capacity. The resulting loss is of the form $\mathcal{L} =  \mathcal{L}_{data}  +  \mathcal{L}_{physics} $ where $\mathcal{L}_{physics}$ is a weighted sum of a PDE residual loss (in strong or variational form), an initial condition loss, a boundary condition loss, and any additional loss term penalizing deviations from appropriate physics or conservation laws. Minimizing this composite loss is more computationally expensive and challenging, although strategies exist to update adaptively the coefficients in the weighted sum of loss terms during training (such as Learning Rate Annealing~\cite{Wang2021}, GradNorm~\cite{Chen2018}, SoftAdapt~\cite{Heydari2019}, and Relative Loss Balancing with Random Lookback (ReLoBRaLo)~\cite{Bischof2021}).

	\newpage

	\subsection{Spatiotemporal Super-Resolution of Scientific Simulations}
	
	\hfill \\
	
	In numerous contexts, high-resolution (HR) solutions to PDEs are required to capture faithfully essential dynamics which occur at small spatiotemporal scales, but these HR solutions can be very difficult to obtain using traditional methods due to limited computational resources. A recent research direction to circumvent these limitations is machine-learning-based super-resolution, where deep learning techniques are used to reconstruct HR solutions from low-resolution (LR) simulations which can be obtained more efficiently using traditional methods. Super-resolution of time-dependent solutions to PDEs is the special case of video super-resolution where the spatiotemporal data has additional structure coming from the laws of physics encompassed by the governing PDEs. \\   
	
	Many existing super-resolution approaches for spatiotemporal scientific data coming from dynamical systems constrained by physics laws and differential equations are based on or inspired by traditional computer vision approaches for images and videos. Image and video super-resolution are well-studied problems in computer vision (see~\cite{Nasrollahi2014,Wang2021_2,Liu2022,Anwar2020,Li2020} for recent surveys), and a plethora of significantly different machine learning approaches have been proposed. 
	
	For images, existing approaches include (Fast) Super Resolution Convolutional Neural Network ((Fast) SRCNN)~\cite{Dong2014,Dong2014_2,Dong2016}, Very Deep Super Resolution (VDSR)~\cite{Kim2016}, Dense Skip Connections~\cite{Tong2017}, Deep Recursive Residual Networks (DRRN)~\cite{Tai2017}, Enhanced Deep Residual Networks (EDSR)~\cite{Lim2017}, deep Laplacian Pyramid Networks (LapSRN)~\cite{Lai2017}, Cascading Residual Networks (CARN)~\cite{Ahn2018}, Residual Dense Networks (RDN)~\cite{Zhang2018_9}, (Enhanced) Super Resolution Generative Adversarial Networks ((E)SRGAN)~\cite{Ledig2017,Wang2018}, deep Residual Channel Attention Networks (RCAN)~\cite{RCAN}, Second-Order Attention Networks~\cite{Dai2019}, and Diffusion-based Models (SRDiff)~\cite{Li2022_8}. 
	
	For videos, existing approaches include Video Super Resolution Networks (VSRNet)~\cite{Kappeler2016}, Video Efficient Sub-Pixel Convolutional Networks (VESPCN)~\cite{Caballero2017}, Detail-Revealing deep Video Super Resolution (DRVSR)~\cite{Tao2017}, Frame Recurrent Video Super Resolution (FRVSR)~\cite{Sajjadi2018}, Spatio-Temporal Transformer Networks (STTN)~\cite{Kim2018}, Recurrent Back-Projection Networks (RBPN)~\cite{Haris2019}, Task-Oriented Flows (TOFlow)~\cite{Xue2019}, Enhanced Deformable Video Restoration (EDVR)~\cite{Wang2019_5}, Fast Spatio-Temporal Residual Networks (FSTRN)~\cite{Li2019_6}, Recurrent Structure-Detail Networks (RSDN)~\cite{Isobe2020}, and Temporally Coherent GANs (TecoGAN)~\cite{Chu2020}.  \\

	For spatiotemporal scientific simulations, SRCNN~\cite{Dong2014,Dong2014_2} has been adapted for super-resolution of turbulent flows~\cite{Fukami2019}, a physics-informed super-resolution architecture for hyper-elastic modeling based on RDN~\cite{Zhang2018_9} was presented in~\cite{Arora2022}, and a modified FRVSR~\cite{Sajjadi2018} algorithm was designed to perform super-resolution for precipitation modeling in~\cite{Teufel2023}. Approaches involving convolutional neural networks have been considered in \cite{Gao2021,Fukami2021} for fluid flows, and a physics-informed combination of temporal interpolation, convolutional recurrent neural networks, residual blocks and pixelshuffle layers was carefully constructed in~\cite{Ren2022}. GAN-based approaches have also been explored in \cite{Li2022_7,Bode2021,PINN_SR} where modified and physics-informed versions of (E)SRGAN~\cite{Ledig2017,Wang2018} are introduced for super-resolution of turbulent flows and advection diffusion models, and a physics-informed diffusion-based super-resolution approach for of fluid flows was recently proposed in~\cite{Shu2023}. Closer in spirit to our approach, MeshFreeFlowNet~\cite{MeshFreeFlowNet} is a physics-informed super-resolution approach learning a continuous representation of the solution operator which can then be evaluated at any location and in particular on a higher resolution grid to obtain a high-resolution simulation.
	
In this paper, we propose an operator learning framework capable of removing constraints on the sensor and prediction locations when performing super-resolution to learn continuous representations of solutions to parametric differential equations from low-resolution approximations. In particular, our approach allows the consideration of numerous dynamical systems of interest in practice where the data is collected at random times by non-stationary sensors, that most of the other existing super-resolution and dynamics learning approaches listed here are not capable of addressing well.

	\newpage 
	
	\section{Super-resolution as an Operator Learning Problem}

	\subsection{Super Resolution Operator Networks (SROpNets)} 
	
	\hfill \\
	
	The proposed approach frames super-resolution as the operator learning problem of learning the operator $\mathcal{S}$ which returns the function $u$ given a LR representation $ u_{LR}$ of $u$. In practice and in the context of solving families of parametric PDEs, we often need to work with discrete data, so we wish to obtain a representation $u_\theta = \mathcal{S}_\theta(u_{LR})$ of the solution $u$ of a parametric differential equation which can then be evaluated on a higher-resolution grid to obtain a HR numerical approximation $u_{HR}$ to the function $u$. This allows us to leverage the advantages of the operator learning framework presented in Section~\ref{subsubsec: Neural Operators} (with $\mathcal{U} = \mathcal{V}$, $D := D_1 = D_2$ and $d := d_1 = d_2$) to learn parametric families of solutions without imposing restrictions on the spatiotemporal locations for the LR input and HR output. The proposed framework draws inspiration from existing operator learning architectures to combine deep learning methods for spatiotemporal data with additional neural networks for the input and prediction locations to allow mesh-free prediction without constraining the locations $\{ x_1, \ldots, x_s \}\subset D $ of the input values in the LR representation (aside from $s$ being fixed). 
	
	A first possibility is to treat time just like any other spatial coordinate and work with points in spacetime. In this case, our \textbf{Super Resolution Operator Network (SROpNet)} looks like \\

\input{diagrams/SR3Branch_Points}

\vspace{3mm}

	\noindent We can also fix the super-resolution scaling factor in time and work with sequences of spatial input, in which case our \textbf{Super Resolution Operator Network (SROpNet)} looks like \\

	\input{diagrams/SR3Branch_Sequences}

\vspace{3mm}
	
	In our framework, the branch and sensor networks can be based on any existing super-resolution or computer vision algorithm, and the trunk network is a multilayer perceptron. The sensor network allows to learn and predict from low-resolution data without requiring the input sensor locations~$x$ to be fixed, while the trunk network allows for mesh-free prediction. Note that MeshFreeFlowNet~\cite{MeshFreeFlowNet} is a special case of a slightly modified version of our approach without sensor network, where the input $u_{LR}$ is encoded into a latent vector which is stacked with the prediction location $y$ before going through a multilayer perceptron to produce the solution model prediction at $y$.
	
	In summary, we combine the use of sensor and trunk networks to propose an operator learning framework without constraints on the sensor and prediction locations, which allows to consider problems that vanilla DeepONet and most existing super-resolution approaches cannot address well.

	\subsubsection{Physics-Informing}
	
	\hfill \\
	
	As for  PINNs in function learning and for learning PDE solution operators from initial and boundary conditions, we can incorporate prior knowledge of the dynamics by adding extra loss terms, to minimize deviations from physics laws and governing differential equations in addition to minimizing the distance to reference solutions.  \\
	
While adding a physics loss when training neural architectures has proven very useful and successful for numerous applications, minimizing the resulting composite loss $$\mathcal{L} =  \mathcal{L}_{data}  +  \mathcal{L}_{physics} $$ can be much more challenging, and each iteration in the training process can become significantly more expensive in terms of computational time and memory. The optimization task might also become more prone to numerical issues in practice~\cite{Krishnapriyan2021,Wang2021,Wang2022_3,Rohrhofer2022,Leiteritz2021}. For instance, the composite objective function could have much worse conditioning, there could be conflicts between the multiple objectives, or the model could converge to a trivial or non-desired solution that satisfies well the physics constraints on the discrete set of collocation points where the physics loss is computed. In addition, the numerical method and the choice of discrete collocation points used to approximate the derivatives in the physics loss also induce errors that may bias the model towards non-desired and possibly non-physical solutions. \\

	Although strategies have been developed to mitigate the presence and impact of these issues, they usually come with an extra additional cost, so it is worth evaluating whether incorporating prior physics knowledge in the loss leads to better solutions at an acceptable additional computational cost for the problem of interest. This is even more relevant in the super-resolution context, since a lot of information about the physics and the PDE dynamics is already encoded in the low-resolution simulation inputs of the models. Even without an explicit loss term enforcing the physics constraints, our framework is already physics-informed or at least physics-aware. In some cases, the information provided by a physics loss might mostly be redundant, and in other cases it could be conflicting with the information already provided by the simulation data, thereby making the optimization task significantly harder and more ill-conditioned. 
	
	In most of our numerical experiments, adding a PDE loss came at a significant additional computational cost and lead to conflicting objectives issues which we were not able to resolve satisfactorily with our limited computational resources, despite numerous attempts using adaptive weighing schemes for the different loss terms such as ReLoBRaLo~\cite{Bischof2021}.   \\

	There is however one very important advantage of using a physics loss for solving families of parametric PDEs or performing super-resolution of parametric PDE solutions. When good high-resolution simulation data is not available, we can still try to learn the parametric solution operator by minimizing the physics loss solely, that is $\mathcal{L} = \mathcal{L}_{physics}$. As before, this task can prove very challenging, but would be very fruitful if successful, since it is very often the case that generating good high-resolution training data is very expensive or not realistic. 
	
	\hfill \\

	\subsection{SROpNet Architecture Examples}
	
	\hfill \\
	
	Although SROpNets can in principle be used with any choice of deep learning architecture in the branch and sensor subnetworks, we will restrict ourselves to relatively simple and small architectures for simplicity of exposition and because of limited computational resources.  
	
	\vspace{1mm}

	\newpage

	\subsubsection{Examples of SROpNets using simple combinations of MLPs, CNNs, and LSTMs} \label{subsec: Example CNN+LSTM+MLP}
	
	\hfill \\
	
In our numerical experiments, we will mostly use architectures composed of Multilayer Perceptrons (MLPs), Convolutional Neural Networks (CNNs), and Long Short-Term Memory networks (LSTMs). When the sensor locations are fixed, we do not need the sensor network and use variants of the 2-Subnetwork SROpNet depicted below. We also propose a similar architecture which only takes the initial state as input, to later assess the benefits of taking the full sequence of low-resolution states as input. We also depict the 3-Subnetwork SROpNet used in our numerical experiment where the sensor and prediction locations are not fixed. \\

\input{diagrams/Architecture_Example}

\newpage

	\subsubsection{Examples of SROpNets based on existing super-resolution algorithms}  \label{subsec: Example Computer Vision}
	
	\hfill \\
	
	\vspace{-1.5mm} 
	
	We can use existing super-resolution approaches from computer vision in our branch network, and connect them to a trunk (and sensor) network to obtain a mesh-free super-resolution model:

	\vspace{1mm} 
	
	\input{diagrams/ComputerVision_Example}
	
	\vspace{1.5mm} 
	
	Here, we use CARN~\cite{Ahn2018}, EDSR~\cite{Lim2017}, FSRCNN~\cite{Dong2016} and RCAN~\cite{RCAN}, whose architecture diagrams (extracted from \cite{Dong2016,Lim2017,Ahn2018,RCAN}) are displayed below. We can also use modified versions of existing approaches and start from pre-trained models to improve efficiency and reduce training time.

	\vspace{1mm}

	\begin{figure}[ht] 
		\begin{subfigure}[b]{0.67\linewidth}
		\centering
		\includegraphics[width=\linewidth]{CARN.png}  \vspace{0.4mm}
		\caption{CARN~\cite{Ahn2018}} 
	\end{subfigure} 
	\hfill 
	\begin{subfigure}[b]{0.29\linewidth}
		\centering
		\includegraphics[width=\linewidth]{EDSR.png}  \vspace{-4.8mm}
		\caption{EDSR~\cite{Lim2017}} 
	\end{subfigure}%% 
		
		\vspace{4mm}

		\begin{subfigure}[b]{\linewidth}
		\centering
		\includegraphics[width=0.93\linewidth]{FSRCNN.png} 
		\caption{FSRCNN~\cite{Dong2016}} 
	\end{subfigure}%%
	\vspace{4mm}
		
		\begin{subfigure}[b]{\linewidth}
			\centering
			\includegraphics[width=0.83\linewidth]{RCAN.png}  \vspace{-0.8mm}
			\caption{RCAN~\cite{RCAN}} 
		\end{subfigure}%%
	\end{figure}

	\newpage

	\section{Numerical Experiments} 
	
	\hfill 
	
	We now test SROpNets on diverse super-resolution problems:
	
	\begin{table}[!h]
		\begin{tabular}{l|c|c|c}  
			\hline 
			\textbf{Problem Considered}    & \ \textbf{Super-resolution}  \ & \textbf{Quantities Varied}  &  \textbf{Section}   \\ \hline
			1D Forced Diffusion
			&   
			Space $\times 4$, Time $\times 2$   &   Forcing Term                                             &           \ref{subsubsec: 1D Diffusion with Parametric Forcing}           \\ \hline   1D Forced Diffusion  &  
			Space $\times 4$, Time $\times 2$    &  Initial State                                         &          \ref{subsubsec: Forced 1D Diffusion with Varied Initial Conditions}       
			\\ \hline   1D Forced Diffusion
			&     Spacetime $\times 25$                 &  \  Forcing Term     $\  \big| \   $     LR/HR Locations \   &       \ref{subsubsec: 1D Diffusion Varied Locations}                           \\ \hline    2D Diffusion
			&  
			Space $\times 9$, Time $\times 2$                      &  Initial State       &                 \ref{subsubsec: 2D Diffusion with Fixed Diffusion Constant}                \\ \hline    2D Diffusion 
			&  Space $\times 9$                            &   Diffusion Constant     $\  \big| \   $      Initial State                              &    \ref{subsubsec: 2D Diffusion with Variable Diffusion Constant}     \\ \hline   2D Forced Diffusion
			&           Space $\times 9$                &   Forcing Term     $\  \big| \   $     Initial State    &                  \ref{subsec: 2D Forced Diffusion}    \\ \hline 2D Kolmogorov Flow
			&      Space $\times 16$                   &    Initial State       &      \ref{subsec: 2D Kolmogorov Flow}             
			\\ \hline  2D Kolmogorov Flow
			&             Space $\times 16$        &   Reynolds Number   $\  \big| \   $   Initial State      &  \ref{subsec: 2D Kolmogorov Flow}  \\ \hline 
		\end{tabular}
	\end{table}

	Note that most of the numerical results in this manuscript are best displayed as short videos, which can be found at \href{https://github.com/vduruiss/SROpNet}{github.com/vduruiss/SROpNet}. Note as well that
	the hyperparameters in our architectures have not been optimized to maximize the quality of our training outcomes, and the results could benefit from further hyperparameter tuning or be improved with more sophisticated and larger networks.
	
	\hfill \\
	
	\subsection{1D Forced Diffusion}

	\hfill \\
	
	In this section, we will consider various datasets for super-resolution of numerical solutions to the 1D Forced Diffusion equation
	\begin{equation}
		\frac{\partial u}{\partial t} = D \frac{\partial^2 u}{\partial x^2} + F(x,t),
	\end{equation}
	with force term $F(x,t)$ and constant diffusion coefficient $D$.

	\hfill

	\subsubsection{1D Diffusion with Parametric Forcing} \label{subsubsec: 1D Diffusion with Parametric Forcing}

	\hfill \\

	We first consider the parametric 1D Forced Diffusion equation
	\begin{equation} \label{eq: Diffusion 1D Exp1} 
		\frac{\partial u}{\partial t} = D \frac{\partial^2 u}{\partial x^2} + \frac{  \beta + D \alpha}{50} e^{-2\beta t}  \sin (\alpha x)  \qquad x\in [-1,1], \ t\in [0,2],
	\end{equation}
	with zero boundary and initial conditions, and constant diffusion coefficient $D = 1/1000$. Each sample in the dataset has different randomly-selected values for the parameters $\alpha \in [-6,6]$ and $\beta \in [-1,1]$. The low-resolution and high-resolution simulations in the dataset are generated using a simple finite difference solver. Here, we perform super-resolution both in space ($\times 4$ from $16$ to $64$) and in time ($\times 2$ from $40$ to $80$).  \\
	 
	We first use a LSTM-based 2-Subnetwork SROpNet with a [Time-Upscaling $\rightarrow$ LSTM $\rightarrow$ MLP] branch network and a MLP trunk network. This is essentially the architecture displayed in Section~\ref{subsec: Example CNN+LSTM+MLP} without CNN in the branch network. We also use the 2-Subnetwork SROpNets presented in Section~\ref{subsec: Example Computer Vision}, using the existing super-resolution approaches CARN~\cite{Ahn2018}, EDSR~\cite{Lim2017}, FSRCNN~\cite{Dong2016} and RCAN~\cite{RCAN}, as part of the branch network. 
	
	We can see from Figure~\ref{fig: Diffusion 1D Exp1} that all the different SROpNets used learn perfectly the super-resolution map of the diffusion dynamics for this parametric forced diffusion dataset.

\input{TexFiles/Forced1D_Diffusion_Exp1.tex}

\newpage

	\subsubsection{Forced 1D Diffusion with Varied Initial Conditions} \label{subsubsec: Forced 1D Diffusion with Varied Initial Conditions}
	
	\hfill  \\

	Next, we consider the 1D Forced Diffusion PDE
	\begin{equation} \label{eq: Diffusion 1D Exp2}
		\frac{\partial u}{\partial t} \  =  \ \frac{1}{50} \frac{\partial^2 u}{\partial x^2} \ + \ \frac{3}{5}  \sin  (12x)   e^{0.2x - 0.5t}       \qquad x\in [0,2], \ t\in [0,1],
	\end{equation}
	with zero boundary conditions. In this numerical experiment, each sample in the dataset has a different initial state $u(x,0)$, obtained by specifying random constant values on a random number of randomly-sized and randomly-located intervals in $[0,2]$. The low-resolution and high-resolution simulations in the dataset are generated using a simple finite difference solver. Here, we perform super-resolution both in space ($\times 4$ from $24$ to $96$) and in time ($\times 2$ from $50$ to $100$). 
	
	We use the 2-Subnetwork SROpNet displayed in Section~\ref{subsec: Example CNN+LSTM+MLP} without CNN in the branch network. Figure~\ref{fig: Diffusion 1D Exp2} shows that our SROpNet learns correctly the super-resolution map for these dynamics.

\input{TexFiles/Forced1D_Diffusion_Exp2.tex}

	\subsubsection{1D Diffusion with Parametric Forcing and Varied Sensor and Prediction Locations} \label{subsubsec: 1D Diffusion Varied Locations}

	\hfill  \\

	Finally, we consider the parametric 1D Forced Diffusion equation
	\begin{equation}   \label{eq: Diffusion 1D with Random Grids}
		\frac{\partial u}{\partial t} = \frac{\partial^2 u}{\partial x^2} + \frac{1}{2}  \left[ (x^2-1)(\alpha^2 + \beta ) -2
		\right] \sin (\alpha x) e^{\beta t} - 2 \alpha x  \cos (\alpha x) e^{\beta t}  \quad \  x\in [-1,1], \ t\in [0,2],
	\end{equation}
	with zero boundary conditions, and initial condition $ u(x,0) = 0.5 + 0.5 (x^2 - 1) \sin (\alpha x)$ for $x \in [-1,1]$. Each sample in the dataset has different randomly-selected values for the parameters $\alpha \in [-8,8]$ and $\beta \in [-1,0]$, and has randomly-located low-resolution input locations and high-resolution output locations. Although it is not required in our framework, we keep the time locations uniformly spaced and fixed across the different samples in the dataset due to limitations in our computational resources. Only the spatial locations are randomly chosen and different for each data sample.

\begin{figure}[!h]
	\vspace{-1mm}
	\centering
	\includegraphics[width=0.92\textwidth]{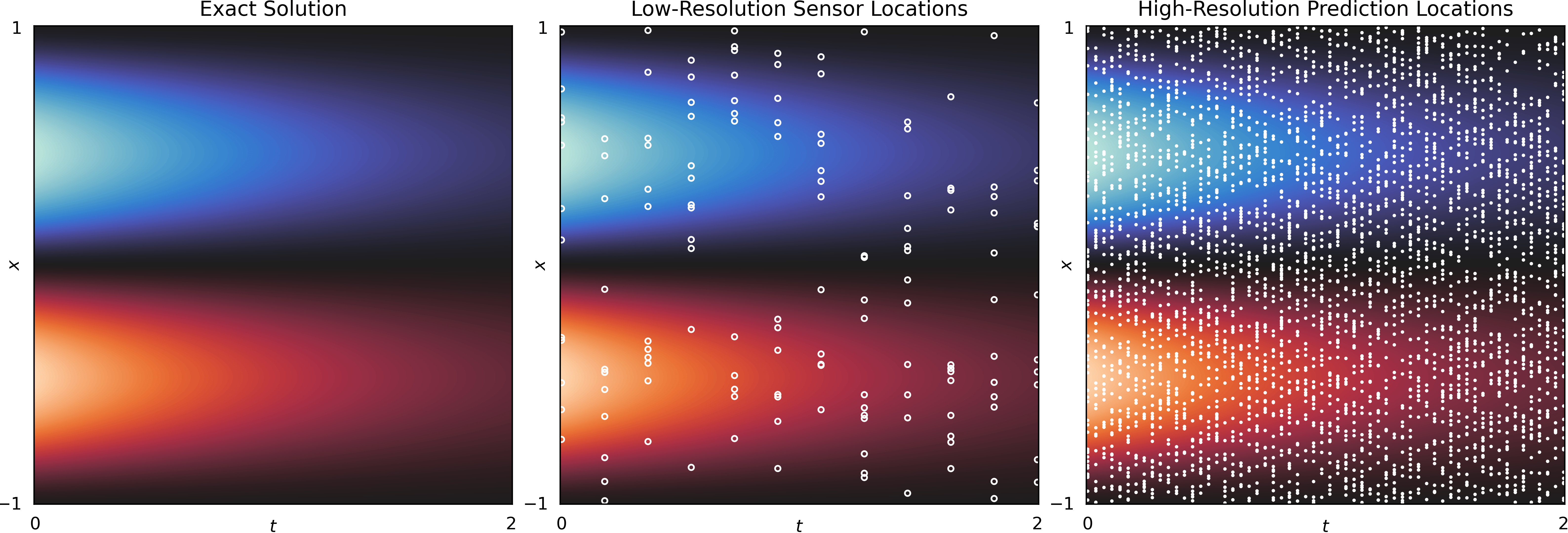}  \vspace{-3.2mm} 
	\caption{Randomly-selected 244 sensor locations and 3600 prediction locations. \label{fig: 1D Exp3 Locations} } \vspace{-0.6mm}
\end{figure}

	We perform $\times 25$ super-resolution from $12^2 = 244$ to $60^2 = 3600$ spatiotemporal points. Figure~\ref{fig: 1D Exp3 Locations} shows a solution of the forced diffusion equation~\eqref{eq: Diffusion 1D with Random Grids} with random sensor and prediction locations. The dataset is generated by evaluating the exact solution $u(x,t)  =  0.5 +  0.5 e^{\beta t} (x^2-1) \sin (\alpha x) $ at the randomly-selected low-resolution sensor and high-resolution prediction locations. 

	To take into account the varied sensor locations, we use the 3-Subnetwork SROpNet displayed in Section~\ref{subsec: Example CNN+LSTM+MLP}. Figure~\ref{fig: 1D Exp3 Results} shows that the SROpNet learns well the operator from low-resolution simulations at random sensor locations to continuous representations of the solutions. On the other hand, cubic interpolation does not extrapolate (missing pixels near edges) and does not perform well on certain examples due to the scarcity of sensor information in certain regions.

\input{TexFiles/Forced1D_Diffusion_Exp3.tex}

	\newpage 
	
	\subsection{2D Diffusion}
	
	\hfill \\
	
	We now consider super-resolution of numerical solutions to the 2D Diffusion PDE
	\begin{equation} \label{eq: 2D Diffusion}
		\frac{\partial u}{\partial t} = D  \left( \frac{\partial^2 u}{\partial x_1^2} + \frac{\partial^2 u}{\partial x_2^2} \right) \qquad x_1,x_2 \in [0,4], \ t\in [0,1],
	\end{equation}
	with constant diffusion coefficient $D$ and zero boundary conditions. In this section, the low-resolution and high-resolution simulations in the dataset are generated using a finite difference solver.

	\hfill 
	
	\subsubsection{2D Diffusion with Fixed Diffusion Constant}  \label{subsubsec: 2D Diffusion with Fixed Diffusion Constant}
	
	\hfill \\
	
	We first consider the 2D Diffusion equation~\eqref{eq: 2D Diffusion} with fixed diffusion constant $D=0.15$. Each sample in the dataset has a different initial state $u(x_1,x_2,0)$, obtained by specifying random constant values on a random number of randomly-sized and randomly-located disks in $[0,4]^2$. Here, we perform super-resolution both in space ($\times 9$ from $24\times 24$ to $72\times 72$) and in time ($\times 2$ from $25$ to $50$). 
	
\begin{figure}[!h]
		\centering
		\includegraphics[width=0.995\textwidth]{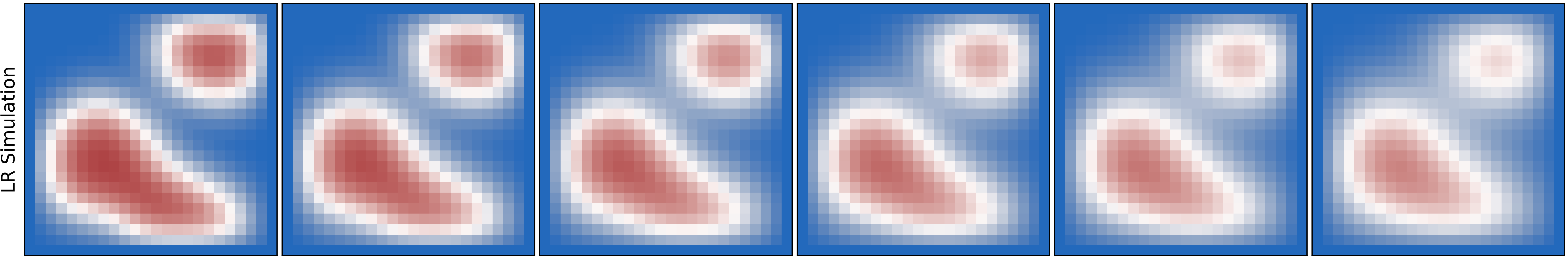} \vspace{-6.5mm}   
		\caption{Example of low-resolution simulation from the dataset of solutions to the 2D Diffusion equation~\eqref{eq: 2D Diffusion} with fixed diffusion constant. \label{fig: Diffusion 2D LR Example} } \vspace{2.5mm} 
	\end{figure}

	We use a 2-Subnetwork SROpNet with a [Time-Upscaling $\rightarrow$ CNN $\rightarrow$ LSTM $\rightarrow$ MLP] branch network and a MLP trunk network. This is the super-resolution architecture displayed in Section~\ref{subsec: Example CNN+LSTM+MLP}. We can see from Figure~\ref{fig: Diffusion 2D} that our architecture learns perfectly the map from low-resolution numerical solutions (such as the one shown in Figure~\ref{fig: Diffusion 2D LR Example}) to high-resolution numerical solutions for the diffusion dynamics of this dataset.

	\subsubsection{2D Diffusion with Variable Diffusion Constant}  \label{subsubsec: 2D Diffusion with Variable Diffusion Constant}

	\hfill \\
	
	Next, we consider the 2D Diffusion equation~\eqref{eq: 2D Diffusion} with variable diffusion constant $D$. Each sample in the dataset now has a different randomly-selected diffusion constant $D$ in $ [0.1,0.4]$ and a different initial state $u(x_1,x_2,0)$, obtained by specifying random constant values on a random number of randomly-sized and randomly-located disks in $[0,4]^2$. Here, we perform super-resolution in space ($\times 9$ from $24\times 24$ to $72\times 72$) while the time resolution remained at $100$. 
	
	We use a 2-Subnetwork SROpNet with [CNN $\rightarrow$ LSTM $\rightarrow$ MLP] branch network and a MLP trunk network, which is the architecture displayed in Section~\ref{subsec: Example CNN+LSTM+MLP} without the time-upscaling layer. We can see from Figure~\ref{fig: Diffusion 2D Speeds} that our SROpNet learns perfectly the map from low-resolution to high-resolution numerical solutions for the diffusion dynamics in this dataset. 
	
	To explore the benefits of taking the full sequence of low-resolution states as input in our approach, we performed a comparison with a similar operator learning architecture (the nonSR architecture displayed in Section~\ref{subsec: Example CNN+LSTM+MLP}) which only takes the high-resolution initial state as input of the branch network. Furthermore, we tested the trained architectures on diffusion dynamics with larger diffusion constants $D=0.5, \ 0.6, \ 0.8$ outside the interval $[0.1,0.4]$ of diffusion constants experienced during training. We can clearly see from Figure~\ref{fig: Diffusion 2D Speeds Comparison} that our super-resolution approach generalizes seamlessly to these faster diffusing dynamics while the other approach fails to simulate the diffusion mechanism at the correct speed due to insufficient information, as expected.

\input{TexFiles/2DDiffusion_Fixed.tex}

\input{TexFiles/2DDiffusion_VariableSpeeds.tex}

	\newpage

	\subsection{2D Forced Diffusion}  \label{subsec: 2D Forced Diffusion}
	
	\hfill \\

	We now consider super-resolution of numerical solutions to the 2D Forced Diffusion PDE
	\begin{equation} \label{eq: Forced 2D Diffusion}
		\frac{\partial u}{\partial t} = D  \left( \frac{\partial^2 u}{\partial x_1^2} + \frac{\partial^2 u}{\partial x_2^2} \right) + F(x_1, x_2,t),
	\end{equation}
	with force term $F(x_1,x_2,t)$ and constant diffusion coefficient $D=0.1$.

	Each sample in the dataset has a different initial state $u(x_1,x_2,0)$, obtained by specifying random constant values on a random number of randomly-sized and randomly-located disks in $[0,4]^2$. Each sample in the dataset also has a different forcing term $F(x_1,x_2,t)$ consisting of a randomly-located constant value spiraling structure. As before, the low-resolution and high-resolution datasets are generated using a simple finite difference solver. Here, we perform super-resolution in space ($\times 9$ from $24\times 24$ to $72\times 72$) while maintaining the time resolution to $30$. This is a more challenging problem because both the initial condition and forcing term are varied, and because the spiraling forcing structures are not well-resolved in the low-resolution simulations. \\
	
	We use a 2-Subnetwork SROpNet with a [CNN $\rightarrow$ LSTM $\rightarrow$ MLP] branch network and a MLP trunk network. This is essentially the architecture displayed in Section~\ref{subsec: Example CNN+LSTM+MLP} without the time-upscaling layer. We can see from Figure~\ref{fig: Forced Diffusion 2D} that the SROpNet learns well the map to go from low-resolution to high-resolution simulations for the forced diffusion dynamics, although the spiraling forcing structures tend to be blurrier and thicker in the model predictions.

\input{TexFiles/2DForcedDiffusion.tex}

	\hfill \\
	
	\subsection{2D Kolmogorov Flow} \label{subsec: 2D Kolmogorov Flow}
	
	\hfill \\

	Finally, we consider super-resolution of numerical solutions to the 2D Navier--Stokes equations in vorticity form for a viscous incompressible fluid:
	\begin{equation}
		\partial_t \omega (x,t) + u(x,t) \cdot \nabla \omega(x,t) = \frac{1}{Re} \Delta \omega(x,t) + f(x)  \qquad  x\in (0,1)^2, \ \ t \in (0,T],
	\end{equation}
	\begin{equation}
		\nabla \cdot u(x,t) = 0   \ \  \qquad  x\in (0,1)^2, \ \ t \in (0,T],
	\end{equation}
	\begin{equation}
		\omega(x,0) = \omega_0(x)  \qquad \ \    x\in (0,1)^2,
	\end{equation}
	where $\omega$ is vorticity, $u$ represents the velocity field, and $Re$ is the Reynolds number. Here, the initial condition $\omega_0(x)$ is sampled from a Gaussian random field, and the source term $f(x)$ is given by  \begin{equation}
		f(x) = 0.1 \left[ \sin(2\pi(x_1 + x_2)) + \cos(2\pi(x_1 + x_2))\right] .
	\end{equation}
	
	As described in \citep{li2023}, this dataset, currently available online\footnote{ \ \url{https://drive.google.com/drive/folders/1z-0V6NSl2STzrSA6QkzYWOGHSTgiOSYq} } under the courtesy of the authors of~\citep{FNO}, was generated by solving the stream function formulation of the Navier--Stokes equation using a pseudo-spectral method on a $256\times 256$ grid. At each time step, the stream function $\psi$ (such that $u = \nabla \times \psi)$ is first calculated
	by solving a Poisson equation $\Delta \psi = -\omega$. Then the nonlinear term is calculated by differentiating
	vorticity in Fourier space, dealiased and multiplied by velocities in physical space. The
	Crank--Nicolson scheme is then applied to update the state. The corresponding low-resolution simulation is then obtained by downsampling this numerical solution. \\ 
	
	In both the numerical experiments of this section, we will use the 2-Subnetwork SROpNet presented in Section~\ref{subsec: Example CNN+LSTM+MLP} which is composed of a [CNN $\rightarrow$ LSTM $\rightarrow$ MLP] branch network and a MLP trunk network.

	We first consider the case where all the samples in the dataset correspond to a 2D Kolmogorov flow with the same Reynolds number $Re = 20$. We can see from Figure~\ref{fig: Kolmogorov Re20} that our architecture learns perfectly the super-resolution map for the fluid dynamics of this dataset.

\input{TexFiles/Kolmogorov1.tex}

Next, we repeat the same experiment for the 2D Kolmogorov flow with Reynolds number $Re = 20$, but now only use a partial low-resolution simulation (in our case the first half, on $[0,T/2]$) as input of the branch network, but still predict on the full time interval $[0,T]$.

	We can see from Figure~\ref{fig: Kolmogorov Re20_Partial} that the SROpNet learns perfectly the super-resolution map of the fluid dynamics for this dataset, both in the first half of the time interval for which the low-resolution simulation was included as an input, and in the future in the second half of the time interval where no low-resolution counterpart was provided to the SROpNet as input. 
	
	\newpage

\input{TexFiles/Kolmogorov2.tex}

Lastly, we consider the case where each sample in the dataset corresponds to a 2D Kolmogorov flow with a different randomly-selected Reynolds number $Re$ in  $[200,500]$. This is a significantly more challenging problems since different values of $Re$ can have different characteristic dynamics and higher values of $Re$ lead to more complicated and faster-varying flows with finer structures. 
	
	The results for 3 previously-unobserved sequences (displayed as rows in Figure~\ref{fig: Kolmogorov MixedRe}) show that the SROpNet was able to capture well these fluid dynamics. Note however that bigger subnetworks and more careful hyperparameter tuning were needed, and that the later frames with finer structures are not resolved as perfectly.

\input{TexFiles/Kolmogorov3.tex}

	\newpage 

\vspace{2mm}
			
	\section{Conclusion}
	
		\vspace{2mm} 
		
			In this article, we have introduced a novel general framework for spatiotemporal super-resolution of numerical solutions to parametric partial differential equations based on operator learning. The proposed approach, the Super Resolution Operator Network (SROpNet), frames super-resolution as an operator learning problem to obtain continuous representations of solutions to parametric differential equations from low-resolution numerical approximations. 
			
			No restrictions are imposed on the spatiotemporal sensor locations for the low-resolution approximations (aside from the number of sensors having to be fixed), and the learnt continuous representations can be evaluated at any desired location. This allows to deal with numerous systems of interest in practice where the data is collected at non-uniformly spaced times (which can be either voluntary based on some prior knowledge of the system's dynamics, or involuntary due to precision and accuracy limitations of sensing equipment) by non-stationary sensors (such as sensors moving naturally in a fluid, weather stations for meteorological forecasting, or guided probing devices), where most other existing super-resolution approaches do not apply naturally. This also permits the use of adaptive meshes to obtain a higher-resolution simulation in regions of fast-varying dynamics while reducing the number of points and computations where dynamics vary slowly (as is frequently done by traditional PDE solvers via domain decomposition and adaptive mesh refinement techniques). Note that our approach could be used for image and video mesh-free super-resolution as well, although most of traditional computer vision and image rendering is performed using fixed uniform spatial grids and equidistant time frames. 
			
			We have demonstrated the effectiveness of our approach on a range of families of parametric PDEs. Equipped with relatively simple deep learning architectures as subnetworks, our framework successfully learnt diffusion and fluid dynamics from datasets where different data samples may satisfy different partial differential equations within a parametric family and can have different initial states and sensor/prediction spatiotemporal locations. We have also seen an example of a family of 2D diffusion equations where the learnt model achieves good results seamlessly on examples with parameter values outside the distribution of values experienced during training, where approaches learning from initial state only failed as expected due to insufficient information.	 \\
			
			There are many possible directions worth exploring in the future, which we have not been able to address in this article due to limitations in our computational resources. 
			
			It would be interesting to see how our approach performs on significantly more challenging problems in higher dimensions and for higher resolutions, when paired up with more sophisticated and larger computer vision architectures for the subnetworks. In particular, it would be very interesting to test our framework on large real-world problems or industrial dynamical systems where the data is collected at random times by non-stationary sensors, since these are problems that most other existing methods are not capable of addressing well.
			
			 In the spirit of implementing existing computer vision architectures within our framework, it would also be useful to investigate how well pre-trained models can be used as subnetworks to reduce significantly the amount of data samples and training needed. 
			
			It could also be beneficial to investigate and design new ways to combine the different subnetworks in our framework, and study when partial low-resolution simulations are sufficient (and explore how the quality of the results changes with the number or frequency of frames in the low-resolution input) instead of using the full low-resolution simulations as inputs of the branch subnetwork. 
			
			One could also try to replicate within our SROpNet framework the numerous modified and generalized versions of PINNs and DeepONets which have been designed in the past few years to improve robustness and efficiency of these function and operator learning approaches to tackle specific classes of dynamical systems in various contexts. 
			
			It is also worth exploring whether we can remove the restriction on the number of sensor locations for the low-resolution simulation, which has to be fixed and common to all training and prediction examples within our framework. There might also be interesting parallels to draw with the theory of compressed sensing, that could better inform the choice of the number of sensors for the low-resolution simulations as well as the locations of these sensors to obtain optimal reconstruction power given a SROpNet architecture and a problem of interest.
			
			Finally, it would be worth easing the process of adding a physics loss and training the resulting composite loss in our framework. Although we did not need it and it did not help for the examples considered in this paper, there might be more complicated problems where adding a physics loss could facilitate unraveling additional finer dynamical structures which only appear as the resolution increases (e.g. small eddies in turbulent fluid flows). Most importantly, it would be very important for practical applications to explore learning SROpNets without high-resolution training data, by minimizing only carefully defined physics loss functions only.  \\

	\bibliographystyle{abbrvnat}
	\bibliography{Bibliography}

\end{document}

%% file: diagrams/General2Branch.tex
\begin{center}

\resizebox{0.96\linewidth}{!}{

\tikzset{every picture/.style={line width=0.75pt}} %set default line width to 0.75pt        

\begin{tikzpicture}[x=0.75pt,y=0.75pt,yscale=-1,xscale=1]
%uncomment if require: \path (0,1580); %set diagram left start at 0, and has height of 1580

%Shape: Ellipse [id:dp6216862369514728] 
\draw   (30,925) .. controls (30,916.72) and (41.19,910) .. (55,910) .. controls (68.81,910) and (80,916.72) .. (80,925) .. controls (80,933.28) and (68.81,940) .. (55,940) .. controls (41.19,940) and (30,933.28) .. (30,925) -- cycle ;
%Straight Lines [id:da4605386849559763] 
\draw    (80,925.8) -- (157,925.61) ;
\draw [shift={(159,925.6)}, rotate = 179.85] [color={rgb, 255:red, 0; green, 0; blue, 0 }  ][line width=0.75]    (10.93,-3.29) .. controls (6.95,-1.4) and (3.31,-0.3) .. (0,0) .. controls (3.31,0.3) and (6.95,1.4) .. (10.93,3.29)   ;
%Shape: Rectangle [id:dp875865494480526] 
\draw   (161,911) -- (261,911) -- (261,941) -- (161,941) -- cycle ;
%Shape: Ellipse [id:dp8583423057557908] 
\draw   (30,984) .. controls (30,975.72) and (41.19,969) .. (55,969) .. controls (68.81,969) and (80,975.72) .. (80,984) .. controls (80,992.28) and (68.81,999) .. (55,999) .. controls (41.19,999) and (30,992.28) .. (30,984) -- cycle ;
%Straight Lines [id:da205347114668762] 
\draw    (80,985) -- (157,984.61) ;
\draw [shift={(159,984.6)}, rotate = 179.71] [color={rgb, 255:red, 0; green, 0; blue, 0 }  ][line width=0.75]    (10.93,-3.29) .. controls (6.95,-1.4) and (3.31,-0.3) .. (0,0) .. controls (3.31,0.3) and (6.95,1.4) .. (10.93,3.29)   ;
%Shape: Rectangle [id:dp7879675806429662] 
\draw   (161,970) -- (261,970) -- (261,1000) -- (161,1000) -- cycle ;
%Shape: Rectangle [id:dp39079505605519627] 
\draw   (370,940.5) -- (510,940.5) -- (510,970.5) -- (370,970.5) -- cycle ;
%Curve Lines [id:da8969334419941608] 
\draw    (261,926) .. controls (319.41,925.51) and (310.19,949.12) .. (366.28,950.56) ;
\draw [shift={(368,950.6)}, rotate = 180.99] [color={rgb, 255:red, 0; green, 0; blue, 0 }  ][line width=0.75]    (10.93,-3.29) .. controls (6.95,-1.4) and (3.31,-0.3) .. (0,0) .. controls (3.31,0.3) and (6.95,1.4) .. (10.93,3.29)   ;
%Curve Lines [id:da8672498381040883] 
\draw    (261,985) .. controls (321.39,984.51) and (311.21,961.08) .. (366.31,959.64) ;
\draw [shift={(368,959.6)}, rotate = 178.99] [color={rgb, 255:red, 0; green, 0; blue, 0 }  ][line width=0.75]    (10.93,-3.29) .. controls (6.95,-1.4) and (3.31,-0.3) .. (0,0) .. controls (3.31,0.3) and (6.95,1.4) .. (10.93,3.29)   ;
%Straight Lines [id:da13652105764168243] 
\draw    (510,956) -- (542,955.53) ;
\draw [shift={(544,955.5)}, rotate = 179.16] [color={rgb, 255:red, 0; green, 0; blue, 0 }  ][line width=0.75]    (10.93,-3.29) .. controls (6.95,-1.4) and (3.31,-0.3) .. (0,0) .. controls (3.31,0.3) and (6.95,1.4) .. (10.93,3.29)   ;
%Shape: Ellipse [id:dp49624443665106566] 
\draw   (547,956) .. controls (547,944.68) and (566.92,935.5) .. (591.5,935.5) .. controls (616.08,935.5) and (636,944.68) .. (636,956) .. controls (636,967.32) and (616.08,976.5) .. (591.5,976.5) .. controls (566.92,976.5) and (547,967.32) .. (547,956) -- cycle ;

% Text Node
\draw (50,919.5) node [anchor=north west][inner sep=0.75pt]  [font=\large] [align=left] {$\displaystyle u$};
% Text Node
\draw (81,908) node [anchor=north west][inner sep=0.75pt]  [font=\scriptsize,color={rgb, 255:red, 31; green, 19; blue, 254 }  ,opacity=1 ] [align=left] {$\displaystyle ( s)$};
% Text Node
\draw (168,918) node [anchor=north west][inner sep=0.75pt]  [font=\large] [align=left] {Branch Net};
% Text Node
\draw (50,978.5) node [anchor=north west][inner sep=0.75pt]  [font=\large] [align=left] {$\displaystyle y$};
% Text Node
\draw (81,989) node [anchor=north west][inner sep=0.75pt]  [font=\scriptsize,color={rgb, 255:red, 31; green, 19; blue, 254 }  ,opacity=1 ] [align=left] {$\displaystyle ( d_{v})$};
% Text Node
\draw (172,977) node [anchor=north west][inner sep=0.75pt]  [font=\large] [align=left] {Trunk Net};
% Text Node
\draw (392,947.5) node [anchor=north west][inner sep=0.75pt]  [font=\large] [align=left] {Inner Product};
% Text Node
\draw (265,989) node [anchor=north west][inner sep=0.75pt]  [font=\scriptsize,color={rgb, 255:red, 31; green, 19; blue, 254 }  ,opacity=1 ] [align=left] {$\displaystyle ( K)$};
% Text Node
\draw (264,909) node [anchor=north west][inner sep=0.75pt]  [font=\scriptsize,color={rgb, 255:red, 31; green, 19; blue, 254 }  ,opacity=1 ] [align=left] {$\displaystyle ( K)$};
% Text Node
\draw (561,948) node [anchor=north west][inner sep=0.75pt]  [font=\large] [align=left] {$\displaystyle \mathcal{G}_{\theta }( u)( y)$};
% Text Node
\draw (513,936) node [anchor=north west][inner sep=0.75pt]  [font=\scriptsize,color={rgb, 255:red, 31; green, 19; blue, 254 }  ,opacity=1 ] [align=left] {$\displaystyle ( 1)$};

\end{tikzpicture}

}

\end{center}

%% file: diagrams/General3Branch.tex
\begin{center}

\resizebox{0.95\linewidth}{!}{

\tikzset{every picture/.style={line width=0.75pt}} %set default line width to 0.75pt        

\begin{tikzpicture}[x=0.75pt,y=0.75pt,yscale=-1,xscale=1]
%uncomment if require: \path (0,1580); %set diagram left start at 0, and has height of 1580

%Shape: Ellipse [id:dp6216862369514728] 
\draw   (30,895) .. controls (30,886.72) and (41.19,880) .. (55,880) .. controls (68.81,880) and (80,886.72) .. (80,895) .. controls (80,903.28) and (68.81,910) .. (55,910) .. controls (41.19,910) and (30,903.28) .. (30,895) -- cycle ;
%Straight Lines [id:da4605386849559763] 
\draw    (80,895.8) -- (157,895.61) ;
\draw [shift={(159,895.6)}, rotate = 179.85] [color={rgb, 255:red, 0; green, 0; blue, 0 }  ][line width=0.75]    (10.93,-3.29) .. controls (6.95,-1.4) and (3.31,-0.3) .. (0,0) .. controls (3.31,0.3) and (6.95,1.4) .. (10.93,3.29)   ;
%Shape: Rectangle [id:dp875865494480526] 
\draw   (161,881) -- (261,881) -- (261,911) -- (161,911) -- cycle ;
%Shape: Ellipse [id:dp8583423057557908] 
\draw   (30,1015) .. controls (30,1006.72) and (41.19,1000) .. (55,1000) .. controls (68.81,1000) and (80,1006.72) .. (80,1015) .. controls (80,1023.28) and (68.81,1030) .. (55,1030) .. controls (41.19,1030) and (30,1023.28) .. (30,1015) -- cycle ;
%Straight Lines [id:da205347114668762] 
\draw    (80,1016) -- (157,1015.61) ;
\draw [shift={(159,1015.6)}, rotate = 179.71] [color={rgb, 255:red, 0; green, 0; blue, 0 }  ][line width=0.75]    (10.93,-3.29) .. controls (6.95,-1.4) and (3.31,-0.3) .. (0,0) .. controls (3.31,0.3) and (6.95,1.4) .. (10.93,3.29)   ;
%Shape: Rectangle [id:dp7879675806429662] 
\draw   (161,1001) -- (261,1001) -- (261,1031) -- (161,1031) -- cycle ;
%Shape: Rectangle [id:dp39079505605519627] 
\draw   (370,941.5) -- (510,941.5) -- (510,971.5) -- (370,971.5) -- cycle ;
%Curve Lines [id:da8969334419941608] 
\draw    (261,896) .. controls (319.41,895.51) and (309.21,943.82) .. (365.28,945.76) ;
\draw [shift={(367,945.8)}, rotate = 180.99] [color={rgb, 255:red, 0; green, 0; blue, 0 }  ][line width=0.75]    (10.93,-3.29) .. controls (6.95,-1.4) and (3.31,-0.3) .. (0,0) .. controls (3.31,0.3) and (6.95,1.4) .. (10.93,3.29)   ;
%Curve Lines [id:da8672498381040883] 
\draw    (261,1016) .. controls (321.39,1015.51) and (310.23,966.79) .. (365.31,964.84) ;
\draw [shift={(367,964.8)}, rotate = 178.99] [color={rgb, 255:red, 0; green, 0; blue, 0 }  ][line width=0.75]    (10.93,-3.29) .. controls (6.95,-1.4) and (3.31,-0.3) .. (0,0) .. controls (3.31,0.3) and (6.95,1.4) .. (10.93,3.29)   ;
%Straight Lines [id:da13652105764168243] 
\draw    (510,957) -- (542,956.53) ;
\draw [shift={(544,956.5)}, rotate = 179.16] [color={rgb, 255:red, 0; green, 0; blue, 0 }  ][line width=0.75]    (10.93,-3.29) .. controls (6.95,-1.4) and (3.31,-0.3) .. (0,0) .. controls (3.31,0.3) and (6.95,1.4) .. (10.93,3.29)   ;
%Shape: Ellipse [id:dp49624443665106566] 
\draw   (547,957) .. controls (547,945.68) and (566.92,936.5) .. (591.5,936.5) .. controls (616.08,936.5) and (636,945.68) .. (636,957) .. controls (636,968.32) and (616.08,977.5) .. (591.5,977.5) .. controls (566.92,977.5) and (547,968.32) .. (547,957) -- cycle ;
%Shape: Ellipse [id:dp9339236734269698] 
\draw   (30,955) .. controls (30,946.72) and (41.19,940) .. (55,940) .. controls (68.81,940) and (80,946.72) .. (80,955) .. controls (80,963.28) and (68.81,970) .. (55,970) .. controls (41.19,970) and (30,963.28) .. (30,955) -- cycle ;
%Straight Lines [id:da48292758827035653] 
\draw    (80,955.8) -- (157,955.61) ;
\draw [shift={(159,955.6)}, rotate = 179.85] [color={rgb, 255:red, 0; green, 0; blue, 0 }  ][line width=0.75]    (10.93,-3.29) .. controls (6.95,-1.4) and (3.31,-0.3) .. (0,0) .. controls (3.31,0.3) and (6.95,1.4) .. (10.93,3.29)   ;
%Shape: Rectangle [id:dp40606700780506966] 
\draw   (161,941) -- (261,941) -- (261,971) -- (161,971) -- cycle ;
%Straight Lines [id:da35168746449256116] 
\draw    (261,956.2) -- (365,955.81) ;
\draw [shift={(367,955.8)}, rotate = 179.78] [color={rgb, 255:red, 0; green, 0; blue, 0 }  ][line width=0.75]    (10.93,-3.29) .. controls (6.95,-1.4) and (3.31,-0.3) .. (0,0) .. controls (3.31,0.3) and (6.95,1.4) .. (10.93,3.29)   ;

% Text Node
\draw (50,889.5) node [anchor=north west][inner sep=0.75pt]  [font=\large] [align=left] {$\displaystyle u$};
% Text Node
\draw (81,878) node [anchor=north west][inner sep=0.75pt]  [font=\scriptsize,color={rgb, 255:red, 31; green, 19; blue, 254 }  ,opacity=1 ] [align=left] {$\displaystyle ( s)$};
% Text Node
\draw (168,888) node [anchor=north west][inner sep=0.75pt]  [font=\large] [align=left] {Branch Net};
% Text Node
\draw (50,1009.5) node [anchor=north west][inner sep=0.75pt]  [font=\large] [align=left] {$\displaystyle y$};
% Text Node
\draw (81,1020) node [anchor=north west][inner sep=0.75pt]  [font=\scriptsize,color={rgb, 255:red, 31; green, 19; blue, 254 }  ,opacity=1 ] [align=left] {$\displaystyle ( d_{v})$};
% Text Node
\draw (172,1008) node [anchor=north west][inner sep=0.75pt]  [font=\large] [align=left] {Trunk Net};
% Text Node
\draw (392,948.5) node [anchor=north west][inner sep=0.75pt]  [font=\large] [align=left] {Inner Product};
% Text Node
\draw (265,1020) node [anchor=north west][inner sep=0.75pt]  [font=\scriptsize,color={rgb, 255:red, 31; green, 19; blue, 254 }  ,opacity=1 ] [align=left] {$\displaystyle ( K)$};
% Text Node
\draw (264,879) node [anchor=north west][inner sep=0.75pt]  [font=\scriptsize,color={rgb, 255:red, 31; green, 19; blue, 254 }  ,opacity=1 ] [align=left] {$\displaystyle ( K)$};
% Text Node
\draw (561,949) node [anchor=north west][inner sep=0.75pt]  [font=\large] [align=left] {$\displaystyle \mathcal{G}_{\theta }( u)( y)$};
% Text Node
\draw (50,949.5) node [anchor=north west][inner sep=0.75pt]  [font=\large] [align=left] {$\displaystyle x$};
% Text Node
\draw (81,960) node [anchor=north west][inner sep=0.75pt]  [font=\scriptsize,color={rgb, 255:red, 31; green, 19; blue, 254 }  ,opacity=1 ] [align=left] {$\displaystyle ( d_{u} ,s)$};
% Text Node
\draw (171,948) node [anchor=north west][inner sep=0.75pt]  [font=\large] [align=left] {Sensor Net};
% Text Node
\draw (265,960) node [anchor=north west][inner sep=0.75pt]  [font=\scriptsize,color={rgb, 255:red, 31; green, 19; blue, 254 }  ,opacity=1 ] [align=left] {$\displaystyle ( K)$};
% Text Node
\draw (513,937) node [anchor=north west][inner sep=0.75pt]  [font=\scriptsize,color={rgb, 255:red, 31; green, 19; blue, 254 }  ,opacity=1 ] [align=left] {$\displaystyle ( 1)$};

\end{tikzpicture}

}

\end{center}

%% file: diagrams/General3BranchOtherExamples.tex
\begin{center}

\resizebox{0.86 \linewidth}{!}{

\tikzset{every picture/.style={line width=0.75pt}} %set default line width to 0.75pt        

\begin{tikzpicture}[x=0.75pt,y=0.75pt,yscale=-1,xscale=1]
	%uncomment if require: \path (0,1580); %set diagram left start at 0, and has height of 1580
	
	%Shape: Ellipse [id:dp5428993451288555] 
	\draw   (80,175) .. controls (80,166.72) and (91.19,160) .. (105,160) .. controls (118.81,160) and (130,166.72) .. (130,175) .. controls (130,183.28) and (118.81,190) .. (105,190) .. controls (91.19,190) and (80,183.28) .. (80,175) -- cycle ;
	%Straight Lines [id:da9895671965183732] 
	\draw    (130,175) -- (180,174.81) ;
	\draw [shift={(182,174.8)}, rotate = 179.78] [color={rgb, 255:red, 0; green, 0; blue, 0 }  ][line width=0.75]    (10.93,-3.29) .. controls (6.95,-1.4) and (3.31,-0.3) .. (0,0) .. controls (3.31,0.3) and (6.95,1.4) .. (10.93,3.29)   ;
	%Shape: Ellipse [id:dp9094848183393782] 
	\draw   (80,295) .. controls (80,286.72) and (91.19,280) .. (105,280) .. controls (118.81,280) and (130,286.72) .. (130,295) .. controls (130,303.28) and (118.81,310) .. (105,310) .. controls (91.19,310) and (80,303.28) .. (80,295) -- cycle ;
	%Shape: Ellipse [id:dp5733676586815641] 
	\draw   (80,235) .. controls (80,226.72) and (91.19,220) .. (105,220) .. controls (118.81,220) and (130,226.72) .. (130,235) .. controls (130,243.28) and (118.81,250) .. (105,250) .. controls (91.19,250) and (80,243.28) .. (80,235) -- cycle ;
	%Curve Lines [id:da5261918290916219] 
	\draw    (130,236) .. controls (172.63,236.49) and (190.77,223.05) .. (199.95,194.28) ;
	\draw [shift={(200.5,192.5)}, rotate = 106.7] [color={rgb, 255:red, 0; green, 0; blue, 0 }  ][line width=0.75]    (10.93,-3.29) .. controls (6.95,-1.4) and (3.31,-0.3) .. (0,0) .. controls (3.31,0.3) and (6.95,1.4) .. (10.93,3.29)   ;
	%Shape: Rectangle [id:dp12753155308265107] 
	\draw   (184,161) -- (234,161) -- (234,191) -- (184,191) -- cycle ;
	%Straight Lines [id:da3254396865399476] 
	\draw    (234,175.8) -- (316,175.6) ;
	\draw [shift={(318,175.6)}, rotate = 179.86] [color={rgb, 255:red, 0; green, 0; blue, 0 }  ][line width=0.75]    (10.93,-3.29) .. controls (6.95,-1.4) and (3.31,-0.3) .. (0,0) .. controls (3.31,0.3) and (6.95,1.4) .. (10.93,3.29)   ;
	%Shape: Rectangle [id:dp15545145619848277] 
	\draw   (319.5,161) -- (419.5,161) -- (419.5,191) -- (319.5,191) -- cycle ;
	%Shape: Rectangle [id:dp3519042947117821] 
	\draw   (319.5,221) -- (419.5,221) -- (419.5,251) -- (319.5,251) -- cycle ;
	%Shape: Rectangle [id:dp4478559566109799] 
	\draw   (491.5,191.5) -- (521.73,191.5) -- (521.73,221.5) -- (491.5,221.5) -- cycle ;
	%Curve Lines [id:da4157239510025228] 
	\draw    (419.5,176) .. controls (447.57,179.94) and (449.93,195.52) .. (487.27,200.29) ;
	\draw [shift={(489,200.5)}, rotate = 186.58] [color={rgb, 255:red, 0; green, 0; blue, 0 }  ][line width=0.75]    (10.93,-3.29) .. controls (6.95,-1.4) and (3.31,-0.3) .. (0,0) .. controls (3.31,0.3) and (6.95,1.4) .. (10.93,3.29)   ;
	%Curve Lines [id:da5821090404920708] 
	\draw    (419.5,236) .. controls (448.56,231.08) and (449.96,216.45) .. (487.27,210.75) ;
	\draw [shift={(489,210.5)}, rotate = 171.97] [color={rgb, 255:red, 0; green, 0; blue, 0 }  ][line width=0.75]    (10.93,-3.29) .. controls (6.95,-1.4) and (3.31,-0.3) .. (0,0) .. controls (3.31,0.3) and (6.95,1.4) .. (10.93,3.29)   ;
	%Straight Lines [id:da13793204545441773] 
	\draw    (521.5,205.5) -- (560,205.02) ;
	\draw [shift={(562,205)}, rotate = 179.29] [color={rgb, 255:red, 0; green, 0; blue, 0 }  ][line width=0.75]    (10.93,-3.29) .. controls (6.95,-1.4) and (3.31,-0.3) .. (0,0) .. controls (3.31,0.3) and (6.95,1.4) .. (10.93,3.29)   ;
	%Shape: Ellipse [id:dp6349612571176779] 
	\draw   (563.5,205) .. controls (563.5,193.68) and (581.86,184.5) .. (604.5,184.5) .. controls (627.14,184.5) and (645.5,193.68) .. (645.5,205) .. controls (645.5,216.32) and (627.14,225.5) .. (604.5,225.5) .. controls (581.86,225.5) and (563.5,216.32) .. (563.5,205) -- cycle ;
	%Curve Lines [id:da10293634030876153] 
	\draw    (130,295) .. controls (229,278.58) and (239.4,244.84) .. (315.35,237.61) ;
	\draw [shift={(316.5,237.5)}, rotate = 174.81] [color={rgb, 255:red, 0; green, 0; blue, 0 }  ][line width=0.75]    (10.93,-3.29) .. controls (6.95,-1.4) and (3.31,-0.3) .. (0,0) .. controls (3.31,0.3) and (6.95,1.4) .. (10.93,3.29)   ;
	
	% Text Node
	\draw (100,169.5) node [anchor=north west][inner sep=0.75pt]  [font=\large] [align=left] {$\displaystyle u$};
	% Text Node
	\draw (131,158) node [anchor=north west][inner sep=0.75pt]  [font=\scriptsize,color={rgb, 255:red, 31; green, 19; blue, 254 }  ,opacity=1 ] [align=left] {$\displaystyle ( s)$};
	% Text Node
	\draw (100,289.5) node [anchor=north west][inner sep=0.75pt]  [font=\large] [align=left] {$\displaystyle y$};
	% Text Node
	\draw (131,299) node [anchor=north west][inner sep=0.75pt]  [font=\scriptsize,color={rgb, 255:red, 31; green, 19; blue, 254 }  ,opacity=1 ] [align=left] {$\displaystyle ( d_{v})$};
	% Text Node
	\draw (100,229.5) node [anchor=north west][inner sep=0.75pt]  [font=\large] [align=left] {$\displaystyle x$};
	% Text Node
	\draw (131,240) node [anchor=north west][inner sep=0.75pt]  [font=\scriptsize,color={rgb, 255:red, 31; green, 19; blue, 254 }  ,opacity=1 ] [align=left] {$\displaystyle ( d_{u} ,s)$};
	% Text Node
	\draw (188,168) node [anchor=north west][inner sep=0.75pt]  [font=\large] [align=left] {Stack};
	% Text Node
	\draw (235,158) node [anchor=north west][inner sep=0.75pt]  [font=\scriptsize,color={rgb, 255:red, 31; green, 19; blue, 254 }  ,opacity=1 ] [align=left] {$\displaystyle ( d_{u} \! +\! 1,s)$};
	% Text Node
	\draw (326.5,168) node [anchor=north west][inner sep=0.75pt]  [font=\large] [align=left] {Branch Net};
	% Text Node
	\draw (330.5,228) node [anchor=north west][inner sep=0.75pt]  [font=\large] [align=left] {Trunk Net};
	% Text Node
	\draw (501.5,202.5) node [anchor=north west][inner sep=0.75pt]  [font=\Huge] [align=left] {$\displaystyle \mathbf{\cdot }$};
	% Text Node
	\draw (423.5,240) node [anchor=north west][inner sep=0.75pt]  [font=\scriptsize,color={rgb, 255:red, 31; green, 19; blue, 254 }  ,opacity=1 ] [align=left] {$\displaystyle ( K)$};
	% Text Node
	\draw (422.5,159) node [anchor=north west][inner sep=0.75pt]  [font=\scriptsize,color={rgb, 255:red, 31; green, 19; blue, 254 }  ,opacity=1 ] [align=left] {$\displaystyle ( K)$};
	% Text Node
	\draw (574.5,197) node [anchor=north west][inner sep=0.75pt]  [font=\large] [align=left] {$\displaystyle \mathcal{G}_{\theta }( u)( y)$};
	% Text Node
	\draw (524.5,187) node [anchor=north west][inner sep=0.75pt]  [font=\scriptsize,color={rgb, 255:red, 31; green, 19; blue, 254 }  ,opacity=1 ] [align=left] {$\displaystyle ( 1)$};

\end{tikzpicture}

}

\vspace{-0.4mm}
 
\noindent \hspace{7mm}   \dotfill \hspace{9mm} 
 
\vspace{2.6mm}

\resizebox{0.86 \linewidth}{!}{

	\tikzset{every picture/.style={line width=0.75pt}} %set default line width to 0.75pt        
	
	\begin{tikzpicture}[x=0.75pt,y=0.75pt,yscale=-1,xscale=1]
		%uncomment if require: \path (0,1580); %set diagram left start at 0, and has height of 1580
		
		%Shape: Ellipse [id:dp04172903564570585] 
		\draw   (79,354) .. controls (79,345.72) and (90.19,339) .. (104,339) .. controls (117.81,339) and (129,345.72) .. (129,354) .. controls (129,362.28) and (117.81,369) .. (104,369) .. controls (90.19,369) and (79,362.28) .. (79,354) -- cycle ;
		%Shape: Ellipse [id:dp8183043858640087] 
		\draw   (80,474) .. controls (80,465.72) and (91.19,459) .. (105,459) .. controls (118.81,459) and (130,465.72) .. (130,474) .. controls (130,482.28) and (118.81,489) .. (105,489) .. controls (91.19,489) and (80,482.28) .. (80,474) -- cycle ;
		%Shape: Ellipse [id:dp25925887924702695] 
		\draw   (80,414) .. controls (80,405.72) and (91.19,399) .. (105,399) .. controls (118.81,399) and (130,405.72) .. (130,414) .. controls (130,422.28) and (118.81,429) .. (105,429) .. controls (91.19,429) and (80,422.28) .. (80,414) -- cycle ;
		%Shape: Rectangle [id:dp38584073281249864] 
		\draw   (190,430.5) -- (269.5,430.5) -- (269.5,460.5) -- (190,460.5) -- cycle ;
		%Straight Lines [id:da3132165003159728] 
		\draw    (269.86,446) -- (316.53,445.52) ;
		\draw [shift={(318.53,445.5)}, rotate = 179.41] [color={rgb, 255:red, 0; green, 0; blue, 0 }  ][line width=0.75]    (10.93,-3.29) .. controls (6.95,-1.4) and (3.31,-0.3) .. (0,0) .. controls (3.31,0.3) and (6.95,1.4) .. (10.93,3.29)   ;
		%Curve Lines [id:da9046757023030201] 
		\draw    (130,471) .. controls (160.54,464.6) and (149.35,452.86) .. (187.23,449.64) ;
		\draw [shift={(189,449.5)}, rotate = 175.71] [color={rgb, 255:red, 0; green, 0; blue, 0 }  ][line width=0.75]    (10.93,-3.29) .. controls (6.95,-1.4) and (3.31,-0.3) .. (0,0) .. controls (3.31,0.3) and (6.95,1.4) .. (10.93,3.29)   ;
		%Curve Lines [id:da6555612163585958] 
		\draw    (128,420) .. controls (158.54,425.42) and (150.26,437.14) .. (187.28,439.4) ;
		\draw [shift={(189,439.5)}, rotate = 182.94] [color={rgb, 255:red, 0; green, 0; blue, 0 }  ][line width=0.75]    (10.93,-3.29) .. controls (6.95,-1.4) and (3.31,-0.3) .. (0,0) .. controls (3.31,0.3) and (6.95,1.4) .. (10.93,3.29)   ;
		%Curve Lines [id:da37961949472913936] 
		\draw    (129,354) .. controls (243.43,354.5) and (210.34,386.18) .. (317.38,386.5) ;
		\draw [shift={(319,386.5)}, rotate = 180] [color={rgb, 255:red, 0; green, 0; blue, 0 }  ][line width=0.75]    (10.93,-3.29) .. controls (6.95,-1.4) and (3.31,-0.3) .. (0,0) .. controls (3.31,0.3) and (6.95,1.4) .. (10.93,3.29)   ;
		%Shape: Rectangle [id:dp007846103679048477] 
		\draw   (320.5,371) -- (420.5,371) -- (420.5,401) -- (320.5,401) -- cycle ;
		%Shape: Rectangle [id:dp044820888804993464] 
		\draw   (320.5,431) -- (420.5,431) -- (420.5,461) -- (320.5,461) -- cycle ;
		%Shape: Rectangle [id:dp2532698259112487] 
		\draw   (492.5,401.5) -- (522.73,401.5) -- (522.73,431.5) -- (492.5,431.5) -- cycle ;
		%Curve Lines [id:da3839053959241452] 
		\draw    (420.5,386) .. controls (448.57,389.94) and (450.93,405.52) .. (488.27,410.29) ;
		\draw [shift={(490,410.5)}, rotate = 186.58] [color={rgb, 255:red, 0; green, 0; blue, 0 }  ][line width=0.75]    (10.93,-3.29) .. controls (6.95,-1.4) and (3.31,-0.3) .. (0,0) .. controls (3.31,0.3) and (6.95,1.4) .. (10.93,3.29)   ;
		%Curve Lines [id:da6951565557082096] 
		\draw    (420.5,446) .. controls (449.56,441.08) and (450.96,426.45) .. (488.27,420.75) ;
		\draw [shift={(490,420.5)}, rotate = 171.97] [color={rgb, 255:red, 0; green, 0; blue, 0 }  ][line width=0.75]    (10.93,-3.29) .. controls (6.95,-1.4) and (3.31,-0.3) .. (0,0) .. controls (3.31,0.3) and (6.95,1.4) .. (10.93,3.29)   ;
		%Straight Lines [id:da19122885320108374] 
		\draw    (522.5,415.5) -- (561,415.02) ;
		\draw [shift={(563,415)}, rotate = 179.29] [color={rgb, 255:red, 0; green, 0; blue, 0 }  ][line width=0.75]    (10.93,-3.29) .. controls (6.95,-1.4) and (3.31,-0.3) .. (0,0) .. controls (3.31,0.3) and (6.95,1.4) .. (10.93,3.29)   ;
		%Shape: Ellipse [id:dp4211951395603333] 
		\draw   (564.5,415) .. controls (564.5,403.68) and (582.86,394.5) .. (605.5,394.5) .. controls (628.14,394.5) and (646.5,403.68) .. (646.5,415) .. controls (646.5,426.32) and (628.14,435.5) .. (605.5,435.5) .. controls (582.86,435.5) and (564.5,426.32) .. (564.5,415) -- cycle ;
		
		% Text Node
		\draw (99,348.5) node [anchor=north west][inner sep=0.75pt]  [font=\large] [align=left] {$\displaystyle u$};
		% Text Node
		\draw (130,337) node [anchor=north west][inner sep=0.75pt]  [font=\scriptsize,color={rgb, 255:red, 31; green, 19; blue, 254 }  ,opacity=1 ] [align=left] {$\displaystyle ( s)$};
		% Text Node
		\draw (100,468.5) node [anchor=north west][inner sep=0.75pt]  [font=\large] [align=left] {$\displaystyle y$};
		% Text Node
		\draw (130,475) node [anchor=north west][inner sep=0.75pt]  [font=\scriptsize,color={rgb, 255:red, 31; green, 19; blue, 254 }  ,opacity=1 ] [align=left] {$\displaystyle ( d_{u})$};
		% Text Node
		\draw (100,408.5) node [anchor=north west][inner sep=0.75pt]  [font=\large] [align=left] {$\displaystyle x$};
		% Text Node
		\draw (131,401) node [anchor=north west][inner sep=0.75pt]  [font=\scriptsize,color={rgb, 255:red, 31; green, 19; blue, 254 }  ,opacity=1 ] [align=left] {$\displaystyle ( d_{u} ,s)$};
		% Text Node
		\draw (198.56,437.5) node [anchor=north west][inner sep=0.75pt]  [font=\large] [align=left] {Distance};
		% Text Node
		\draw (272,450) node [anchor=north west][inner sep=0.75pt]  [font=\scriptsize,color={rgb, 255:red, 31; green, 19; blue, 254 }  ,opacity=1 ] [align=left] {$\displaystyle ( d_{u} ,s)$};
		% Text Node
		\draw (327.5,378) node [anchor=north west][inner sep=0.75pt]  [font=\large] [align=left] {Branch Net};
		% Text Node
		\draw (331.5,438) node [anchor=north west][inner sep=0.75pt]  [font=\large] [align=left] {Trunk Net};
		% Text Node
		\draw (502.5,412.5) node [anchor=north west][inner sep=0.75pt]  [font=\Huge] [align=left] {$\displaystyle \mathbf{\cdot }$};
		% Text Node
		\draw (424.5,450) node [anchor=north west][inner sep=0.75pt]  [font=\scriptsize,color={rgb, 255:red, 31; green, 19; blue, 254 }  ,opacity=1 ] [align=left] {$\displaystyle ( K)$};
		% Text Node
		\draw (423.5,369) node [anchor=north west][inner sep=0.75pt]  [font=\scriptsize,color={rgb, 255:red, 31; green, 19; blue, 254 }  ,opacity=1 ] [align=left] {$\displaystyle ( K)$};
		% Text Node
		\draw (575.5,407) node [anchor=north west][inner sep=0.75pt]  [font=\large] [align=left] {$\displaystyle \mathcal{G}_{\theta }( u)( y)$};
		% Text Node
		\draw (525.5,397) node [anchor=north west][inner sep=0.75pt]  [font=\scriptsize,color={rgb, 255:red, 31; green, 19; blue, 254 }  ,opacity=1 ] [align=left] {$\displaystyle ( 1)$};

	\end{tikzpicture}
	
}

\end{center}

%% file: diagrams/SR3Branch_Points.tex
\begin{center}

\resizebox{0.94\linewidth}{!}{

\tikzset{every picture/.style={line width=0.75pt}} %set default line width to 0.75pt        

\begin{tikzpicture}[x=0.75pt,y=0.75pt,yscale=-1,xscale=1]
%uncomment if require: \path (0,1580); %set diagram left start at 0, and has height of 1580

%Shape: Ellipse [id:dp6216862369514728] 
\draw   (1,895) .. controls (1,886.72) and (12.19,880) .. (26,880) .. controls (39.81,880) and (51,886.72) .. (51,895) .. controls (51,903.28) and (39.81,910) .. (26,910) .. controls (12.19,910) and (1,903.28) .. (1,895) -- cycle ;
%Straight Lines [id:da4605386849559763] 
\draw    (51,895) -- (168.4,895.39) ;
\draw [shift={(170.4,895.4)}, rotate = 180.19] [color={rgb, 255:red, 0; green, 0; blue, 0 }  ][line width=0.75]    (10.93,-3.29) .. controls (6.95,-1.4) and (3.31,-0.3) .. (0,0) .. controls (3.31,0.3) and (6.95,1.4) .. (10.93,3.29)   ;
%Shape: Rectangle [id:dp875865494480526] 
\draw   (172,881) -- (272,881) -- (272,911) -- (172,911) -- cycle ;
%Shape: Ellipse [id:dp8583423057557908] 
\draw   (1,1015) .. controls (1,1006.72) and (12.19,1000) .. (26,1000) .. controls (39.81,1000) and (51,1006.72) .. (51,1015) .. controls (51,1023.28) and (39.81,1030) .. (26,1030) .. controls (12.19,1030) and (1,1023.28) .. (1,1015) -- cycle ;
%Straight Lines [id:da205347114668762] 
\draw    (51,1016) -- (168.4,1016.39) ;
\draw [shift={(170.4,1016.4)}, rotate = 180.19] [color={rgb, 255:red, 0; green, 0; blue, 0 }  ][line width=0.75]    (10.93,-3.29) .. controls (6.95,-1.4) and (3.31,-0.3) .. (0,0) .. controls (3.31,0.3) and (6.95,1.4) .. (10.93,3.29)   ;
%Shape: Rectangle [id:dp7879675806429662] 
\draw   (172,1001) -- (272,1001) -- (272,1031) -- (172,1031) -- cycle ;
%Shape: Rectangle [id:dp39079505605519627] 
\draw   (381,941.5) -- (521,941.5) -- (521,971.5) -- (381,971.5) -- cycle ;
%Curve Lines [id:da8969334419941608] 
\draw    (272,896) .. controls (330.41,895.51) and (320.21,943.82) .. (376.28,945.76) ;
\draw [shift={(378,945.8)}, rotate = 180.99] [color={rgb, 255:red, 0; green, 0; blue, 0 }  ][line width=0.75]    (10.93,-3.29) .. controls (6.95,-1.4) and (3.31,-0.3) .. (0,0) .. controls (3.31,0.3) and (6.95,1.4) .. (10.93,3.29)   ;
%Curve Lines [id:da8672498381040883] 
\draw    (272,1016) .. controls (332.39,1015.51) and (321.23,966.79) .. (376.31,964.84) ;
\draw [shift={(378,964.8)}, rotate = 178.99] [color={rgb, 255:red, 0; green, 0; blue, 0 }  ][line width=0.75]    (10.93,-3.29) .. controls (6.95,-1.4) and (3.31,-0.3) .. (0,0) .. controls (3.31,0.3) and (6.95,1.4) .. (10.93,3.29)   ;
%Straight Lines [id:da13652105764168243] 
\draw    (521,957) -- (553,956.53) ;
\draw [shift={(555,956.5)}, rotate = 179.16] [color={rgb, 255:red, 0; green, 0; blue, 0 }  ][line width=0.75]    (10.93,-3.29) .. controls (6.95,-1.4) and (3.31,-0.3) .. (0,0) .. controls (3.31,0.3) and (6.95,1.4) .. (10.93,3.29)   ;
%Shape: Ellipse [id:dp49624443665106566] 
\draw   (558,957) .. controls (558,945.68) and (577.92,936.5) .. (602.5,936.5) .. controls (627.08,936.5) and (647,945.68) .. (647,957) .. controls (647,968.32) and (627.08,977.5) .. (602.5,977.5) .. controls (577.92,977.5) and (558,968.32) .. (558,957) -- cycle ;
%Shape: Ellipse [id:dp9339236734269698] 
\draw   (1,955) .. controls (1,946.72) and (12.19,940) .. (26,940) .. controls (39.81,940) and (51,946.72) .. (51,955) .. controls (51,963.28) and (39.81,970) .. (26,970) .. controls (12.19,970) and (1,963.28) .. (1,955) -- cycle ;
%Straight Lines [id:da48292758827035653] 
\draw    (51,955) -- (168.4,955.39) ;
\draw [shift={(170.4,955.4)}, rotate = 180.19] [color={rgb, 255:red, 0; green, 0; blue, 0 }  ][line width=0.75]    (10.93,-3.29) .. controls (6.95,-1.4) and (3.31,-0.3) .. (0,0) .. controls (3.31,0.3) and (6.95,1.4) .. (10.93,3.29)   ;
%Shape: Rectangle [id:dp40606700780506966] 
\draw   (172,941) -- (272,941) -- (272,971) -- (172,971) -- cycle ;
%Straight Lines [id:da35168746449256116] 
\draw    (272,956.2) -- (376,955.81) ;
\draw [shift={(378,955.8)}, rotate = 179.78] [color={rgb, 255:red, 0; green, 0; blue, 0 }  ][line width=0.75]    (10.93,-3.29) .. controls (6.95,-1.4) and (3.31,-0.3) .. (0,0) .. controls (3.31,0.3) and (6.95,1.4) .. (10.93,3.29)   ;

% Text Node
\draw (13,888.5) node [anchor=north west][inner sep=0.75pt]  [font=\large] [align=left] {$\displaystyle u_{LR}$};
% Text Node
\draw (179,888) node [anchor=north west][inner sep=0.75pt]  [font=\large] [align=left] {Branch Net};
% Text Node
\draw (21,1009.5) node [anchor=north west][inner sep=0.75pt]  [font=\large] [align=left] {$\displaystyle y$};
% Text Node
\draw (52,1020) node [anchor=north west][inner sep=0.75pt]  [font=\scriptsize,color={rgb, 255:red, 31; green, 19; blue, 254 }  ,opacity=1 ] [align=left] {$\displaystyle ( d\!+\!1)$};
% Text Node
\draw (183,1008) node [anchor=north west][inner sep=0.75pt]  [font=\large] [align=left] {Trunk Net};
% Text Node
\draw (403,948.5) node [anchor=north west][inner sep=0.75pt]  [font=\large] [align=left] {Inner Product};
% Text Node
\draw (276,1020) node [anchor=north west][inner sep=0.75pt]  [font=\scriptsize,color={rgb, 255:red, 31; green, 19; blue, 254 }  ,opacity=1 ] [align=left] {$\displaystyle ( K)$};
% Text Node
\draw (275,879) node [anchor=north west][inner sep=0.75pt]  [font=\scriptsize,color={rgb, 255:red, 31; green, 19; blue, 254 }  ,opacity=1 ] [align=left] {$\displaystyle ( K)$};
% Text Node
\draw (563,949) node [anchor=north west][inner sep=0.75pt]  [font=\large] [align=left] {$\displaystyle \mathcal{S}_{\theta }( u_{LR})( y)$};
% Text Node
\draw (21,949.5) node [anchor=north west][inner sep=0.75pt]  [font=\large] [align=left] {$\displaystyle x$};
% Text Node
\draw (182,948) node [anchor=north west][inner sep=0.75pt]  [font=\large] [align=left] {Sensor Net};
% Text Node
\draw (276,960) node [anchor=north west][inner sep=0.75pt]  [font=\scriptsize,color={rgb, 255:red, 31; green, 19; blue, 254 }  ,opacity=1 ] [align=left] {$\displaystyle ( K)$};
% Text Node
\draw (524,937) node [anchor=north west][inner sep=0.75pt]  [font=\scriptsize,color={rgb, 255:red, 31; green, 19; blue, 254 }  ,opacity=1 ] [align=left] {$\displaystyle ( 1)$};
% Text Node
\draw (52,877) node [anchor=north west][inner sep=0.75pt]  [font=\scriptsize,color={rgb, 255:red, 31; green, 19; blue, 254 }  ,opacity=1 ] [align=left] {$\displaystyle ( T,N_{1} \ ,...,N_{d})$};
% Text Node
\draw (52,960) node [anchor=north west][inner sep=0.75pt]  [font=\scriptsize,color={rgb, 255:red, 31; green, 19; blue, 254 }  ,opacity=1 ] [align=left] {$\displaystyle ( d\!+\!1,T ,N_{1} \ ,...,N_{d})$};

\end{tikzpicture}

}

\end{center}

%% file: diagrams/SR3Branch_Sequences.tex
\begin{center}

\resizebox{0.94\linewidth}{!}{

\tikzset{every picture/.style={line width=0.75pt}} %set default line width to 0.75pt        

\begin{tikzpicture}[x=0.75pt,y=0.75pt,yscale=-1,xscale=1]
%uncomment if require: \path (0,1580); %set diagram left start at 0, and has height of 1580

%Shape: Ellipse [id:dp19393129622664484] 
\draw   (0,1135) .. controls (0,1126.72) and (11.19,1120) .. (25,1120) .. controls (38.81,1120) and (50,1126.72) .. (50,1135) .. controls (50,1143.28) and (38.81,1150) .. (25,1150) .. controls (11.19,1150) and (0,1143.28) .. (0,1135) -- cycle ;
%Straight Lines [id:da8646990852039178] 
\draw    (50,1135) -- (167.4,1135.39) ;
\draw [shift={(169.4,1135.4)}, rotate = 180.19] [color={rgb, 255:red, 0; green, 0; blue, 0 }  ][line width=0.75]    (10.93,-3.29) .. controls (6.95,-1.4) and (3.31,-0.3) .. (0,0) .. controls (3.31,0.3) and (6.95,1.4) .. (10.93,3.29)   ;
%Shape: Rectangle [id:dp5141277831935935] 
\draw   (171,1121) -- (271,1121) -- (271,1151) -- (171,1151) -- cycle ;
%Shape: Ellipse [id:dp4487467017610567] 
\draw   (0,1255) .. controls (0,1246.72) and (11.19,1240) .. (25,1240) .. controls (38.81,1240) and (50,1246.72) .. (50,1255) .. controls (50,1263.28) and (38.81,1270) .. (25,1270) .. controls (11.19,1270) and (0,1263.28) .. (0,1255) -- cycle ;
%Straight Lines [id:da3709574563174092] 
\draw    (50,1256) -- (167.4,1256.39) ;
\draw [shift={(169.4,1256.4)}, rotate = 180.19] [color={rgb, 255:red, 0; green, 0; blue, 0 }  ][line width=0.75]    (10.93,-3.29) .. controls (6.95,-1.4) and (3.31,-0.3) .. (0,0) .. controls (3.31,0.3) and (6.95,1.4) .. (10.93,3.29)   ;
%Shape: Rectangle [id:dp24635113171604828] 
\draw   (171,1241) -- (271,1241) -- (271,1271) -- (171,1271) -- cycle ;
%Shape: Rectangle [id:dp7753117865016377] 
\draw   (380,1180.5) -- (479.8,1180.5) -- (479.8,1210.5) -- (380,1210.5) -- cycle ;
%Curve Lines [id:da07415710928309394] 
\draw    (271,1136) .. controls (329.41,1135.51) and (319.21,1183.82) .. (375.28,1185.76) ;
\draw [shift={(377,1185.8)}, rotate = 180.99] [color={rgb, 255:red, 0; green, 0; blue, 0 }  ][line width=0.75]    (10.93,-3.29) .. controls (6.95,-1.4) and (3.31,-0.3) .. (0,0) .. controls (3.31,0.3) and (6.95,1.4) .. (10.93,3.29)   ;
%Curve Lines [id:da12498733936514328] 
\draw    (271,1256) .. controls (331.39,1255.51) and (320.23,1206.79) .. (375.31,1204.84) ;
\draw [shift={(377,1204.8)}, rotate = 178.99] [color={rgb, 255:red, 0; green, 0; blue, 0 }  ][line width=0.75]    (10.93,-3.29) .. controls (6.95,-1.4) and (3.31,-0.3) .. (0,0) .. controls (3.31,0.3) and (6.95,1.4) .. (10.93,3.29)   ;
%Straight Lines [id:da9058713814665045] 
\draw    (480.4,1196.6) -- (527.4,1196.6) ;
\draw [shift={(529.4,1196.6)}, rotate = 180] [color={rgb, 255:red, 0; green, 0; blue, 0 }  ][line width=0.75]    (10.93,-3.29) .. controls (6.95,-1.4) and (3.31,-0.3) .. (0,0) .. controls (3.31,0.3) and (6.95,1.4) .. (10.93,3.29)   ;
%Shape: Ellipse [id:dp7982496266530326] 
\draw   (531,1197) .. controls (531,1185.68) and (550.92,1176.5) .. (575.5,1176.5) .. controls (600.08,1176.5) and (620,1185.68) .. (620,1197) .. controls (620,1208.32) and (600.08,1217.5) .. (575.5,1217.5) .. controls (550.92,1217.5) and (531,1208.32) .. (531,1197) -- cycle ;
%Shape: Ellipse [id:dp33952639140977015] 
\draw   (0,1195) .. controls (0,1186.72) and (11.19,1180) .. (25,1180) .. controls (38.81,1180) and (50,1186.72) .. (50,1195) .. controls (50,1203.28) and (38.81,1210) .. (25,1210) .. controls (11.19,1210) and (0,1203.28) .. (0,1195) -- cycle ;
%Straight Lines [id:da09531445098408042] 
\draw    (50,1195) -- (167.4,1195.39) ;
\draw [shift={(169.4,1195.4)}, rotate = 180.19] [color={rgb, 255:red, 0; green, 0; blue, 0 }  ][line width=0.75]    (10.93,-3.29) .. controls (6.95,-1.4) and (3.31,-0.3) .. (0,0) .. controls (3.31,0.3) and (6.95,1.4) .. (10.93,3.29)   ;
%Shape: Rectangle [id:dp3047957422869272] 
\draw   (171,1181) -- (271,1181) -- (271,1211) -- (171,1211) -- cycle ;
%Straight Lines [id:da09928647384255807] 
\draw    (271,1196.2) -- (375,1195.81) ;
\draw [shift={(377,1195.8)}, rotate = 179.78] [color={rgb, 255:red, 0; green, 0; blue, 0 }  ][line width=0.75]    (10.93,-3.29) .. controls (6.95,-1.4) and (3.31,-0.3) .. (0,0) .. controls (3.31,0.3) and (6.95,1.4) .. (10.93,3.29)   ;

% Text Node
\draw (12,1128.5) node [anchor=north west][inner sep=0.75pt]  [font=\large] [align=left] {$\displaystyle u_{LR}$};
% Text Node
\draw (178,1128) node [anchor=north west][inner sep=0.75pt]  [font=\large] [align=left] {Branch Net};
% Text Node
\draw (20,1249.5) node [anchor=north west][inner sep=0.75pt]  [font=\large] [align=left] {$\displaystyle y$};
% Text Node
\draw (51,1260) node [anchor=north west][inner sep=0.75pt]  [font=\scriptsize,color={rgb, 255:red, 31; green, 19; blue, 254 }  ,opacity=1 ] [align=left] {$\displaystyle ( d)$};
% Text Node
\draw (182,1248) node [anchor=north west][inner sep=0.75pt]  [font=\large] [align=left] {Trunk Net};
% Text Node
\draw (392,1187.5) node [anchor=north west][inner sep=0.75pt]  [font=\normalsize] [align=left] {$\displaystyle \Sigma _{j} B_{ij} S_{ij} T_{j}$};
% Text Node
\draw (275,1260) node [anchor=north west][inner sep=0.75pt]  [font=\scriptsize,color={rgb, 255:red, 31; green, 19; blue, 254 }  ,opacity=1 ] [align=left] {$\displaystyle ( K)$};
% Text Node
\draw (275,1117) node [anchor=north west][inner sep=0.75pt]  [font=\scriptsize,color={rgb, 255:red, 31; green, 19; blue, 254 }  ,opacity=1 ] [align=left] {$\displaystyle \left( T^{+} \! \!,K\right)$};
% Text Node
\draw (536,1189) node [anchor=north west][inner sep=0.75pt]  [font=\large] [align=left] {$\displaystyle \mathcal{S}_{\theta }( u_{LR})( y)$};
% Text Node
\draw (20,1189.5) node [anchor=north west][inner sep=0.75pt]  [font=\large] [align=left] {$\displaystyle x$};
% Text Node
\draw (181,1188) node [anchor=north west][inner sep=0.75pt]  [font=\large] [align=left] {Sensor Net};
% Text Node
\draw (275,1200) node [anchor=north west][inner sep=0.75pt]  [font=\scriptsize,color={rgb, 255:red, 31; green, 19; blue, 254 }  ,opacity=1 ] [align=left] {$\displaystyle \left( T^{+} \! \!,K\right)$};
% Text Node
\draw (484,1176) node [anchor=north west][inner sep=0.75pt]  [font=\scriptsize,color={rgb, 255:red, 31; green, 19; blue, 254 }  ,opacity=1 ] [align=left] {$\displaystyle \left( T^{+}\right)$};
% Text Node
\draw (51,1117) node [anchor=north west][inner sep=0.75pt]  [font=\scriptsize,color={rgb, 255:red, 31; green, 19; blue, 254 }  ,opacity=1 ] [align=left] {$\displaystyle ( T,N_{1} \ ,...,N_{d})$};
% Text Node
\draw (51,1200) node [anchor=north west][inner sep=0.75pt]  [font=\scriptsize,color={rgb, 255:red, 31; green, 19; blue, 254 }  ,opacity=1 ] [align=left] {$\displaystyle ( d \! + \! 1,T ,N_{1} \ ,...,N_{d})$};
% Text Node
\draw (345,1212) node [anchor=north west][inner sep=0.75pt]  [font=\scriptsize,color={rgb, 255:red, 0; green, 186; blue, 91 }  ,opacity=1 ] [align=left] {$\displaystyle [ T_{j}]$};
% Text Node
\draw (344,1164) node [anchor=north west][inner sep=0.75pt]  [font=\scriptsize,color={rgb, 255:red, 0; green, 186; blue, 91 }  ,opacity=1 ] [align=left] {$\displaystyle [ B_{ij}]$};
% Text Node
\draw (321,1181) node [anchor=north west][inner sep=0.75pt]  [font=\scriptsize,color={rgb, 255:red, 0; green, 186; blue, 91 }  ,opacity=1 ] [align=left] {$\displaystyle [ S_{ij}]$};

\end{tikzpicture}
}

\end{center}

%% file: diagrams/Architecture_Example.tex
\begin{center}

\resizebox{\linewidth}{!}{

\tikzset{every picture/.style={line width=0.75pt}} %set default line width to 0.75pt        

\begin{tikzpicture}[x=0.75pt,y=0.75pt,yscale=-1,xscale=1]
	%uncomment if require: \path (0,2566); %set diagram left start at 0, and has height of 2566
	
	%Shape: Ellipse [id:dp2062923714580902] 
	\draw   (1049.8,381.4) .. controls (1049.8,373.12) and (1060.99,366.4) .. (1074.8,366.4) .. controls (1088.61,366.4) and (1099.8,373.12) .. (1099.8,381.4) .. controls (1099.8,389.68) and (1088.61,396.4) .. (1074.8,396.4) .. controls (1060.99,396.4) and (1049.8,389.68) .. (1049.8,381.4) -- cycle ;
	%Straight Lines [id:da3282340992984547] 
	\draw    (1073,397) -- (1072.02,512.4) ;
	\draw [shift={(1072,514.4)}, rotate = 270.49] [color={rgb, 255:red, 0; green, 0; blue, 0 }  ][line width=0.75]    (10.93,-3.29) .. controls (6.95,-1.4) and (3.31,-0.3) .. (0,0) .. controls (3.31,0.3) and (6.95,1.4) .. (10.93,3.29)   ;
	%Shape: Rectangle [id:dp26953889698783895] 
	\draw   (1042.8,516.4) -- (1102.8,516.4) -- (1102.8,565.4) -- (1042.8,565.4) -- cycle ;
	%Shape: Rectangle [id:dp8185641984099452] 
	\draw   (945.8,726.4) -- (1027,726.4) -- (1027,752.2) -- (945.8,752.2) -- cycle ;
	%Straight Lines [id:da728710392862379] 
	\draw    (986.5,752.5) -- (987.14,774.4) ;
	\draw [shift={(987.2,776.4)}, rotate = 268.32] [color={rgb, 255:red, 0; green, 0; blue, 0 }  ][line width=0.75]    (10.93,-3.29) .. controls (6.95,-1.4) and (3.31,-0.3) .. (0,0) .. controls (3.31,0.3) and (6.95,1.4) .. (10.93,3.29)   ;
	%Rounded Rect [id:dp5918314069241775] 
	\draw  [color={rgb, 255:red, 22; green, 19; blue, 254 }  ,draw opacity=1 ][line width=1.5]  (1033.6,452.74) .. controls (1033.6,443.99) and (1040.69,436.9) .. (1049.44,436.9) -- (1096.96,436.9) .. controls (1105.71,436.9) and (1112.8,443.99) .. (1112.8,452.74) -- (1112.8,630.06) .. controls (1112.8,638.81) and (1105.71,645.9) .. (1096.96,645.9) -- (1049.44,645.9) .. controls (1040.69,645.9) and (1033.6,638.81) .. (1033.6,630.06) -- cycle ;
	%Shape: Ellipse [id:dp379729144355087] 
	\draw   (877,131.8) .. controls (877,123.52) and (888.19,116.8) .. (902,116.8) .. controls (915.81,116.8) and (927,123.52) .. (927,131.8) .. controls (927,140.08) and (915.81,146.8) .. (902,146.8) .. controls (888.19,146.8) and (877,140.08) .. (877,131.8) -- cycle ;
	%Straight Lines [id:da2888275644263667] 
	\draw    (902,146.8) -- (902.19,199) ;
	\draw [shift={(902.2,201)}, rotate = 269.79] [color={rgb, 255:red, 0; green, 0; blue, 0 }  ][line width=0.75]    (10.93,-3.29) .. controls (6.95,-1.4) and (3.31,-0.3) .. (0,0) .. controls (3.31,0.3) and (6.95,1.4) .. (10.93,3.29)   ;
	%Shape: Rectangle [id:dp08022132343468158] 
	\draw   (832,202.8) -- (972,202.8) -- (972,232.8) -- (832,232.8) -- cycle ;
	%Straight Lines [id:da9242716663953259] 
	\draw    (902,232.8) -- (902,265.8) ;
	\draw [shift={(902,267.8)}, rotate = 270] [color={rgb, 255:red, 0; green, 0; blue, 0 }  ][line width=0.75]    (10.93,-3.29) .. controls (6.95,-1.4) and (3.31,-0.3) .. (0,0) .. controls (3.31,0.3) and (6.95,1.4) .. (10.93,3.29)   ;
	%Shape: Rectangle [id:dp06431399676318228] 
	\draw   (832,272.3) -- (971,272.3) -- (971,519.3) -- (832,519.3) -- cycle ;
	%Shape: Rectangle [id:dp25771018726440587] 
	\draw   (880,275.8) -- (958,275.8) -- (958,322.3) -- (880,322.3) -- cycle ;
	%Straight Lines [id:da8213939088334248] 
	\draw    (902,321.8) -- (902,340.8) ;
	\draw [shift={(902,342.8)}, rotate = 270] [color={rgb, 255:red, 0; green, 0; blue, 0 }  ][line width=0.75]    (10.93,-3.29) .. controls (6.95,-1.4) and (3.31,-0.3) .. (0,0) .. controls (3.31,0.3) and (6.95,1.4) .. (10.93,3.29)   ;
	%Shape: Rectangle [id:dp06525010012392274] 
	\draw   (878,473.8) -- (958,473.8) -- (958,513.8) -- (878,513.8) -- cycle ;
	%Straight Lines [id:da643423485696784] 
	\draw    (902,519.8) -- (902,558.3) ;
	\draw [shift={(902,560.3)}, rotate = 270] [color={rgb, 255:red, 0; green, 0; blue, 0 }  ][line width=0.75]    (10.93,-3.29) .. controls (6.95,-1.4) and (3.31,-0.3) .. (0,0) .. controls (3.31,0.3) and (6.95,1.4) .. (10.93,3.29)   ;
	%Shape: Rectangle [id:dp9991697425913446] 
	\draw   (862,562.8) -- (952,562.8) -- (952,592.8) -- (862,592.8) -- cycle ;
	%Straight Lines [id:da802370737348546] 
	\draw    (902,593.8) -- (902,629.8) ;
	\draw [shift={(902,631.8)}, rotate = 270] [color={rgb, 255:red, 0; green, 0; blue, 0 }  ][line width=0.75]    (10.93,-3.29) .. controls (6.95,-1.4) and (3.31,-0.3) .. (0,0) .. controls (3.31,0.3) and (6.95,1.4) .. (10.93,3.29)   ;
	%Shape: Rectangle [id:dp16959286937313967] 
	\draw   (873,633.8) -- (933,633.8) -- (933,663.8) -- (873,663.8) -- cycle ;
	%Rounded Rect [id:dp4360199615424065] 
	\draw  [color={rgb, 255:red, 22; green, 19; blue, 254 }  ,draw opacity=1 ][line width=1.5]  (821.6,211.21) .. controls (821.6,192.32) and (836.92,177) .. (855.81,177) -- (958.45,177) .. controls (977.35,177) and (992.67,192.32) .. (992.67,211.21) -- (992.67,661.09) .. controls (992.67,679.98) and (977.35,695.3) .. (958.45,695.3) -- (855.81,695.3) .. controls (836.92,695.3) and (821.6,679.98) .. (821.6,661.09) -- cycle ;
	%Straight Lines [id:da24683564704501526] 
	\draw    (902,379.8) -- (902,398.8) ;
	\draw [shift={(902,400.8)}, rotate = 270] [color={rgb, 255:red, 0; green, 0; blue, 0 }  ][line width=0.75]    (10.93,-3.29) .. controls (6.95,-1.4) and (3.31,-0.3) .. (0,0) .. controls (3.31,0.3) and (6.95,1.4) .. (10.93,3.29)   ;
	%Shape: Rectangle [id:dp6789770114680227] 
	\draw   (881,403.8) -- (958,403.8) -- (958,450.3) -- (881,450.3) -- cycle ;
	%Straight Lines [id:da39825188697873926] 
	\draw    (902,449.8) -- (902,468.8) ;
	\draw [shift={(902,470.8)}, rotate = 270] [color={rgb, 255:red, 0; green, 0; blue, 0 }  ][line width=0.75]    (10.93,-3.29) .. controls (6.95,-1.4) and (3.31,-0.3) .. (0,0) .. controls (3.31,0.3) and (6.95,1.4) .. (10.93,3.29)   ;
	%Shape: Free Drawing [id:dp9947408691614927] 
	\draw  [line width=3] [line join = round][line cap = round] (902,368.8) .. controls (902,368.8) and (902,368.8) .. (902,368.8) ;
	%Shape: Free Drawing [id:dp8058436313277377] 
	\draw  [line width=3] [line join = round][line cap = round] (902,358.8) .. controls (902,358.8) and (902,358.8) .. (902,358.8) ;
	%Shape: Free Drawing [id:dp0074434508776084485] 
	\draw  [line width=3] [line join = round][line cap = round] (902,350.8) .. controls (902,350.8) and (902,350.8) .. (902,350.8) ;
	%Curve Lines [id:da5545073977548522] 
	\draw    (902.4,664) .. controls (900.43,706.36) and (930.87,694.96) .. (955.48,721.75) ;
	\draw [shift={(956.6,723)}, rotate = 228.87] [color={rgb, 255:red, 0; green, 0; blue, 0 }  ][line width=0.75]    (10.93,-3.29) .. controls (6.95,-1.4) and (3.31,-0.3) .. (0,0) .. controls (3.31,0.3) and (6.95,1.4) .. (10.93,3.29)   ;
	%Curve Lines [id:da7286112555824726] 
	\draw    (1072,565.4) .. controls (1068.77,702.88) and (1038.55,663.81) .. (1017.24,721.23) ;
	\draw [shift={(1016.6,723)}, rotate = 289.57] [color={rgb, 255:red, 0; green, 0; blue, 0 }  ][line width=0.75]    (10.93,-3.29) .. controls (6.95,-1.4) and (3.31,-0.3) .. (0,0) .. controls (3.31,0.3) and (6.95,1.4) .. (10.93,3.29)   ;
	%Shape: Ellipse [id:dp47475691089166605] 
	\draw   (939,804.58) .. controls (939,790.18) and (960.27,778.5) .. (986.5,778.5) .. controls (1012.73,778.5) and (1034,790.18) .. (1034,804.58) .. controls (1034,818.99) and (1012.73,830.67) .. (986.5,830.67) .. controls (960.27,830.67) and (939,818.99) .. (939,804.58) -- cycle ;
	%Shape: Ellipse [id:dp46650433050692364] 
	\draw   (1551.8,382.4) .. controls (1551.8,374.12) and (1562.99,367.4) .. (1576.8,367.4) .. controls (1590.61,367.4) and (1601.8,374.12) .. (1601.8,382.4) .. controls (1601.8,390.68) and (1590.61,397.4) .. (1576.8,397.4) .. controls (1562.99,397.4) and (1551.8,390.68) .. (1551.8,382.4) -- cycle ;
	%Straight Lines [id:da4153751973509747] 
	\draw    (1575,398) -- (1574.02,513.4) ;
	\draw [shift={(1574,515.4)}, rotate = 270.49] [color={rgb, 255:red, 0; green, 0; blue, 0 }  ][line width=0.75]    (10.93,-3.29) .. controls (6.95,-1.4) and (3.31,-0.3) .. (0,0) .. controls (3.31,0.3) and (6.95,1.4) .. (10.93,3.29)   ;
	%Shape: Rectangle [id:dp6241658343707215] 
	\draw   (1544.8,517.4) -- (1604.8,517.4) -- (1604.8,566.4) -- (1544.8,566.4) -- cycle ;
	%Rounded Rect [id:dp11485157991033579] 
	\draw  [color={rgb, 255:red, 22; green, 19; blue, 254 }  ,draw opacity=1 ][line width=1.5]  (1535.6,453.74) .. controls (1535.6,444.99) and (1542.69,437.9) .. (1551.44,437.9) -- (1598.96,437.9) .. controls (1607.71,437.9) and (1614.8,444.99) .. (1614.8,453.74) -- (1614.8,631.06) .. controls (1614.8,639.81) and (1607.71,646.9) .. (1598.96,646.9) -- (1551.44,646.9) .. controls (1542.69,646.9) and (1535.6,639.81) .. (1535.6,631.06) -- cycle ;
	%Shape: Rectangle [id:dp15602177271678652] 
	\draw   (1334,273.3) -- (1473,273.3) -- (1473,520.3) -- (1334,520.3) -- cycle ;
	%Shape: Rectangle [id:dp10948074776436068] 
	\draw   (1382,276.8) -- (1460,276.8) -- (1460,323.3) -- (1382,323.3) -- cycle ;
	%Straight Lines [id:da5586773823764615] 
	\draw    (1404,322.8) -- (1404,341.8) ;
	\draw [shift={(1404,343.8)}, rotate = 270] [color={rgb, 255:red, 0; green, 0; blue, 0 }  ][line width=0.75]    (10.93,-3.29) .. controls (6.95,-1.4) and (3.31,-0.3) .. (0,0) .. controls (3.31,0.3) and (6.95,1.4) .. (10.93,3.29)   ;
	%Shape: Rectangle [id:dp609396345452712] 
	\draw   (1380,474.8) -- (1460,474.8) -- (1460,514.8) -- (1380,514.8) -- cycle ;
	%Straight Lines [id:da9808595069290464] 
	\draw    (1404,520.8) -- (1404,559.3) ;
	\draw [shift={(1404,561.3)}, rotate = 270] [color={rgb, 255:red, 0; green, 0; blue, 0 }  ][line width=0.75]    (10.93,-3.29) .. controls (6.95,-1.4) and (3.31,-0.3) .. (0,0) .. controls (3.31,0.3) and (6.95,1.4) .. (10.93,3.29)   ;
	%Shape: Rectangle [id:dp5611243139666844] 
	\draw   (1364,563.8) -- (1454,563.8) -- (1454,593.8) -- (1364,593.8) -- cycle ;
	%Straight Lines [id:da7883151459745008] 
	\draw    (1404,594.8) -- (1404,630.8) ;
	\draw [shift={(1404,632.8)}, rotate = 270] [color={rgb, 255:red, 0; green, 0; blue, 0 }  ][line width=0.75]    (10.93,-3.29) .. controls (6.95,-1.4) and (3.31,-0.3) .. (0,0) .. controls (3.31,0.3) and (6.95,1.4) .. (10.93,3.29)   ;
	%Shape: Rectangle [id:dp0029192300673992477] 
	\draw   (1375,634.8) -- (1435,634.8) -- (1435,664.8) -- (1375,664.8) -- cycle ;
	%Rounded Rect [id:dp6051270158962716] 
	\draw  [color={rgb, 255:red, 22; green, 19; blue, 254 }  ,draw opacity=1 ][line width=1.5]  (1323.6,279.21) .. controls (1323.6,260.32) and (1338.92,245) .. (1357.81,245) -- (1460.45,245) .. controls (1479.35,245) and (1494.67,260.32) .. (1494.67,279.21) -- (1494.67,662.09) .. controls (1494.67,680.98) and (1479.35,696.3) .. (1460.45,696.3) -- (1357.81,696.3) .. controls (1338.92,696.3) and (1323.6,680.98) .. (1323.6,662.09) -- cycle ;
	%Straight Lines [id:da16827341228315795] 
	\draw    (1404,380.8) -- (1404,399.8) ;
	\draw [shift={(1404,401.8)}, rotate = 270] [color={rgb, 255:red, 0; green, 0; blue, 0 }  ][line width=0.75]    (10.93,-3.29) .. controls (6.95,-1.4) and (3.31,-0.3) .. (0,0) .. controls (3.31,0.3) and (6.95,1.4) .. (10.93,3.29)   ;
	%Shape: Rectangle [id:dp7687128135883785] 
	\draw   (1383,404.8) -- (1460,404.8) -- (1460,451.3) -- (1383,451.3) -- cycle ;
	%Straight Lines [id:da38969964352605246] 
	\draw    (1404,450.8) -- (1404,469.8) ;
	\draw [shift={(1404,471.8)}, rotate = 270] [color={rgb, 255:red, 0; green, 0; blue, 0 }  ][line width=0.75]    (10.93,-3.29) .. controls (6.95,-1.4) and (3.31,-0.3) .. (0,0) .. controls (3.31,0.3) and (6.95,1.4) .. (10.93,3.29)   ;
	%Shape: Free Drawing [id:dp667606956269299] 
	\draw  [line width=3] [line join = round][line cap = round] (1404,369.8) .. controls (1404,369.8) and (1404,369.8) .. (1404,369.8) ;
	%Shape: Free Drawing [id:dp7048956739590668] 
	\draw  [line width=3] [line join = round][line cap = round] (1404,359.8) .. controls (1404,359.8) and (1404,359.8) .. (1404,359.8) ;
	%Shape: Free Drawing [id:dp6292504964126306] 
	\draw  [line width=3] [line join = round][line cap = round] (1404,351.8) .. controls (1404,351.8) and (1404,351.8) .. (1404,351.8) ;
	%Shape: Ellipse [id:dp03951688375224105] 
	\draw   (1378.4,202) .. controls (1378.4,193.72) and (1389.59,187) .. (1403.4,187) .. controls (1417.21,187) and (1428.4,193.72) .. (1428.4,202) .. controls (1428.4,210.28) and (1417.21,217) .. (1403.4,217) .. controls (1389.59,217) and (1378.4,210.28) .. (1378.4,202) -- cycle ;
	%Straight Lines [id:da5839017905984589] 
	\draw    (1403.4,217) -- (1403.21,269) ;
	\draw [shift={(1403.2,271)}, rotate = 270.21] [color={rgb, 255:red, 0; green, 0; blue, 0 }  ][line width=0.75]    (10.93,-3.29) .. controls (6.95,-1.4) and (3.31,-0.3) .. (0,0) .. controls (3.31,0.3) and (6.95,1.4) .. (10.93,3.29)   ;
	%Straight Lines [id:da7374470513565281] 
	\draw [line width=2.25]    (1246.4,142.6) -- (1247,900) ;
	%Shape: Ellipse [id:dp013094486244652481] 
	\draw   (1441,804.58) .. controls (1441,790.18) and (1462.27,778.5) .. (1488.5,778.5) .. controls (1514.73,778.5) and (1536,790.18) .. (1536,804.58) .. controls (1536,818.99) and (1514.73,830.67) .. (1488.5,830.67) .. controls (1462.27,830.67) and (1441,818.99) .. (1441,804.58) -- cycle ;
	%Shape: Rectangle [id:dp4720066325021117] 
	\draw   (1447.8,727.4) -- (1529,727.4) -- (1529,753.2) -- (1447.8,753.2) -- cycle ;
	%Straight Lines [id:da7616806655194175] 
	\draw    (1488.5,753.5) -- (1489.14,775.4) ;
	\draw [shift={(1489.2,777.4)}, rotate = 268.32] [color={rgb, 255:red, 0; green, 0; blue, 0 }  ][line width=0.75]    (10.93,-3.29) .. controls (6.95,-1.4) and (3.31,-0.3) .. (0,0) .. controls (3.31,0.3) and (6.95,1.4) .. (10.93,3.29)   ;
	%Curve Lines [id:da46474645138583903] 
	\draw    (1404.4,665) .. controls (1402.43,707.36) and (1432.87,695.96) .. (1457.48,722.75) ;
	\draw [shift={(1458.6,724)}, rotate = 228.87] [color={rgb, 255:red, 0; green, 0; blue, 0 }  ][line width=0.75]    (10.93,-3.29) .. controls (6.95,-1.4) and (3.31,-0.3) .. (0,0) .. controls (3.31,0.3) and (6.95,1.4) .. (10.93,3.29)   ;
	%Curve Lines [id:da773099594716075] 
	\draw    (1574,566.4) .. controls (1570.77,703.88) and (1540.55,664.81) .. (1519.24,722.23) ;
	\draw [shift={(1518.6,724)}, rotate = 289.57] [color={rgb, 255:red, 0; green, 0; blue, 0 }  ][line width=0.75]    (10.93,-3.29) .. controls (6.95,-1.4) and (3.31,-0.3) .. (0,0) .. controls (3.31,0.3) and (6.95,1.4) .. (10.93,3.29)   ;
	%Shape: Ellipse [id:dp4596435649012236] 
	\draw   (1162.8,1150.53) .. controls (1162.8,1142.25) and (1173.99,1135.53) .. (1187.8,1135.53) .. controls (1201.61,1135.53) and (1212.8,1142.25) .. (1212.8,1150.53) .. controls (1212.8,1158.82) and (1201.61,1165.53) .. (1187.8,1165.53) .. controls (1173.99,1165.53) and (1162.8,1158.82) .. (1162.8,1150.53) -- cycle ;
	%Straight Lines [id:da7459955627650516] 
	\draw    (1212.6,1150.4) -- (1298.8,1150.79) ;
	\draw [shift={(1300.8,1150.8)}, rotate = 180.26] [color={rgb, 255:red, 0; green, 0; blue, 0 }  ][line width=0.75]    (10.93,-3.29) .. controls (6.95,-1.4) and (3.31,-0.3) .. (0,0) .. controls (3.31,0.3) and (6.95,1.4) .. (10.93,3.29)   ;
	%Shape: Rectangle [id:dp623537311114031] 
	\draw   (1513.8,1044.2) -- (1602.6,1044.2) -- (1602.6,1079.2) -- (1513.8,1079.2) -- cycle ;
	%Straight Lines [id:da30508648018191686] 
	\draw    (1602.5,1060.63) -- (1629.6,1060.6) ;
	\draw [shift={(1631.6,1060.6)}, rotate = 179.93] [color={rgb, 255:red, 0; green, 0; blue, 0 }  ][line width=0.75]    (10.93,-3.29) .. controls (6.95,-1.4) and (3.31,-0.3) .. (0,0) .. controls (3.31,0.3) and (6.95,1.4) .. (10.93,3.29)   ;
	%Rounded Rect [id:dp05050681407086133] 
	\draw  [color={rgb, 255:red, 22; green, 19; blue, 254 }  ,draw opacity=1 ][line width=1.5]  (1253.6,1134.2) .. controls (1253.6,1127.9) and (1258.7,1122.8) .. (1265,1122.8) -- (1422.4,1122.8) .. controls (1428.7,1122.8) and (1433.8,1127.9) .. (1433.8,1134.2) -- (1433.8,1168.4) .. controls (1433.8,1174.7) and (1428.7,1179.8) .. (1422.4,1179.8) -- (1265,1179.8) .. controls (1258.7,1179.8) and (1253.6,1174.7) .. (1253.6,1168.4) -- cycle ;
	%Shape: Ellipse [id:dp37140740355341406] 
	\draw   (733,971.93) .. controls (733,963.65) and (744.19,956.93) .. (758,956.93) .. controls (771.81,956.93) and (783,963.65) .. (783,971.93) .. controls (783,980.22) and (771.81,986.93) .. (758,986.93) .. controls (744.19,986.93) and (733,980.22) .. (733,971.93) -- cycle ;
	%Straight Lines [id:da18675690712899895] 
	\draw    (783,971.93) -- (889.6,972.2) ;
	\draw [shift={(891.6,972.2)}, rotate = 180.14] [color={rgb, 255:red, 0; green, 0; blue, 0 }  ][line width=0.75]    (10.93,-3.29) .. controls (6.95,-1.4) and (3.31,-0.3) .. (0,0) .. controls (3.31,0.3) and (6.95,1.4) .. (10.93,3.29)   ;
	%Shape: Rectangle [id:dp5607888695451588] 
	\draw   (894,955.93) -- (1022.8,955.93) -- (1022.8,985.93) -- (894,985.93) -- cycle ;
	%Straight Lines [id:da8173211149318425] 
	\draw    (1022.6,971.2) -- (1050.8,971.01) ;
	\draw [shift={(1052.8,971)}, rotate = 179.62] [color={rgb, 255:red, 0; green, 0; blue, 0 }  ][line width=0.75]    (10.93,-3.29) .. controls (6.95,-1.4) and (3.31,-0.3) .. (0,0) .. controls (3.31,0.3) and (6.95,1.4) .. (10.93,3.29)   ;
	%Rounded Rect [id:dp37418211141309277] 
	\draw  [color={rgb, 255:red, 22; green, 19; blue, 254 }  ,draw opacity=1 ][line width=1.5]  (866.63,953.4) .. controls (866.63,946.66) and (872.1,941.2) .. (878.83,941.2) -- (1421.6,941.2) .. controls (1428.34,941.2) and (1433.8,946.66) .. (1433.8,953.4) -- (1433.8,990) .. controls (1433.8,996.74) and (1428.34,1002.2) .. (1421.6,1002.2) -- (878.83,1002.2) .. controls (872.1,1002.2) and (866.63,996.74) .. (866.63,990) -- cycle ;
	%Shape: Ellipse [id:dp7368770817570345] 
	\draw   (1633,1060.72) .. controls (1633,1046.31) and (1654.27,1034.63) .. (1680.5,1034.63) .. controls (1706.73,1034.63) and (1728,1046.31) .. (1728,1060.72) .. controls (1728,1075.12) and (1706.73,1086.8) .. (1680.5,1086.8) .. controls (1654.27,1086.8) and (1633,1075.12) .. (1633,1060.72) -- cycle ;
	%Shape: Rectangle [id:dp11313100315647362] 
	\draw   (1134,950.93) -- (1214,950.93) -- (1214,990.93) -- (1134,990.93) -- cycle ;
	%Shape: Rectangle [id:dp972974668333418] 
	\draw   (1053.6,955.93) -- (1103.6,955.93) -- (1103.6,985.93) -- (1053.6,985.93) -- cycle ;
	%Straight Lines [id:da3379005705102667] 
	\draw    (1103.6,971.6) -- (1130.6,971.6) ;
	\draw [shift={(1132.6,971.6)}, rotate = 180] [color={rgb, 255:red, 0; green, 0; blue, 0 }  ][line width=0.75]    (10.93,-3.29) .. controls (6.95,-1.4) and (3.31,-0.3) .. (0,0) .. controls (3.31,0.3) and (6.95,1.4) .. (10.93,3.29)   ;
	%Straight Lines [id:da2517646896101644] 
	\draw    (1214.6,972.6) -- (1240.6,972.97) ;
	\draw [shift={(1242.6,973)}, rotate = 180.82] [color={rgb, 255:red, 0; green, 0; blue, 0 }  ][line width=0.75]    (10.93,-3.29) .. controls (6.95,-1.4) and (3.31,-0.3) .. (0,0) .. controls (3.31,0.3) and (6.95,1.4) .. (10.93,3.29)   ;
	%Shape: Rectangle [id:dp02821692770183626] 
	\draw   (1243,955.93) -- (1303.6,955.93) -- (1303.6,985.93) -- (1243,985.93) -- cycle ;
	%Straight Lines [id:da23858138106894966] 
	\draw    (1303.6,971.6) -- (1330.6,971.6) ;
	\draw [shift={(1332.6,971.6)}, rotate = 180] [color={rgb, 255:red, 0; green, 0; blue, 0 }  ][line width=0.75]    (10.93,-3.29) .. controls (6.95,-1.4) and (3.31,-0.3) .. (0,0) .. controls (3.31,0.3) and (6.95,1.4) .. (10.93,3.29)   ;
	%Shape: Rectangle [id:dp24982004276957337] 
	\draw   (1333.6,955.6) -- (1382.6,955.6) -- (1382.6,985.6) -- (1333.6,985.6) -- cycle ;
	%Straight Lines [id:da5076860747966678] 
	\draw    (783,1062.2) -- (890.4,1062.46) ;
	\draw [shift={(892.4,1062.47)}, rotate = 180.14] [color={rgb, 255:red, 0; green, 0; blue, 0 }  ][line width=0.75]    (10.93,-3.29) .. controls (6.95,-1.4) and (3.31,-0.3) .. (0,0) .. controls (3.31,0.3) and (6.95,1.4) .. (10.93,3.29)   ;
	%Shape: Rectangle [id:dp5391324080660576] 
	\draw   (894,1045.93) -- (1022.8,1045.93) -- (1022.8,1075.93) -- (894,1075.93) -- cycle ;
	%Straight Lines [id:da9661749585722679] 
	\draw    (1022.6,1061.2) -- (1050.8,1061.01) ;
	\draw [shift={(1052.8,1061)}, rotate = 179.62] [color={rgb, 255:red, 0; green, 0; blue, 0 }  ][line width=0.75]    (10.93,-3.29) .. controls (6.95,-1.4) and (3.31,-0.3) .. (0,0) .. controls (3.31,0.3) and (6.95,1.4) .. (10.93,3.29)   ;
	%Rounded Rect [id:dp11968653087445658] 
	\draw  [color={rgb, 255:red, 22; green, 19; blue, 254 }  ,draw opacity=1 ][line width=1.5]  (866.63,1043.4) .. controls (866.63,1036.66) and (872.1,1031.2) .. (878.83,1031.2) -- (1421.6,1031.2) .. controls (1428.34,1031.2) and (1433.8,1036.66) .. (1433.8,1043.4) -- (1433.8,1080) .. controls (1433.8,1086.74) and (1428.34,1092.2) .. (1421.6,1092.2) -- (878.83,1092.2) .. controls (872.1,1092.2) and (866.63,1086.74) .. (866.63,1080) -- cycle ;
	%Shape: Rectangle [id:dp747540275875987] 
	\draw   (1134,1040.93) -- (1214,1040.93) -- (1214,1080.93) -- (1134,1080.93) -- cycle ;
	%Shape: Rectangle [id:dp26234816146533624] 
	\draw   (1053.6,1045.93) -- (1103.6,1045.93) -- (1103.6,1075.93) -- (1053.6,1075.93) -- cycle ;
	%Straight Lines [id:da8436830246574876] 
	\draw    (1103.6,1061.6) -- (1130.6,1061.6) ;
	\draw [shift={(1132.6,1061.6)}, rotate = 180] [color={rgb, 255:red, 0; green, 0; blue, 0 }  ][line width=0.75]    (10.93,-3.29) .. controls (6.95,-1.4) and (3.31,-0.3) .. (0,0) .. controls (3.31,0.3) and (6.95,1.4) .. (10.93,3.29)   ;
	%Straight Lines [id:da2681330221070397] 
	\draw    (1214.6,1062.6) -- (1240.6,1062.97) ;
	\draw [shift={(1242.6,1063)}, rotate = 180.82] [color={rgb, 255:red, 0; green, 0; blue, 0 }  ][line width=0.75]    (10.93,-3.29) .. controls (6.95,-1.4) and (3.31,-0.3) .. (0,0) .. controls (3.31,0.3) and (6.95,1.4) .. (10.93,3.29)   ;
	%Shape: Rectangle [id:dp9523613902610786] 
	\draw   (1243,1045.93) -- (1303.6,1045.93) -- (1303.6,1075.93) -- (1243,1075.93) -- cycle ;
	%Straight Lines [id:da7680205646703089] 
	\draw    (1303.6,1061.6) -- (1330.6,1061.6) ;
	\draw [shift={(1332.6,1061.6)}, rotate = 180] [color={rgb, 255:red, 0; green, 0; blue, 0 }  ][line width=0.75]    (10.93,-3.29) .. controls (6.95,-1.4) and (3.31,-0.3) .. (0,0) .. controls (3.31,0.3) and (6.95,1.4) .. (10.93,3.29)   ;
	%Shape: Rectangle [id:dp18663669310366027] 
	\draw   (1333.6,1045.6) -- (1382.6,1045.6) -- (1382.6,1075.6) -- (1333.6,1075.6) -- cycle ;
	%Shape: Ellipse [id:dp004907779089255104] 
	\draw   (733,1063.2) .. controls (733,1054.92) and (744.19,1048.2) .. (758,1048.2) .. controls (771.81,1048.2) and (783,1054.92) .. (783,1063.2) .. controls (783,1071.48) and (771.81,1078.2) .. (758,1078.2) .. controls (744.19,1078.2) and (733,1071.48) .. (733,1063.2) -- cycle ;
	%Shape: Rectangle [id:dp3390619697078696] 
	\draw   (1302.6,1136.3) -- (1392.6,1136.3) -- (1392.6,1166.3) -- (1302.6,1166.3) -- cycle ;
	%Curve Lines [id:da4953067158082647] 
	\draw    (1393,1151.2) .. controls (1473,1151.8) and (1435.78,1083.09) .. (1510.27,1070.98) ;
	\draw [shift={(1511.4,1070.8)}, rotate = 171.32] [color={rgb, 255:red, 0; green, 0; blue, 0 }  ][line width=0.75]    (10.93,-3.29) .. controls (6.95,-1.4) and (3.31,-0.3) .. (0,0) .. controls (3.31,0.3) and (6.95,1.4) .. (10.93,3.29)   ;
	%Straight Lines [id:da6847760351613372] 
	\draw    (1383.6,1062.4) -- (1509.4,1062.79) ;
	\draw [shift={(1511.4,1062.8)}, rotate = 180.18] [color={rgb, 255:red, 0; green, 0; blue, 0 }  ][line width=0.75]    (10.93,-3.29) .. controls (6.95,-1.4) and (3.31,-0.3) .. (0,0) .. controls (3.31,0.3) and (6.95,1.4) .. (10.93,3.29)   ;
	%Curve Lines [id:da1951803383910915] 
	\draw    (1383,972.2) .. controls (1462,971.8) and (1434.68,1039.12) .. (1510.25,1054.57) ;
	\draw [shift={(1511.4,1054.8)}, rotate = 191.02] [color={rgb, 255:red, 0; green, 0; blue, 0 }  ][line width=0.75]    (10.93,-3.29) .. controls (6.95,-1.4) and (3.31,-0.3) .. (0,0) .. controls (3.31,0.3) and (6.95,1.4) .. (10.93,3.29)   ;
	%Straight Lines [id:da2944989110544123] 
	\draw [line width=2.25]    (1680,899.6) -- (762,902) ;
	
	% Text Node
	\draw (1068.8,377.4) node [anchor=north west][inner sep=0.75pt]  [font=\large] [align=left] {$\displaystyle y$};
	% Text Node
	\draw (1079.8,400.4) node [anchor=north west][inner sep=0.75pt]  [font=\scriptsize,color={rgb, 255:red, 31; green, 19; blue, 254 }  ,opacity=1 ] [align=left] {$\displaystyle ( d)$};
	% Text Node
	\draw (1053.8,534.4) node [anchor=north west][inner sep=0.75pt]  [font=\large] [align=left] {MLP};
	% Text Node
	\draw (1076.8,568.4) node [anchor=north west][inner sep=0.75pt]  [font=\scriptsize,color={rgb, 255:red, 31; green, 19; blue, 254 }  ,opacity=1 ] [align=left] {$\displaystyle ( K)$};
	% Text Node
	\draw (958.07,730.93) node [anchor=north west][inner sep=0.75pt]  [font=\normalsize] [align=left] {$\displaystyle \Sigma _{j} B_{ij} T_{j}$};
	% Text Node
	\draw (991.5,754.5) node [anchor=north west][inner sep=0.75pt]  [font=\scriptsize,color={rgb, 255:red, 31; green, 19; blue, 254 }  ,opacity=1 ] [align=left] {$\displaystyle \left( T^{+}\right)$};
	% Text Node
	\draw (1011.89,587.89) node [anchor=north west][inner sep=0.75pt]  [rotate=-270.02] [align=left] {\textbf{{\Large \textcolor[rgb]{0.07,0.09,1}{TrunkNet}}}};
	% Text Node
	\draw (887,126.8) node [anchor=north west][inner sep=0.75pt]  [font=\large] [align=left] {$\displaystyle u_{LR}$};
	% Text Node
	\draw (906,152.8) node [anchor=north west][inner sep=0.75pt]  [font=\scriptsize,color={rgb, 255:red, 31; green, 19; blue, 254 }  ,opacity=1 ] [align=left] {$\displaystyle ( T,N_{1} \ ,...,N_{d})$};
	% Text Node
	\draw (846,209.8) node [anchor=north west][inner sep=0.75pt]  [font=\large] [align=left] {Time Upscaling};
	% Text Node
	\draw (907,237.8) node [anchor=north west][inner sep=0.75pt]  [font=\scriptsize,color={rgb, 255:red, 31; green, 19; blue, 254 }  ,opacity=1 ] [align=left] {$\displaystyle \left( T^{+} \! \! ,N_{1} \ ,...,N_{d}\right)$};
	% Text Node
	\draw (884,270.8) node [anchor=north west][inner sep=0.75pt]  [font=\small] [align=left] {\begin{minipage}[lt]{51.2pt}\setlength\topsep{0pt}
			\begin{center}
				Convolution\\BatchNorm\\Activation
			\end{center}
			
	\end{minipage}};
	% Text Node
	\draw (884,478.8) node [anchor=north west][inner sep=0.75pt]  [font=\normalsize] [align=left] {\begin{minipage}[lt]{47.52pt}\setlength\topsep{0pt}
			\begin{center}
				Spatial\\Flattening
			\end{center}
			
	\end{minipage}};
	% Text Node
	\draw (836,276.3) node [anchor=north west][inner sep=0.75pt]   [align=left] {{\large CNN}};
	% Text Node
	\draw (880,571.8) node [anchor=north west][inner sep=0.75pt]  [font=\large] [align=left] {LSTM};
	% Text Node
	\draw (908,597.8) node [anchor=north west][inner sep=0.75pt]  [font=\scriptsize,color={rgb, 255:red, 31; green, 19; blue, 254 }  ,opacity=1 ] [align=left] {$\displaystyle \left( T^{+} \! \! ,K_{2}\right)$};
	% Text Node
	\draw (883,641.8) node [anchor=north west][inner sep=0.75pt]  [font=\large] [align=left] {MLP};
	% Text Node
	\draw (908,667.8) node [anchor=north west][inner sep=0.75pt]  [font=\scriptsize,color={rgb, 255:red, 31; green, 19; blue, 254 }  ,opacity=1 ] [align=left] {$\displaystyle \left( T^{+} \! \! ,K\right)$};
	% Text Node
	\draw (907,524.8) node [anchor=north west][inner sep=0.75pt]  [font=\scriptsize,color={rgb, 255:red, 31; green, 19; blue, 254 }  ,opacity=1 ] [align=left] {$\displaystyle \left( T^{+} \! \! ,K_{1}\right)$};
	% Text Node
	\draw (800.15,462.48) node [anchor=north west][inner sep=0.75pt]  [rotate=-269.35] [align=left] {\textbf{{\Large \textcolor[rgb]{0.07,0.09,1}{BranchNet}}}};
	% Text Node
	\draw (884,398.8) node [anchor=north west][inner sep=0.75pt]  [font=\small] [align=left] {\begin{minipage}[lt]{51.2pt}\setlength\topsep{0pt}
			\begin{center}
				Convolution\\BatchNorm\\Activation
			\end{center}
			
	\end{minipage}};
	% Text Node
	\draw (946,796) node [anchor=north west][inner sep=0.75pt]  [font=\large] [align=left] {$\displaystyle \mathcal{S}_{\theta }( u_{LR})( y)$};
	% Text Node
	\draw (1570.8,378.4) node [anchor=north west][inner sep=0.75pt]  [font=\large] [align=left] {$\displaystyle y$};
	% Text Node
	\draw (1581.8,401.4) node [anchor=north west][inner sep=0.75pt]  [font=\scriptsize,color={rgb, 255:red, 31; green, 19; blue, 254 }  ,opacity=1 ] [align=left] {$\displaystyle ( d)$};
	% Text Node
	\draw (1555.8,535.4) node [anchor=north west][inner sep=0.75pt]  [font=\large] [align=left] {MLP};
	% Text Node
	\draw (1578.8,569.4) node [anchor=north west][inner sep=0.75pt]  [font=\scriptsize,color={rgb, 255:red, 31; green, 19; blue, 254 }  ,opacity=1 ] [align=left] {$\displaystyle ( K)$};
	% Text Node
	\draw (1513.89,588.89) node [anchor=north west][inner sep=0.75pt]  [rotate=-270.02] [align=left] {\textbf{{\Large \textcolor[rgb]{0.07,0.09,1}{TrunkNet}}}};
	% Text Node
	\draw (1386,271.8) node [anchor=north west][inner sep=0.75pt]  [font=\small] [align=left] {\begin{minipage}[lt]{51.2pt}\setlength\topsep{0pt}
			\begin{center}
				Convolution\\BatchNorm\\Activation
			\end{center}
			
	\end{minipage}};
	% Text Node
	\draw (1386,479.8) node [anchor=north west][inner sep=0.75pt]  [font=\normalsize] [align=left] {\begin{minipage}[lt]{47.52pt}\setlength\topsep{0pt}
			\begin{center}
				Spatial\\Flattening
			\end{center}
			
	\end{minipage}};
	% Text Node
	\draw (1338,277.3) node [anchor=north west][inner sep=0.75pt]   [align=left] {{\large CNN}};
	% Text Node
	\draw (1382,572.8) node [anchor=north west][inner sep=0.75pt]  [font=\large] [align=left] {LSTM};
	% Text Node
	\draw (1410,598.8) node [anchor=north west][inner sep=0.75pt]  [font=\scriptsize,color={rgb, 255:red, 31; green, 19; blue, 254 }  ,opacity=1 ] [align=left] {$\displaystyle \left( T^{+} \! \! ,K_{2}\right)$};
	% Text Node
	\draw (1385,642.8) node [anchor=north west][inner sep=0.75pt]  [font=\large] [align=left] {MLP};
	% Text Node
	\draw (1410,668.8) node [anchor=north west][inner sep=0.75pt]  [font=\scriptsize,color={rgb, 255:red, 31; green, 19; blue, 254 }  ,opacity=1 ] [align=left] {$\displaystyle \left( T^{+} \! \! ,K\right)$};
	% Text Node
	\draw (1409,525.8) node [anchor=north west][inner sep=0.75pt]  [font=\scriptsize,color={rgb, 255:red, 31; green, 19; blue, 254 }  ,opacity=1 ] [align=left] {$\displaystyle \left( T^{+} \! \! ,K_{1}\right)$};
	% Text Node
	\draw (1302.15,463.48) node [anchor=north west][inner sep=0.75pt]  [rotate=-269.35] [align=left] {\textbf{{\Large \textcolor[rgb]{0.07,0.09,1}{BranchNet}}}};
	% Text Node
	\draw (1386,399.8) node [anchor=north west][inner sep=0.75pt]  [font=\small] [align=left] {\begin{minipage}[lt]{51.2pt}\setlength\topsep{0pt}
			\begin{center}
				Convolution\\BatchNorm\\Activation
			\end{center}
			
	\end{minipage}};
	% Text Node
	\draw (1387.4,190) node [anchor=north west][inner sep=0.75pt]  [font=\large] [align=left] {$\displaystyle u_{HR}^{( 0)}$};
	% Text Node
	\draw (1408.4,223) node [anchor=north west][inner sep=0.75pt]  [font=\scriptsize,color={rgb, 255:red, 31; green, 19; blue, 254 }  ,opacity=1 ] [align=left] {$\displaystyle ( N_{1} \ ,...,N_{d})$};
	% Text Node
	\draw (824.8,854.4) node [anchor=north west][inner sep=0.75pt]  [font=\fontsize{1.3em}{1.56em}\selectfont] [align=left] {\textbf{{\large Example of 2-SubNetwork SROpNet}}};
	% Text Node
	\draw (1404.8,855.4) node [anchor=north west][inner sep=0.75pt]  [font=\fontsize{1.3em}{1.56em}\selectfont] [align=left] {\textbf{{\large nonSR Counterpart}}};
	% Text Node
	\draw (1025.8,705.4) node [anchor=north west][inner sep=0.75pt]  [font=\scriptsize,color={rgb, 255:red, 0; green, 186; blue, 91 }  ,opacity=1 ] [align=left] {$\displaystyle [ T_{j}]$};
	% Text Node
	\draw (907.45,707.3) node [anchor=north west][inner sep=0.75pt]  [font=\scriptsize,color={rgb, 255:red, 0; green, 186; blue, 81 }  ,opacity=1 ] [align=left] {$\displaystyle [ B_{ij}]$};
	% Text Node
	\draw (1446,789) node [anchor=north west][inner sep=0.75pt]  [font=\large] [align=left] {$\displaystyle \mathcal{G}_{\theta }\left( u_{HR}^{( 0)}\right)( y)$};
	% Text Node
	\draw (1460.07,731.93) node [anchor=north west][inner sep=0.75pt]  [font=\normalsize] [align=left] {$\displaystyle \Sigma _{j} B_{ij} T_{j}$};
	% Text Node
	\draw (1493.5,755.5) node [anchor=north west][inner sep=0.75pt]  [font=\scriptsize,color={rgb, 255:red, 31; green, 19; blue, 254 }  ,opacity=1 ] [align=left] {$\displaystyle \left( T^{+}\right)$};
	% Text Node
	\draw (1527.8,706.4) node [anchor=north west][inner sep=0.75pt]  [font=\scriptsize,color={rgb, 255:red, 0; green, 186; blue, 91 }  ,opacity=1 ] [align=left] {$\displaystyle [ T_{j}]$};
	% Text Node
	\draw (1409.45,708.3) node [anchor=north west][inner sep=0.75pt]  [font=\scriptsize,color={rgb, 255:red, 0; green, 186; blue, 81 }  ,opacity=1 ] [align=left] {$\displaystyle [ B_{ij}]$};
	% Text Node
	\draw (1181.8,1146.53) node [anchor=north west][inner sep=0.75pt]  [font=\large] [align=left] {$\displaystyle y$};
	% Text Node
	\draw (1211.6,1131.4) node [anchor=north west][inner sep=0.75pt]  [font=\scriptsize,color={rgb, 255:red, 31; green, 19; blue, 254 }  ,opacity=1 ] [align=left] {$\displaystyle ( d)$};
	% Text Node
	\draw (1395.6,1131.3) node [anchor=north west][inner sep=0.75pt]  [font=\scriptsize,color={rgb, 255:red, 31; green, 19; blue, 254 }  ,opacity=1 ] [align=left] {$\displaystyle ( K)$};
	% Text Node
	\draw (1519.07,1053.07) node [anchor=north west][inner sep=0.75pt]  [font=\large] [align=left] {$\displaystyle \Sigma _{j} B_{ij} S_{ij} T_{j}$};
	% Text Node
	\draw (1603.5,1034.63) node [anchor=north west][inner sep=0.75pt]  [font=\scriptsize,color={rgb, 255:red, 31; green, 19; blue, 254 }  ,opacity=1 ] [align=left] {$\displaystyle \left( T^{+}\right)$};
	% Text Node
	\draw (743,966.93) node [anchor=north west][inner sep=0.75pt]  [font=\large] [align=left] {$\displaystyle u_{LR}$};
	% Text Node
	\draw (783,952.93) node [anchor=north west][inner sep=0.75pt]  [font=\scriptsize,color={rgb, 255:red, 31; green, 19; blue, 254 }  ,opacity=1 ] [align=left] {$\displaystyle ( T,N_{1} \ ,...,N_{d})$};
	% Text Node
	\draw (901,962.93) node [anchor=north west][inner sep=0.75pt]  [font=\large] [align=left] {Time Upscaling};
	% Text Node
	\draw (881.97,923.02) node [anchor=north west][inner sep=0.75pt]  [rotate=-359.91] [align=left] {\textbf{{\Large \textcolor[rgb]{0.07,0.09,1}{BranchNet}}}};
	% Text Node
	\draw (1640,1052.13) node [anchor=north west][inner sep=0.75pt]  [font=\large] [align=left] {$\displaystyle \mathcal{S}_{\theta }( u_{LR})( y)$};
	% Text Node
	\draw (1472.8,1086.53) node [anchor=north west][inner sep=0.75pt]  [font=\scriptsize,color={rgb, 255:red, 0; green, 186; blue, 91 }  ,opacity=1 ] [align=left] {$\displaystyle [ T_{j}]$};
	% Text Node
	\draw (1140,955.93) node [anchor=north west][inner sep=0.75pt]  [font=\normalsize] [align=left] {\begin{minipage}[lt]{47.52pt}\setlength\topsep{0pt}
			\begin{center}
				Spatial\\Flattening
			\end{center}
			
	\end{minipage}};
	% Text Node
	\draw (1059,964.93) node [anchor=north west][inner sep=0.75pt]  [font=\large] [align=left] {CNN};
	% Text Node
	\draw (1249,964.93) node [anchor=north west][inner sep=0.75pt]  [font=\large] [align=left] {LSTM};
	% Text Node
	\draw (1338.6,963.6) node [anchor=north west][inner sep=0.75pt]  [font=\large] [align=left] {MLP};
	% Text Node
	\draw (1384.6,949.6) node [anchor=north west][inner sep=0.75pt]  [font=\scriptsize,color={rgb, 255:red, 31; green, 19; blue, 254 }  ,opacity=1 ] [align=left] {$\displaystyle \left( T^{+} \! \! ,K\right)$};
	% Text Node
	\draw (1473.45,1024) node [anchor=north west][inner sep=0.75pt]  [font=\scriptsize,color={rgb, 255:red, 0; green, 186; blue, 81 }  ,opacity=1 ] [align=left] {$\displaystyle [ B_{ij}]$};
	% Text Node
	\draw (777,1041.93) node [anchor=north west][inner sep=0.75pt]  [font=\scriptsize,color={rgb, 255:red, 31; green, 19; blue, 254 }  ,opacity=1 ] [align=left] {$\displaystyle ( d,T,N_{1} \ ,...,N_{d})$};
	% Text Node
	\draw (901,1052.93) node [anchor=north west][inner sep=0.75pt]  [font=\large] [align=left] {Time Upscaling};
	% Text Node
	\draw (881.97,1013.02) node [anchor=north west][inner sep=0.75pt]  [rotate=-359.91] [align=left] {\textbf{{\Large \textcolor[rgb]{0.07,0.09,1}{SensorNet}}}};
	% Text Node
	\draw (1140,1045.93) node [anchor=north west][inner sep=0.75pt]  [font=\normalsize] [align=left] {\begin{minipage}[lt]{47.52pt}\setlength\topsep{0pt}
			\begin{center}
				Spatial\\Flattening
			\end{center}
			
	\end{minipage}};
	% Text Node
	\draw (1059,1054.93) node [anchor=north west][inner sep=0.75pt]  [font=\large] [align=left] {CNN};
	% Text Node
	\draw (1249,1054.93) node [anchor=north west][inner sep=0.75pt]  [font=\large] [align=left] {LSTM};
	% Text Node
	\draw (1338.6,1053.6) node [anchor=north west][inner sep=0.75pt]  [font=\large] [align=left] {MLP};
	% Text Node
	\draw (1384.6,1039.6) node [anchor=north west][inner sep=0.75pt]  [font=\scriptsize,color={rgb, 255:red, 31; green, 19; blue, 254 }  ,opacity=1 ] [align=left] {$\displaystyle \left( T^{+} \! \! ,K\right)$};
	% Text Node
	\draw (1452.45,1045) node [anchor=north west][inner sep=0.75pt]  [font=\scriptsize,color={rgb, 255:red, 0; green, 186; blue, 81 }  ,opacity=1 ] [align=left] {$\displaystyle [ S_{ij}]$};
	% Text Node
	\draw (752,1060.2) node [anchor=north west][inner sep=0.75pt]  [font=\large] [align=left] {$\displaystyle x$};
	% Text Node
	\draw (1325.6,1144.6) node [anchor=north west][inner sep=0.75pt]  [font=\large] [align=left] {MLP};
	% Text Node
	\draw (1265.97,1104.02) node [anchor=north west][inner sep=0.75pt]  [rotate=-359.91] [align=left] {\textbf{{\Large \textcolor[rgb]{0.07,0.09,1}{TrunkNet}}}};
	% Text Node
	\draw (1099.8,1205.6) node [anchor=north west][inner sep=0.75pt]  [font=\fontsize{1.3em}{1.56em}\selectfont] [align=left] {\textbf{{\large Example of 3-SubNetwork SROpNet}}};

\end{tikzpicture}

}

\end{center}

%% file: diagrams/ComputerVision_Example.tex
\begin{center}

\resizebox{0.89\linewidth}{!}{

\tikzset{every picture/.style={line width=0.75pt}} %set default line width to 0.75pt        

\begin{tikzpicture}[x=0.75pt,y=0.75pt,yscale=-1,xscale=1]
	%uncomment if require: \path (0,2566); %set diagram left start at 0, and has height of 2566
	
	%Shape: Ellipse [id:dp7806965461673661] 
	\draw   (784.8,2288.51) .. controls (784.8,2280.23) and (795.99,2273.51) .. (809.8,2273.51) .. controls (823.61,2273.51) and (834.8,2280.23) .. (834.8,2288.51) .. controls (834.8,2296.79) and (823.61,2303.51) .. (809.8,2303.51) .. controls (795.99,2303.51) and (784.8,2296.79) .. (784.8,2288.51) -- cycle ;
	%Straight Lines [id:da929647822606833] 
	\draw    (834.6,2288.38) -- (920.8,2288.77) ;
	\draw [shift={(922.8,2288.78)}, rotate = 180.26] [color={rgb, 255:red, 0; green, 0; blue, 0 }  ][line width=0.75]    (10.93,-3.29) .. controls (6.95,-1.4) and (3.31,-0.3) .. (0,0) .. controls (3.31,0.3) and (6.95,1.4) .. (10.93,3.29)   ;
	%Shape: Rectangle [id:dp9933544194403812] 
	\draw   (1119.8,2223.18) -- (1192.5,2223.18) -- (1192.5,2258.18) -- (1119.8,2258.18) -- cycle ;
	%Straight Lines [id:da5973131113434322] 
	\draw    (1192.5,2239.61) -- (1219.6,2239.58) ;
	\draw [shift={(1221.6,2239.58)}, rotate = 179.93] [color={rgb, 255:red, 0; green, 0; blue, 0 }  ][line width=0.75]    (10.93,-3.29) .. controls (6.95,-1.4) and (3.31,-0.3) .. (0,0) .. controls (3.31,0.3) and (6.95,1.4) .. (10.93,3.29)   ;
	%Rounded Rect [id:dp2651814395924781] 
	\draw  [color={rgb, 255:red, 22; green, 19; blue, 254 }  ,draw opacity=1 ][line width=1.5]  (875.6,2272.18) .. controls (875.6,2265.88) and (880.7,2260.78) .. (887,2260.78) -- (1050.1,2260.78) .. controls (1056.4,2260.78) and (1061.5,2265.88) .. (1061.5,2272.18) -- (1061.5,2306.38) .. controls (1061.5,2312.67) and (1056.4,2317.78) .. (1050.1,2317.78) -- (887,2317.78) .. controls (880.7,2317.78) and (875.6,2312.67) .. (875.6,2306.38) -- cycle ;
	%Shape: Ellipse [id:dp21144663410516573] 
	\draw   (560,2188.91) .. controls (560,2180.63) and (571.19,2173.91) .. (585,2173.91) .. controls (598.81,2173.91) and (610,2180.63) .. (610,2188.91) .. controls (610,2197.19) and (598.81,2203.91) .. (585,2203.91) .. controls (571.19,2203.91) and (560,2197.19) .. (560,2188.91) -- cycle ;
	%Straight Lines [id:da802696671682597] 
	\draw    (610,2188.91) -- (682.6,2189.17) ;
	\draw [shift={(684.6,2189.18)}, rotate = 180.2] [color={rgb, 255:red, 0; green, 0; blue, 0 }  ][line width=0.75]    (10.93,-3.29) .. controls (6.95,-1.4) and (3.31,-0.3) .. (0,0) .. controls (3.31,0.3) and (6.95,1.4) .. (10.93,3.29)   ;
	%Shape: Rectangle [id:dp36203439903862367] 
	\draw   (687,2172.91) -- (909,2172.91) -- (909,2202.91) -- (687,2202.91) -- cycle ;
	%Rounded Rect [id:dp6539113125439491] 
	\draw  [color={rgb, 255:red, 22; green, 19; blue, 254 }  ,draw opacity=1 ][line width=1.5]  (659.63,2170.38) .. controls (659.63,2163.64) and (665.1,2158.18) .. (671.83,2158.18) -- (1050.3,2158.18) .. controls (1057.04,2158.18) and (1062.5,2163.64) .. (1062.5,2170.38) -- (1062.5,2206.98) .. controls (1062.5,2213.71) and (1057.04,2219.18) .. (1050.3,2219.18) -- (671.83,2219.18) .. controls (665.1,2219.18) and (659.63,2213.71) .. (659.63,2206.98) -- cycle ;
	%Shape: Ellipse [id:dp08150564255915138] 
	\draw   (1223,2239.69) .. controls (1223,2225.29) and (1244.27,2213.61) .. (1270.5,2213.61) .. controls (1296.73,2213.61) and (1318,2225.29) .. (1318,2239.69) .. controls (1318,2254.1) and (1296.73,2265.78) .. (1270.5,2265.78) .. controls (1244.27,2265.78) and (1223,2254.1) .. (1223,2239.69) -- cycle ;
	%Straight Lines [id:da1591732679335509] 
	\draw    (908.6,2188.58) -- (961.6,2188.58) ;
	\draw [shift={(963.6,2188.58)}, rotate = 180] [color={rgb, 255:red, 0; green, 0; blue, 0 }  ][line width=0.75]    (10.93,-3.29) .. controls (6.95,-1.4) and (3.31,-0.3) .. (0,0) .. controls (3.31,0.3) and (6.95,1.4) .. (10.93,3.29)   ;
	%Shape: Rectangle [id:dp6788517166318704] 
	\draw   (965.6,2172.58) -- (1014.6,2172.58) -- (1014.6,2202.58) -- (965.6,2202.58) -- cycle ;
	%Shape: Rectangle [id:dp06502355447206964] 
	\draw   (924.6,2274.28) -- (1014.6,2274.28) -- (1014.6,2304.28) -- (924.6,2304.28) -- cycle ;
	%Curve Lines [id:da047047774819393595] 
	\draw    (1014,2290.18) .. controls (1060.04,2289.51) and (1059.52,2248.34) .. (1114.81,2246.54) ;
	\draw [shift={(1116.5,2246.5)}, rotate = 178.99] [color={rgb, 255:red, 0; green, 0; blue, 0 }  ][line width=0.75]    (10.93,-3.29) .. controls (6.95,-1.4) and (3.31,-0.3) .. (0,0) .. controls (3.31,0.3) and (6.95,1.4) .. (10.93,3.29)   ;
	%Curve Lines [id:da7919701443789724] 
	\draw    (1015,2189.18) .. controls (1062.03,2189.5) and (1059.55,2233.6) .. (1114.81,2235.46) ;
	\draw [shift={(1116.5,2235.5)}, rotate = 181.01] [color={rgb, 255:red, 0; green, 0; blue, 0 }  ][line width=0.75]    (10.93,-3.29) .. controls (6.95,-1.4) and (3.31,-0.3) .. (0,0) .. controls (3.31,0.3) and (6.95,1.4) .. (10.93,3.29)   ;
	
	% Text Node
	\draw (803.8,2284.51) node [anchor=north west][inner sep=0.75pt]  [font=\large] [align=left] {$\displaystyle y$};
	% Text Node
	\draw (833.6,2269.38) node [anchor=north west][inner sep=0.75pt]  [font=\scriptsize,color={rgb, 255:red, 31; green, 19; blue, 254 }  ,opacity=1 ] [align=left] {$\displaystyle ( 2)$};
	% Text Node
	\draw (1017.6,2269.28) node [anchor=north west][inner sep=0.75pt]  [font=\scriptsize,color={rgb, 255:red, 31; green, 19; blue, 254 }  ,opacity=1 ] [align=left] {$\displaystyle ( K)$};
	% Text Node
	\draw (1129.07,2232.04) node [anchor=north west][inner sep=0.75pt]  [font=\large] [align=left] {$\displaystyle \Sigma _{j} B_{ij} T_{j}$};
	% Text Node
	\draw (1193.5,2216.61) node [anchor=north west][inner sep=0.75pt]  [font=\scriptsize,color={rgb, 255:red, 31; green, 19; blue, 254 }  ,opacity=1 ] [align=left] {$\displaystyle \left( T^{+}\right)$};
	% Text Node
	\draw (570,2183.91) node [anchor=north west][inner sep=0.75pt]  [font=\large] [align=left] {$\displaystyle u_{LR}$};
	% Text Node
	\draw (610,2169.91) node [anchor=north west][inner sep=0.75pt]  [font=\scriptsize,color={rgb, 255:red, 31; green, 19; blue, 254 }  ,opacity=1 ] [align=left] {$\displaystyle ( T,N)$};
	% Text Node
	\draw (705,2178.91) node [anchor=north west][inner sep=0.75pt]  [font=\Large] [align=left] {Existing SR Approach};
	% Text Node
	\draw (674.97,2140) node [anchor=north west][inner sep=0.75pt]  [rotate=-359.91] [align=left] {\textbf{{\Large \textcolor[rgb]{0.07,0.09,1}{BranchNet}}}};
	% Text Node
	\draw (1230,2231.11) node [anchor=north west][inner sep=0.75pt]  [font=\large] [align=left] {$\displaystyle \mathcal{S}_{\theta }( u_{LR})( y)$};
	% Text Node
	\draw (1088.8,2257.51) node [anchor=north west][inner sep=0.75pt]  [font=\scriptsize,color={rgb, 255:red, 0; green, 186; blue, 91 }  ,opacity=1 ] [align=left] {$\displaystyle [ T_{j}]$};
	% Text Node
	\draw (970.6,2180.58) node [anchor=north west][inner sep=0.75pt]  [font=\large] [align=left] {MLP};
	% Text Node
	\draw (1016.6,2169.58) node [anchor=north west][inner sep=0.75pt]  [font=\scriptsize,color={rgb, 255:red, 31; green, 19; blue, 254 }  ,opacity=1 ] [align=left] {$\displaystyle \left( T^{+} \! \! ,K\right)$};
	% Text Node
	\draw (1086.45,2213.98) node [anchor=north west][inner sep=0.75pt]  [font=\scriptsize,color={rgb, 255:red, 0; green, 186; blue, 81 }  ,opacity=1 ] [align=left] {$\displaystyle [ B_{ij}]$};
	% Text Node
	\draw (947.6,2282.58) node [anchor=north west][inner sep=0.75pt]  [font=\large] [align=left] {MLP};
	% Text Node
	\draw (887.97,2242) node [anchor=north west][inner sep=0.75pt]  [rotate=-359.91] [align=left] {\textbf{{\Large \textcolor[rgb]{0.07,0.09,1}{TrunkNet}}}};
	% Text Node
	\draw (910.6,2168.58) node [anchor=north west][inner sep=0.75pt]  [font=\scriptsize,color={rgb, 255:red, 31; green, 19; blue, 254 }  ,opacity=1 ] [align=left] {$\displaystyle \left( T^{+} \! \! ,N^{+}\right)$};

\end{tikzpicture}

}

\end{center}

%% file: TexFiles/Forced1D_Diffusion_Exp1.tex
\begin{figure}[!h]
\centering
\includegraphics[width=0.99\textwidth]{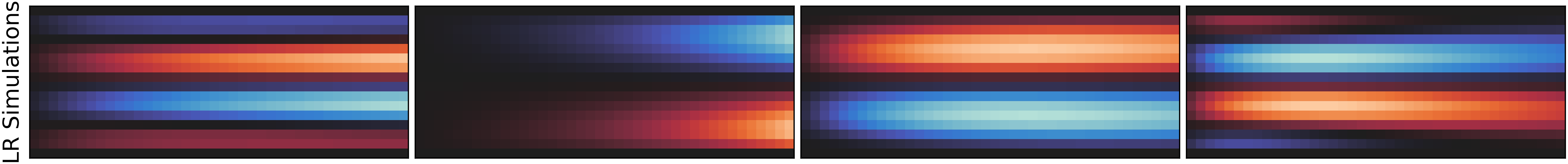} 
\vspace{0.4mm}

\includegraphics[width=0.99\textwidth]{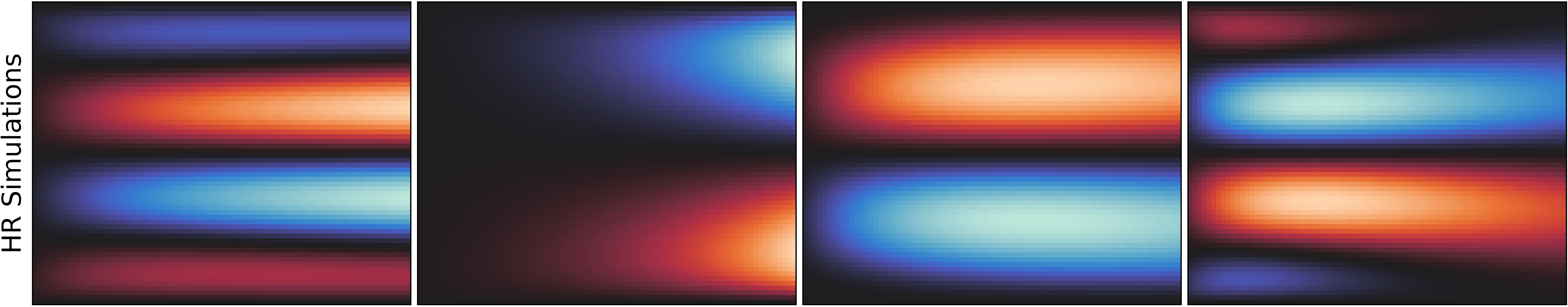} 
\vspace{2mm}.

\includegraphics[width=0.99\textwidth]{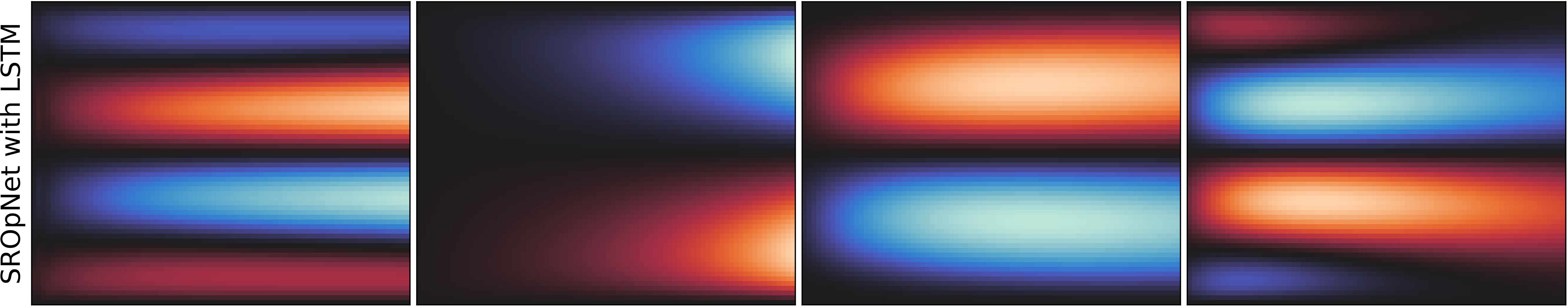} 
\vspace{-0.25mm}

\includegraphics[width=0.99\textwidth]{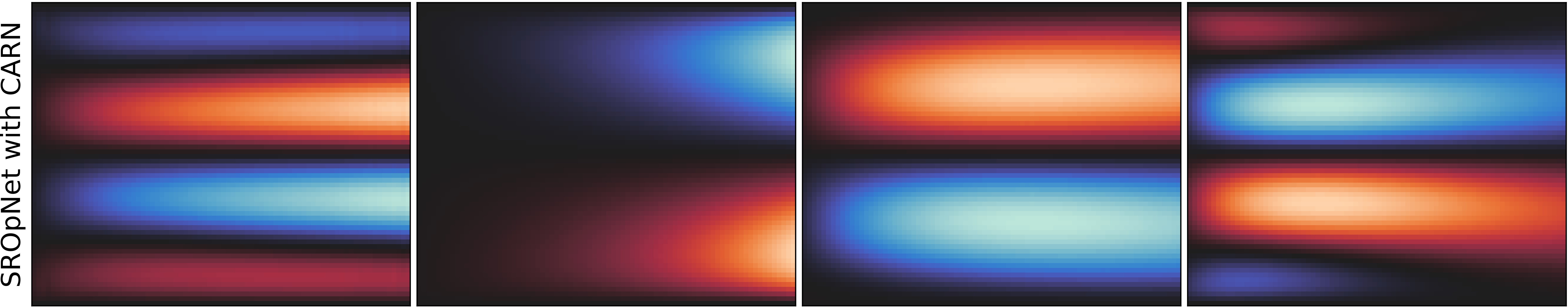} 
\vspace{-0.25mm}

\includegraphics[width=0.99\textwidth]{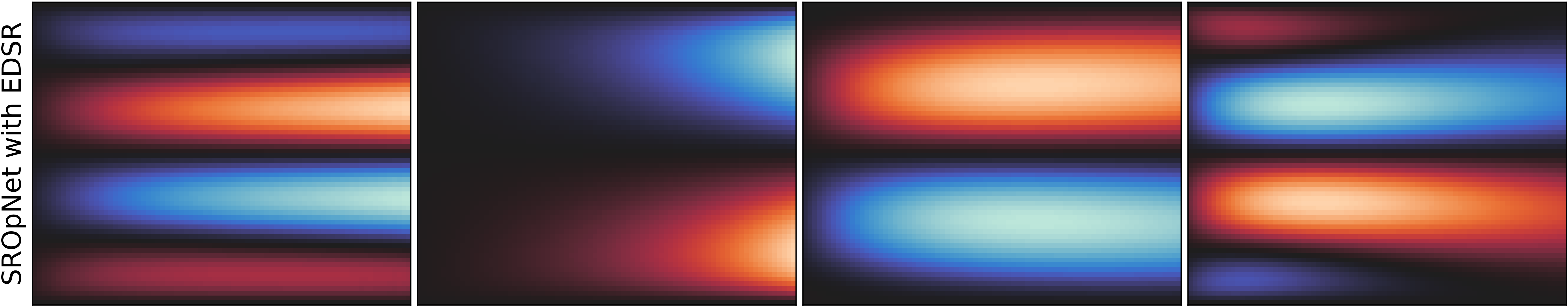} 
\vspace{-0.25mm}

\includegraphics[width=0.99\textwidth]{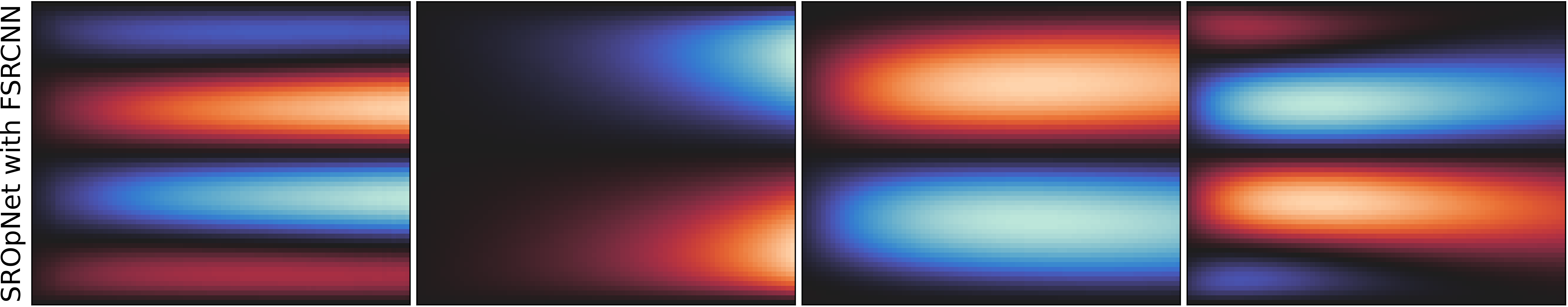} 
\vspace{-0.25mm}

\includegraphics[width=0.99\textwidth]{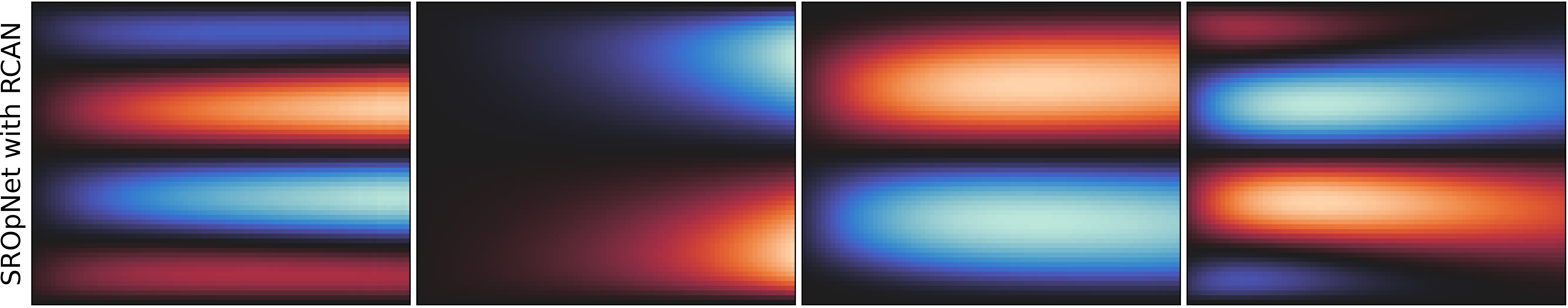} 
\vspace{-2.7mm}

\caption{Results using different SROpNets on 4 previously-unobserved samples from the dataset of solutions to the 1D Forced Diffusion equation~\eqref{eq: Diffusion 1D Exp1}. \label{fig: Diffusion 1D Exp1}  }
\end{figure}

%% file: TexFiles/Forced1D_Diffusion_Exp2.tex
\begin{figure}[!h]
	\vspace{-1.5mm}
\centering
\includegraphics[width=0.98\textwidth]{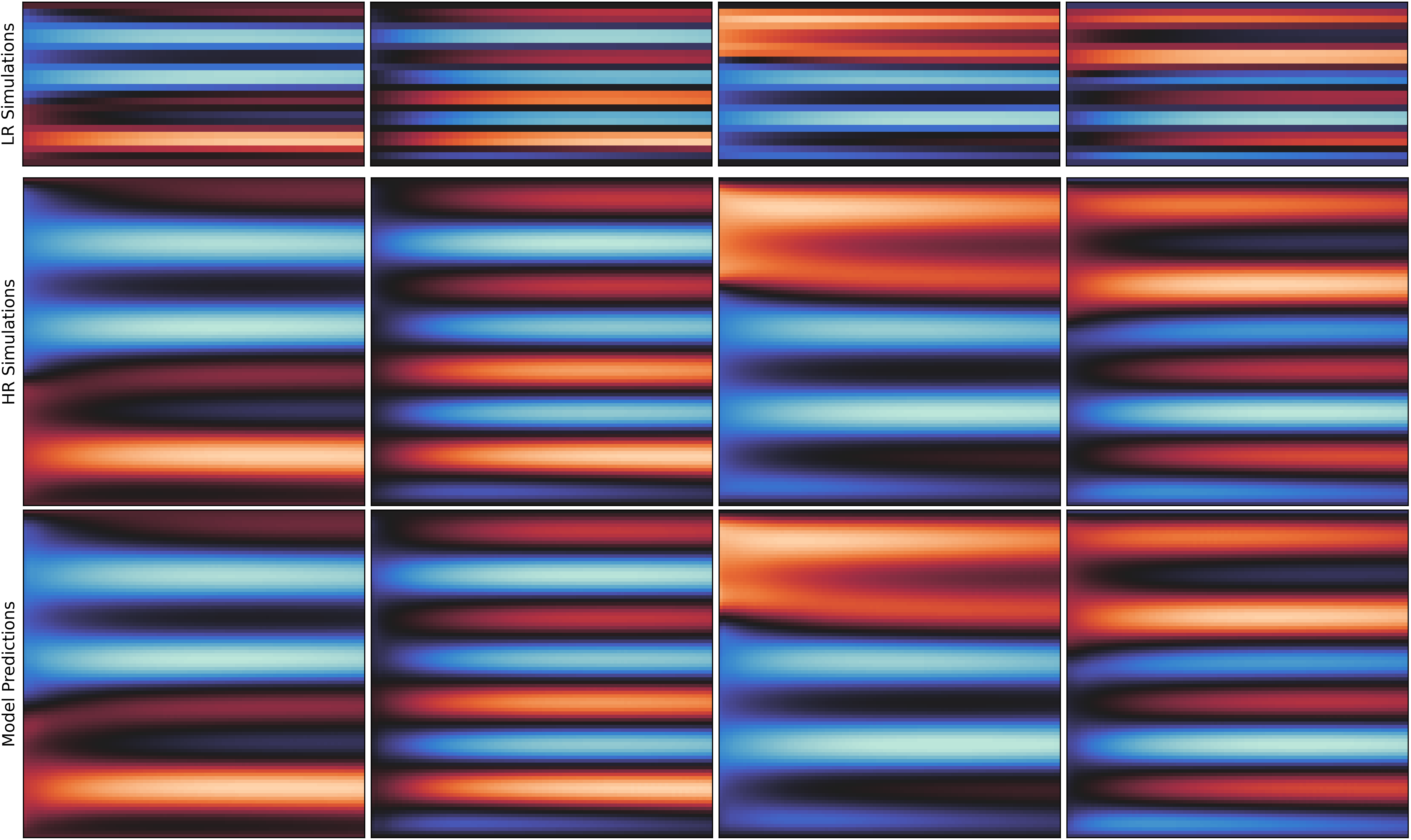}  \vspace{-2.1mm}
\caption{Super-resolution results for 4 previously-unobserved samples from the dataset of solutions to the 1D Forced Diffusion equation~\eqref{eq: Diffusion 1D Exp2}. \label{fig: Diffusion 1D Exp2} }
\end{figure}

%% file: TexFiles/Forced1D_Diffusion_Exp3.tex
\begin{figure}[!h]
	\vspace{-1.4mm}
	\centering
	\includegraphics[width=0.905\textwidth]{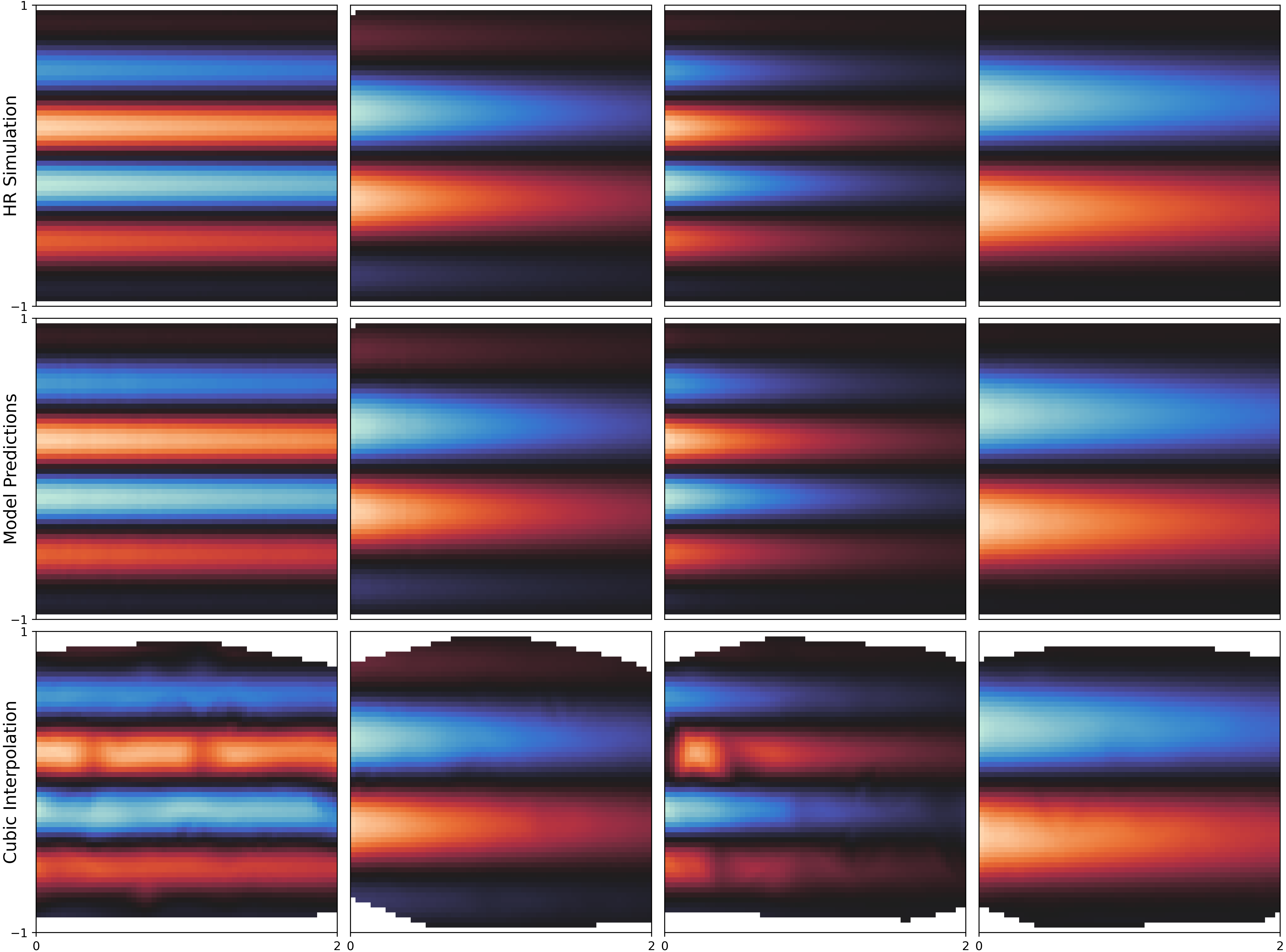}  \vspace{-3.3mm} 
	\caption{Results for 4 previously-unobserved samples from the dataset of solutions to the forced diffusion equation~\eqref{eq: Diffusion 1D with Random Grids} with random sensor and prediction locations. \label{fig: 1D Exp3 Results} }
\end{figure}

%% file: TexFiles/2DDiffusion_Fixed.tex
\begin{figure}[!h]
\vspace{1mm}

\centering

\includegraphics[width=0.93\textwidth]{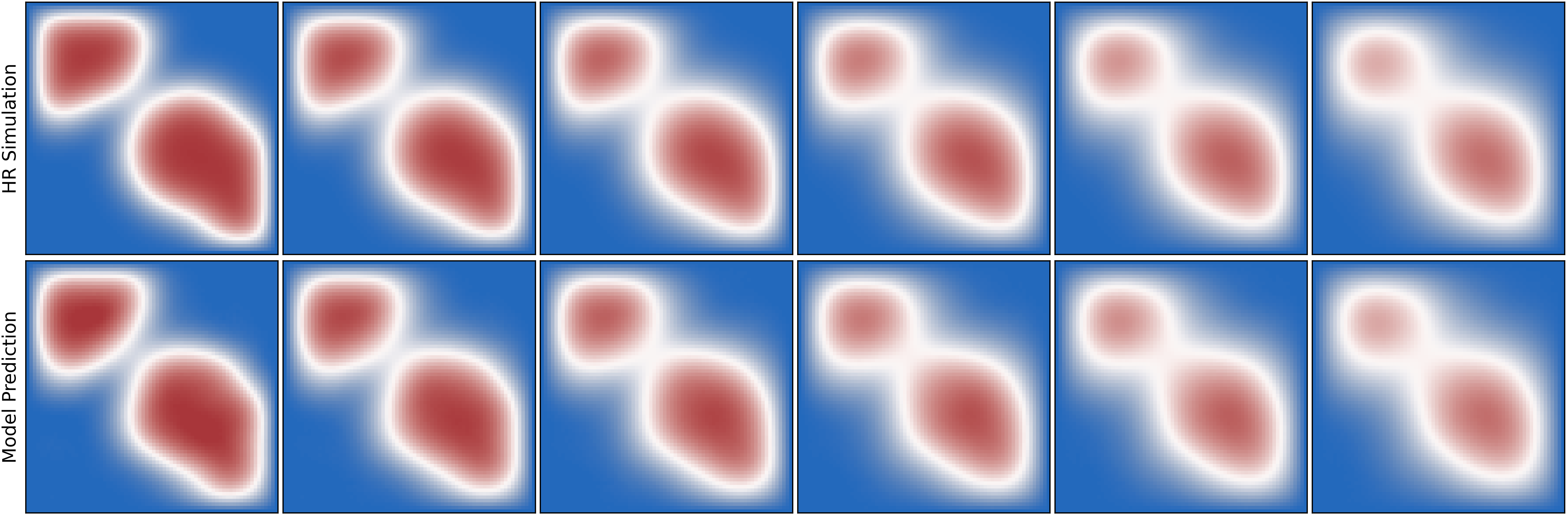} 
\vspace{0.8mm}

\includegraphics[width=0.93\textwidth]{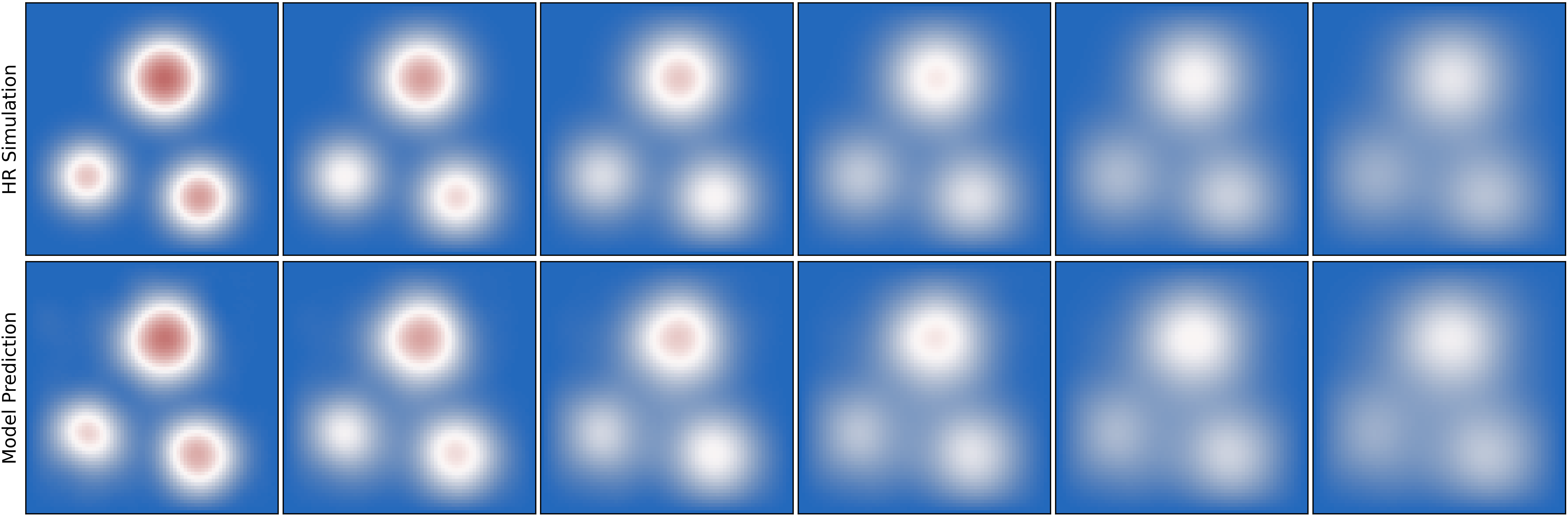} 
\vspace{0.8mm}

\includegraphics[width=0.93\textwidth]{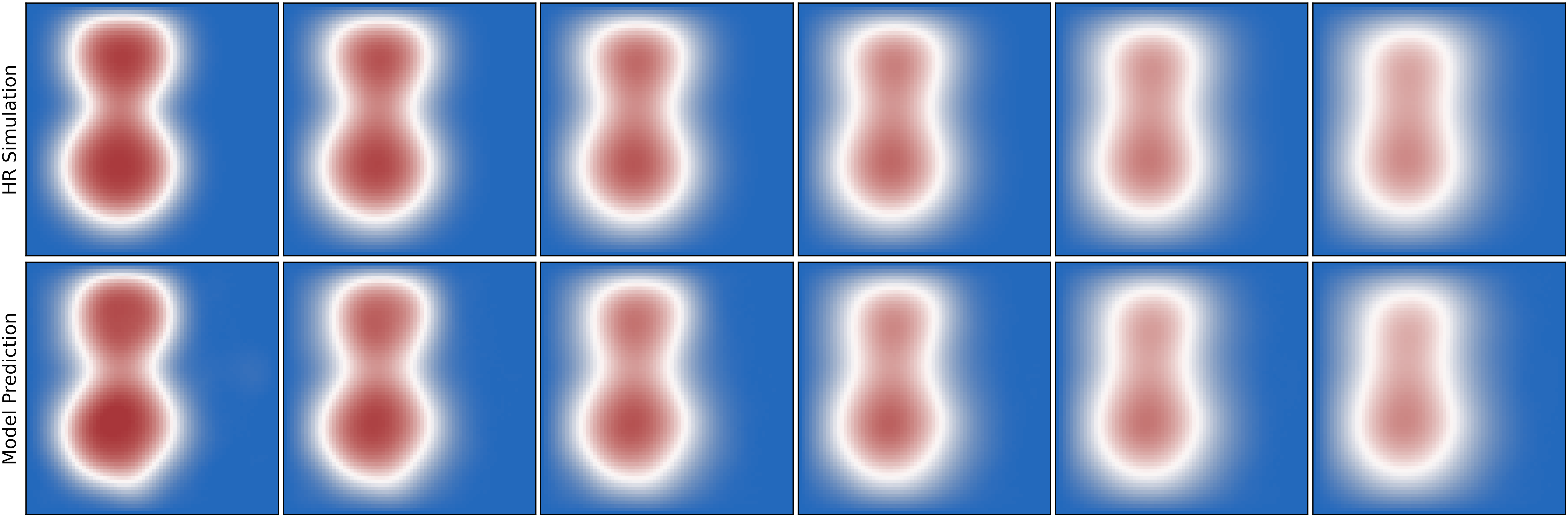} 
\vspace{0.8mm}

\includegraphics[width=0.93\textwidth]{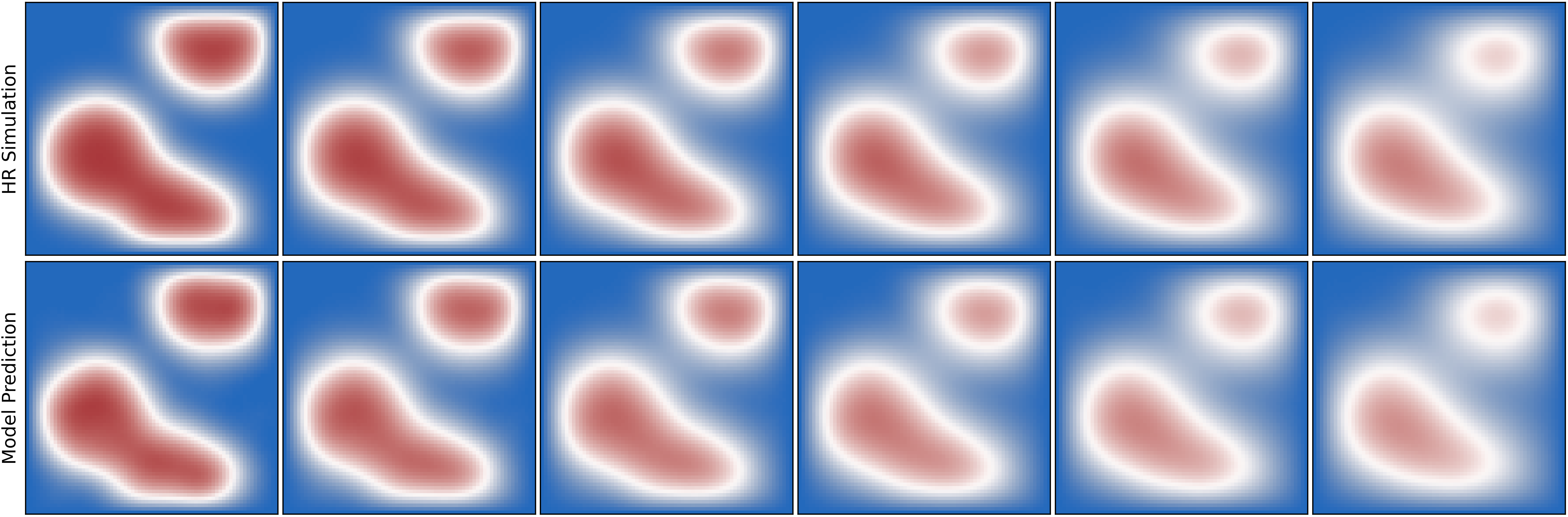} 
\vspace{0mm}

\caption{Super-resolution results for 4 previously-unobserved sequences from the dataset of solutions to the 2D Diffusion equation~\eqref{eq: 2D Diffusion} with fixed diffusion constant. \label{fig: Diffusion 2D} }
\end{figure}

%% file: TexFiles/2DDiffusion_VariableSpeeds.tex
\begin{figure}[!h]

\centering

\includegraphics[width=0.86\textwidth]{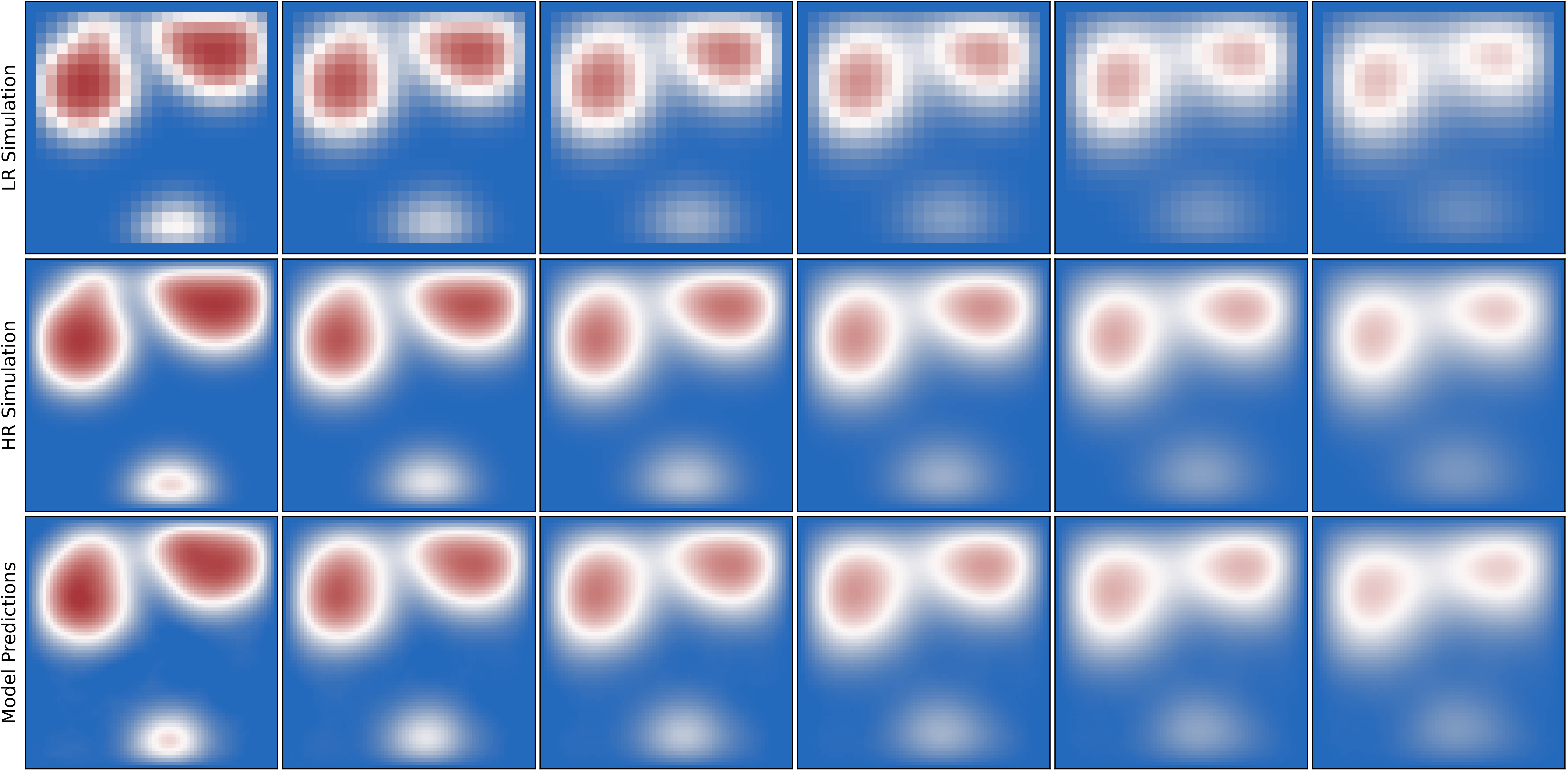} 
\vspace{0.5mm}

\includegraphics[width=0.86\textwidth]{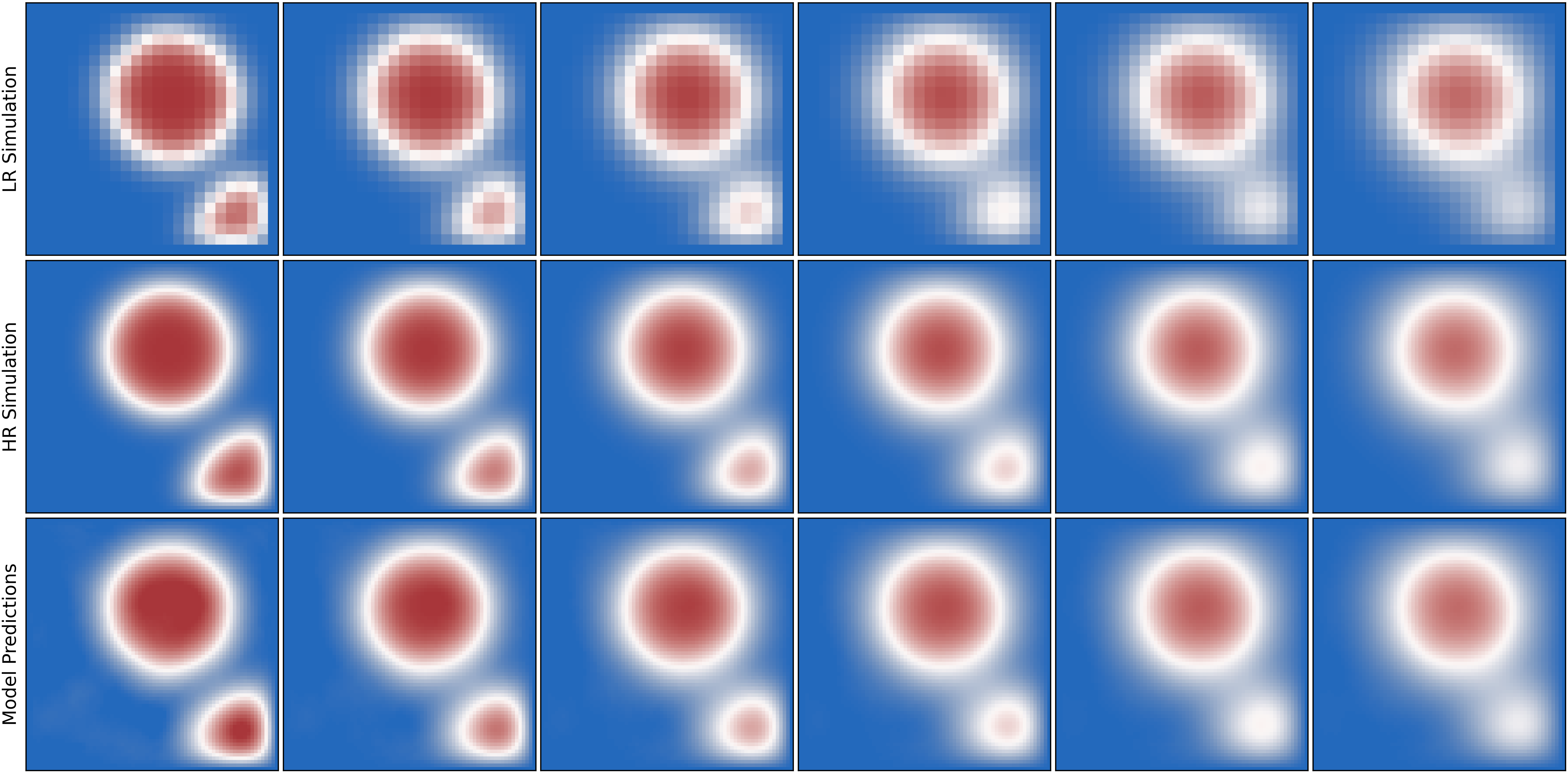}  
\vspace{0.5mm}

\includegraphics[width=0.86\textwidth]{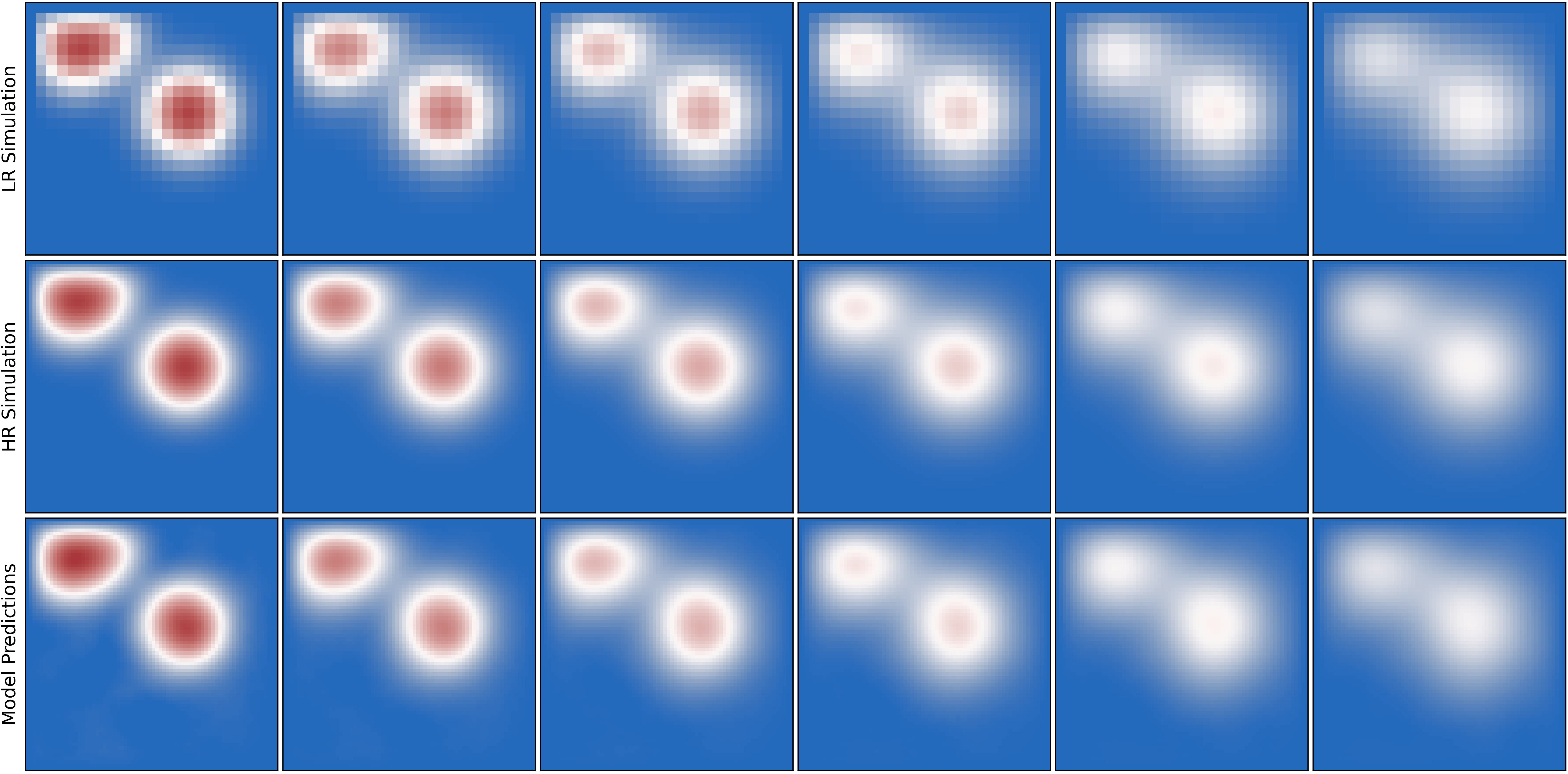} 
\vspace{-2mm}

\caption{Super-resolution results for 3 previously-unobserved sequences from the dataset of solutions to the 2D Diffusion equation with different diffusion constants. \label{fig: Diffusion 2D Speeds Comparison} }
\end{figure}

\newpage

\begin{figure}[!h]

\centering

\includegraphics[width=0.84\textwidth]{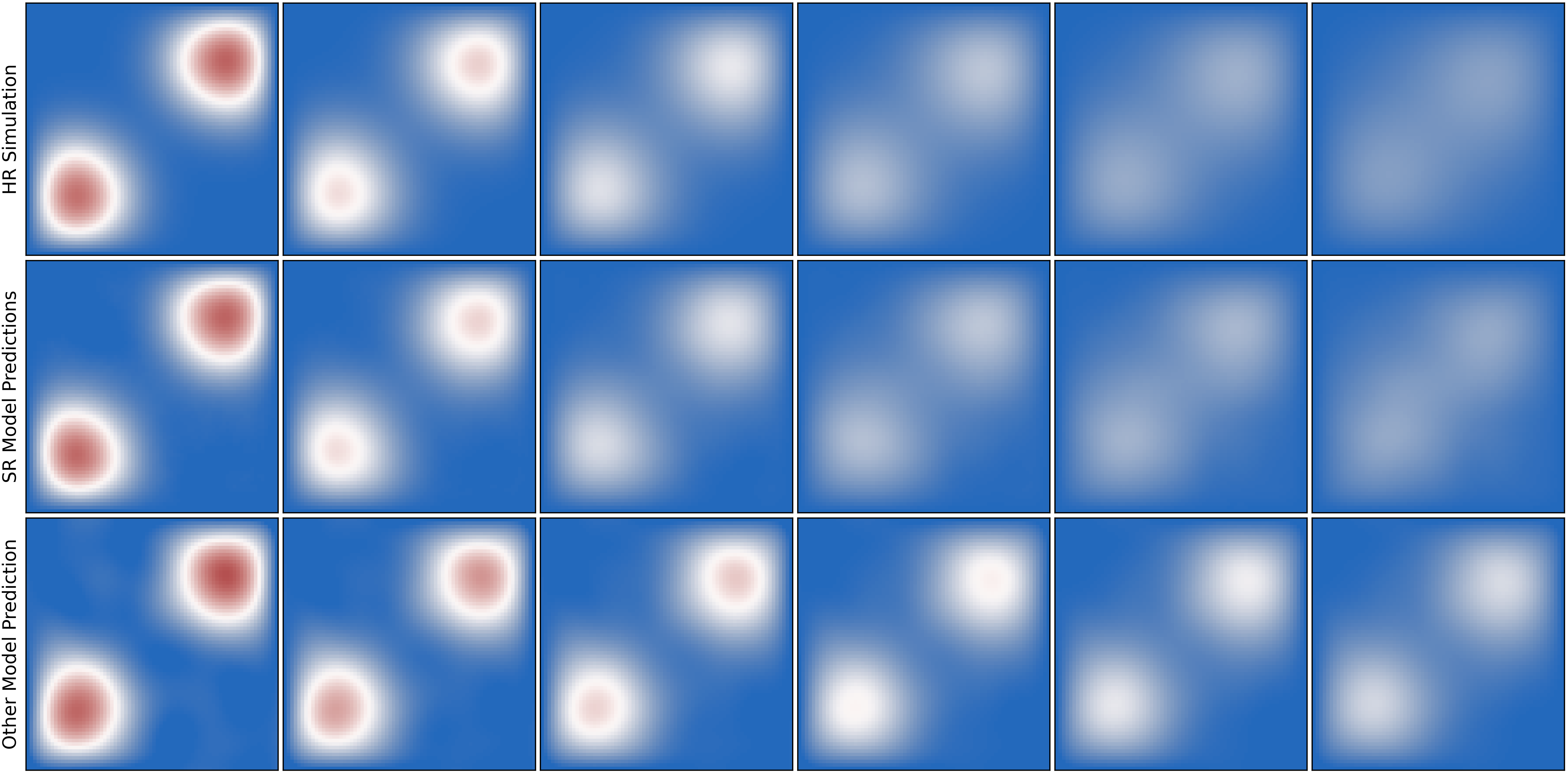} 
\vspace{0.5mm}

\includegraphics[width=0.84\textwidth]{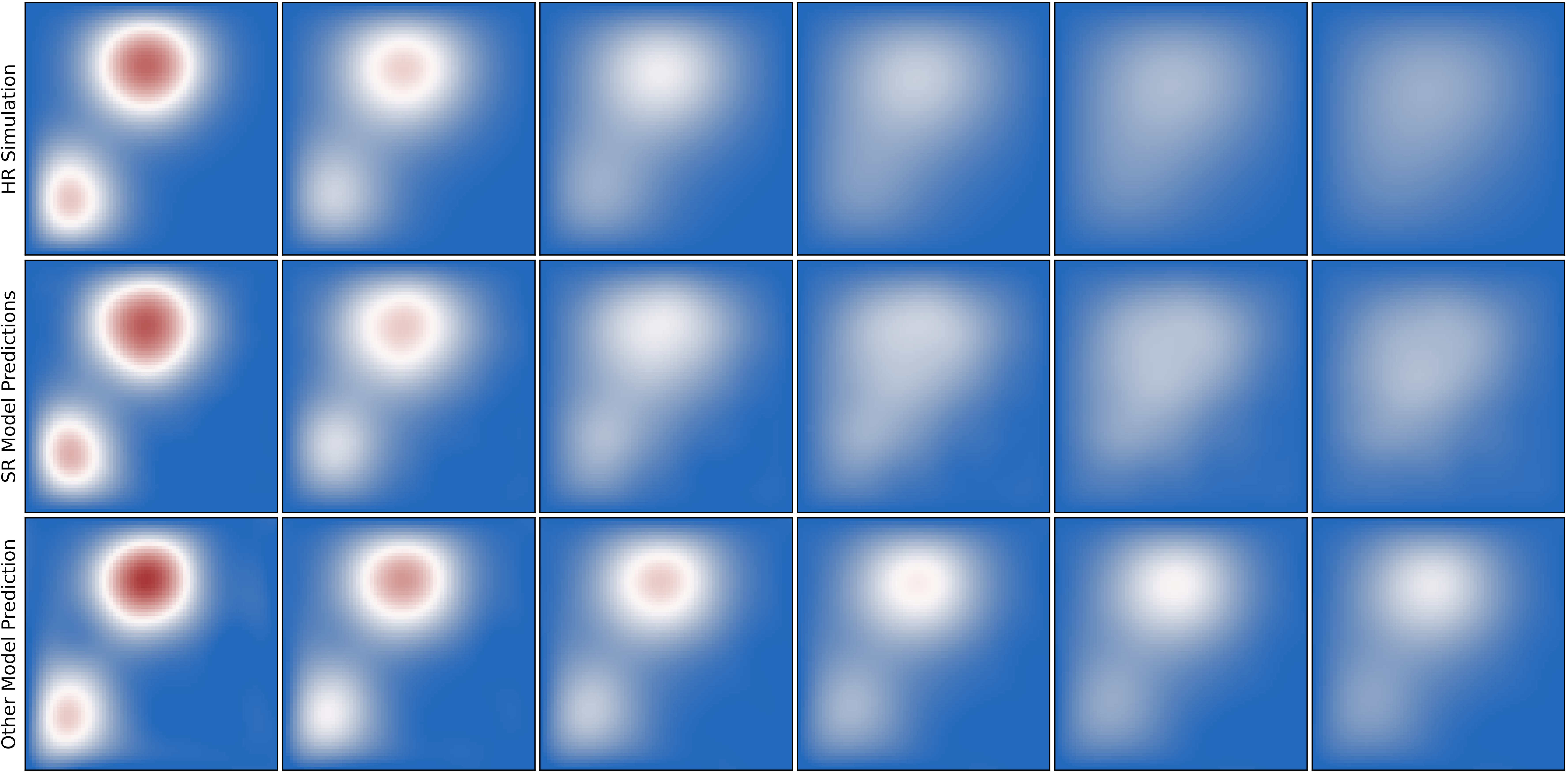} 
\vspace{0.5mm}

\includegraphics[width=0.84\textwidth]{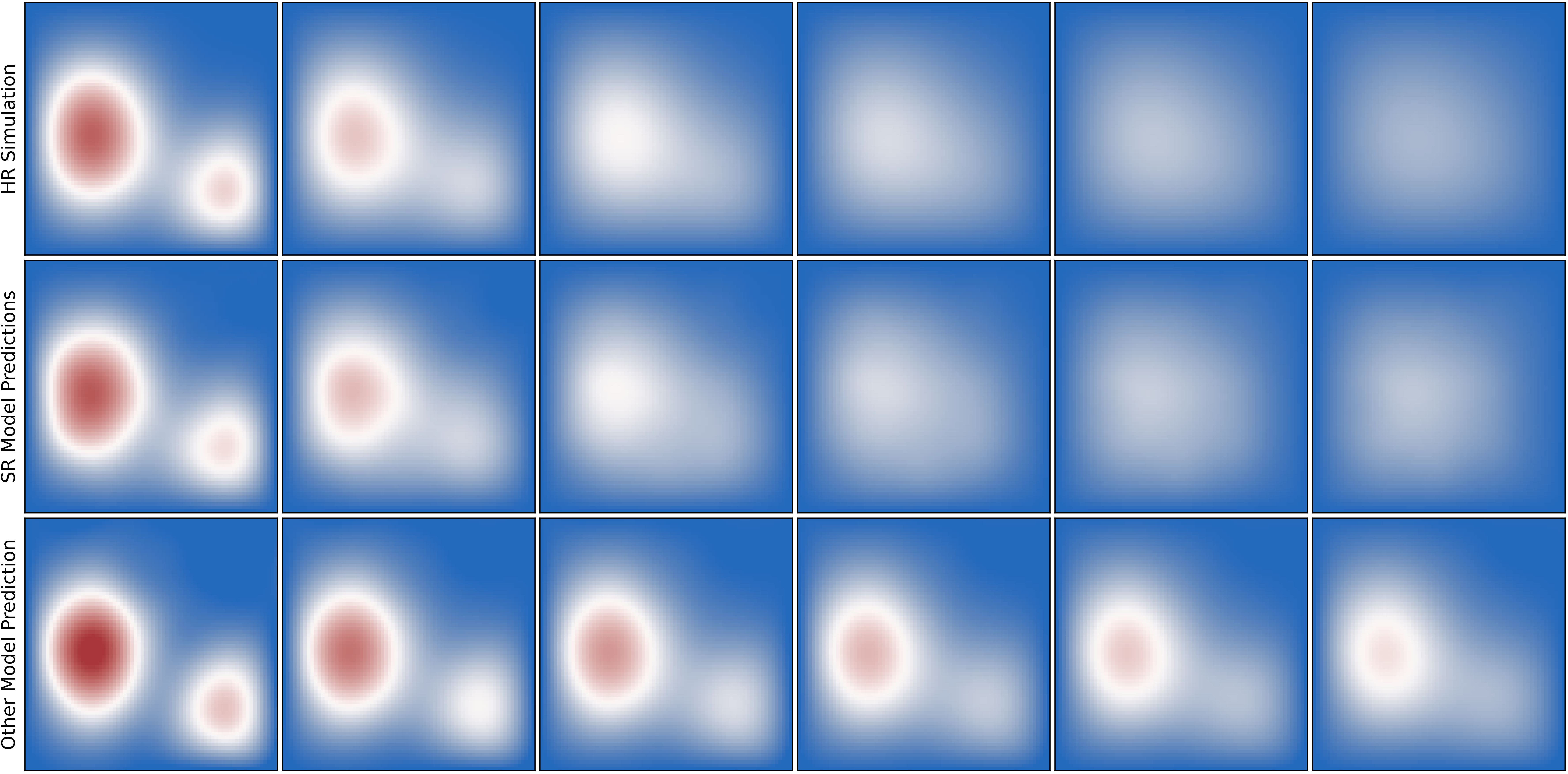} 
\vspace{-2mm}

\caption{Predictions of our approach (SR) and of its counterpart based on initial state only for 3 solutions to the 2D Diffusion equation with diffusion constants $D=0.5, \ 0.6,\ 0.8,$ outside the interval of values experienced in the training dataset. \label{fig: Diffusion 2D Speeds} }
\end{figure}

%% file: TexFiles/2DForcedDiffusion.tex
\begin{figure}[!h]

\centering

\includegraphics[width= 0.99\textwidth]{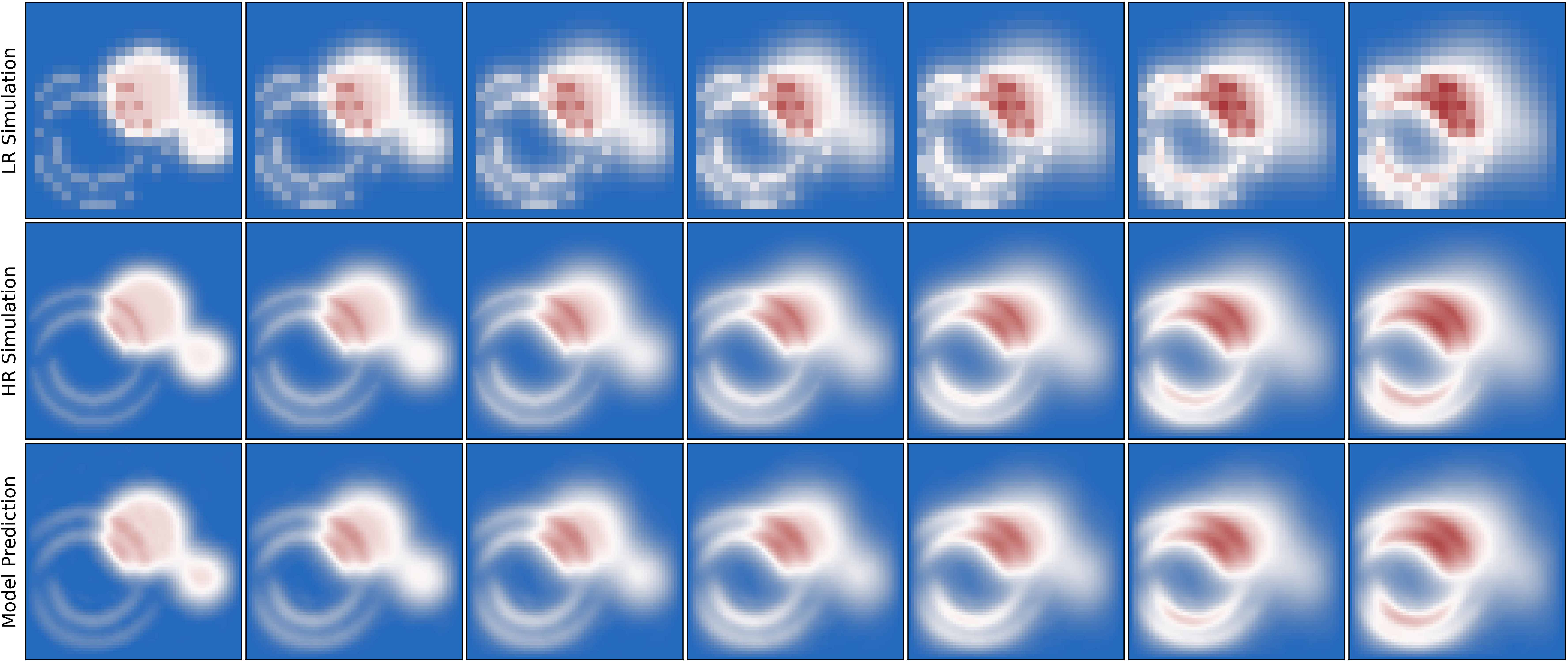} 
\vspace{0.6mm}

\includegraphics[width=0.99\textwidth]{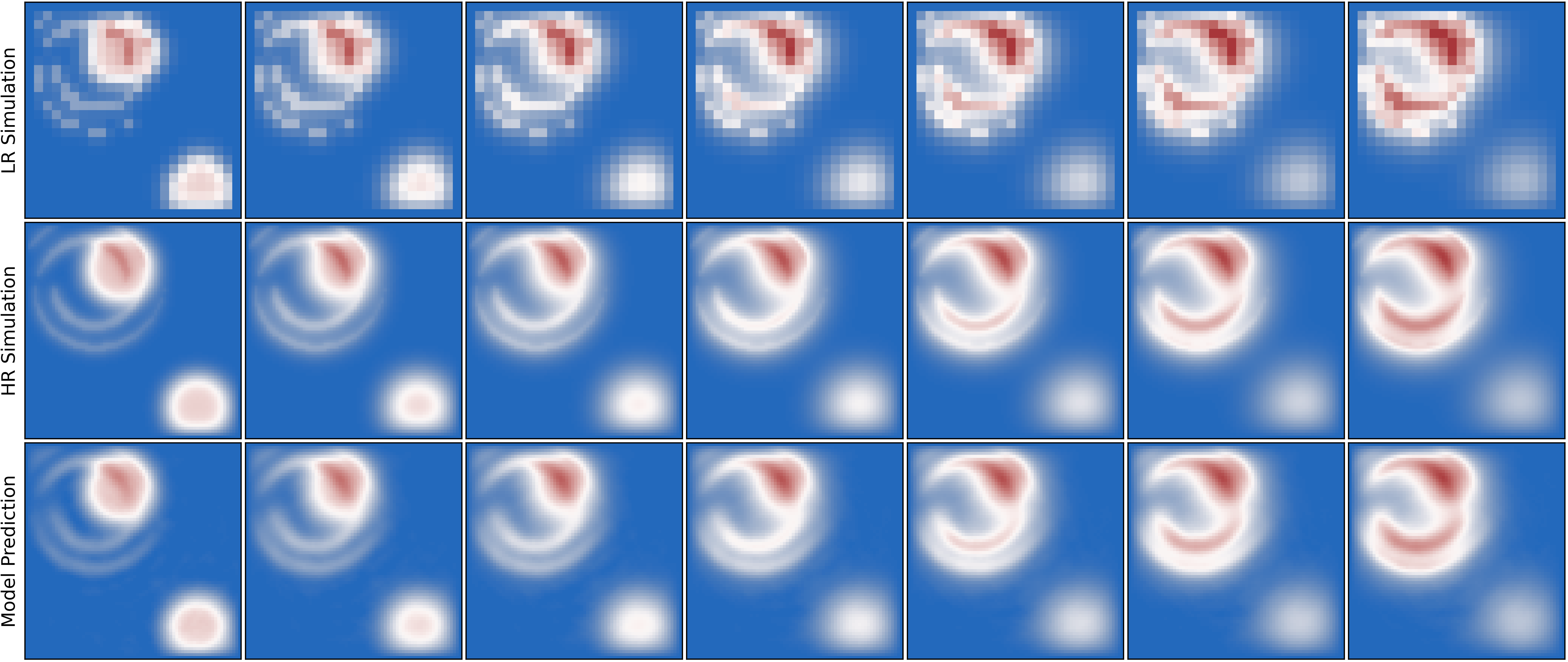} 
\vspace{0.6mm}

\includegraphics[width=0.99\textwidth]{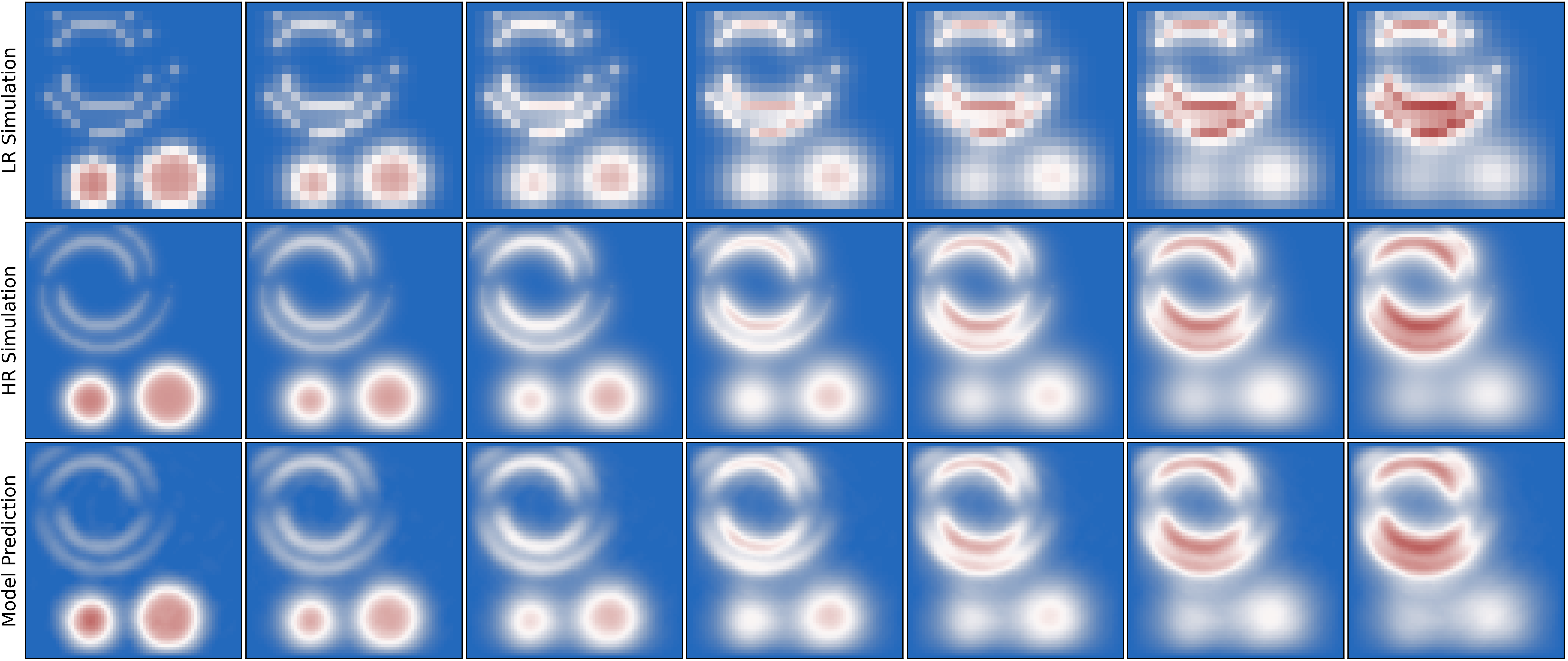} 
\vspace{0mm}

\caption{Super-resolution results for 3 previously-unobserved sequences from the dataset of solutions to the Forced 2D Diffusion equation~\eqref{eq: Forced 2D Diffusion}. \label{fig: Forced Diffusion 2D} }
\end{figure}

%% file: TexFiles/Kolmogorov1.tex
\begin{figure}[!h]
	\vspace{-0.3mm}
\centering
\includegraphics[width=0.995\textwidth]{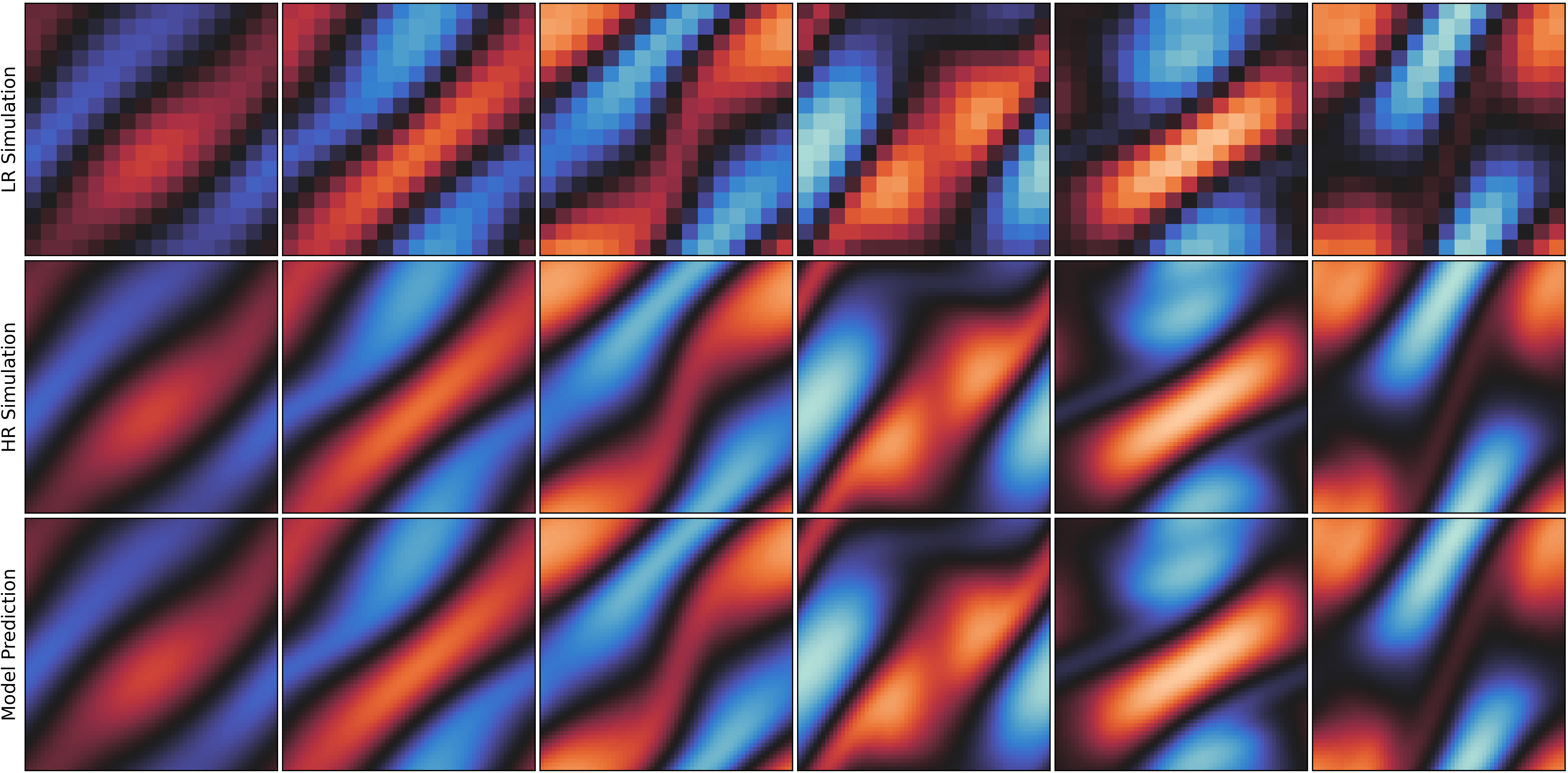} 
\vspace{-2.8mm}

\includegraphics[width=0.995\textwidth]{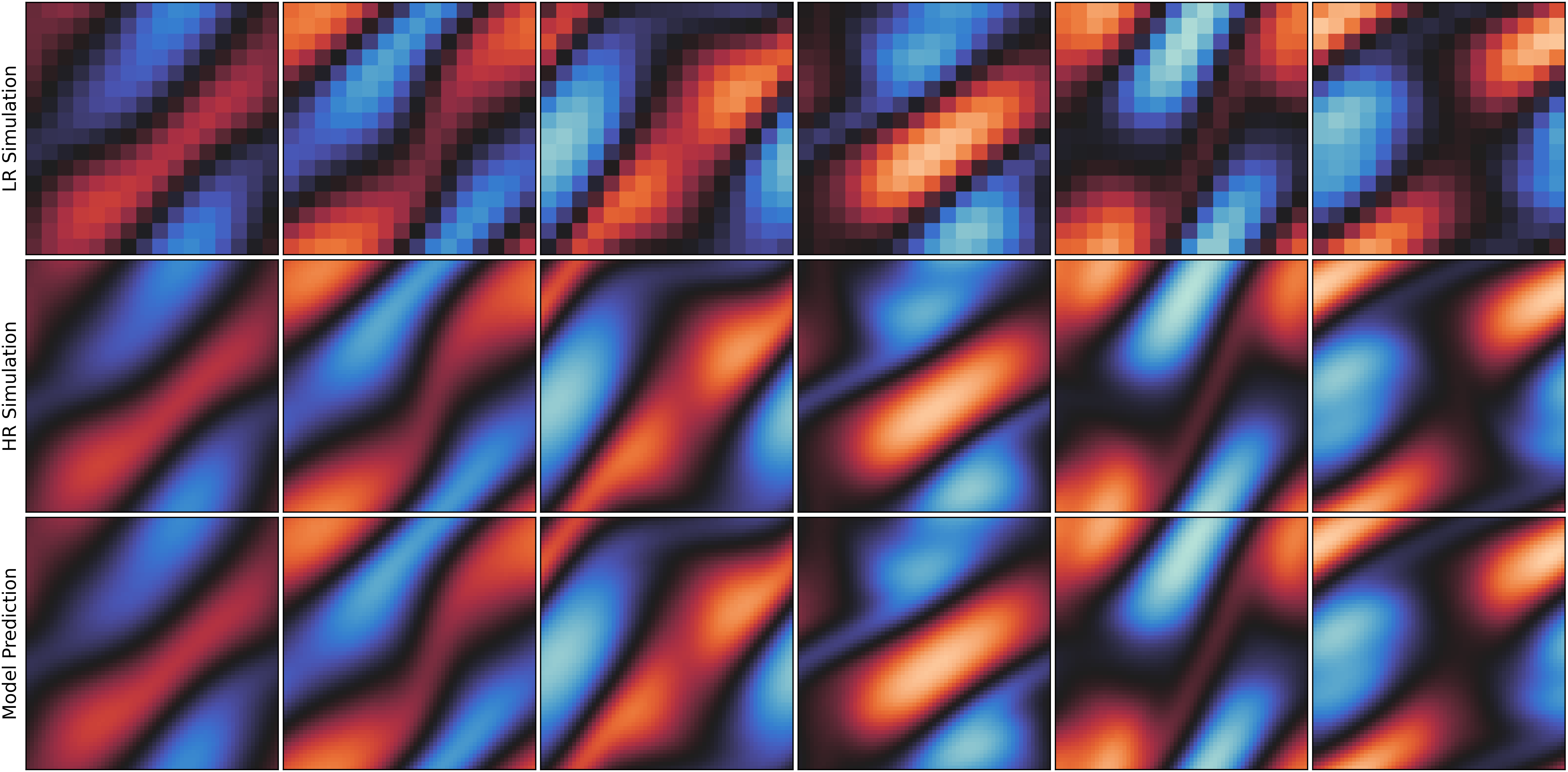} 
\vspace{-5.5mm}

\caption{Super-resolution results for 2 previously-unobserved sequences (in rows) from the 2D Kolmogorov flow dataset with fixed Reynolds number $Re = 20$. \label{fig: Kolmogorov Re20} }  \vspace{2mm}
\end{figure}

%% file: TexFiles/Kolmogorov2.tex
\begin{figure}[!h]
\centering
\includegraphics[width=0.999\textwidth]{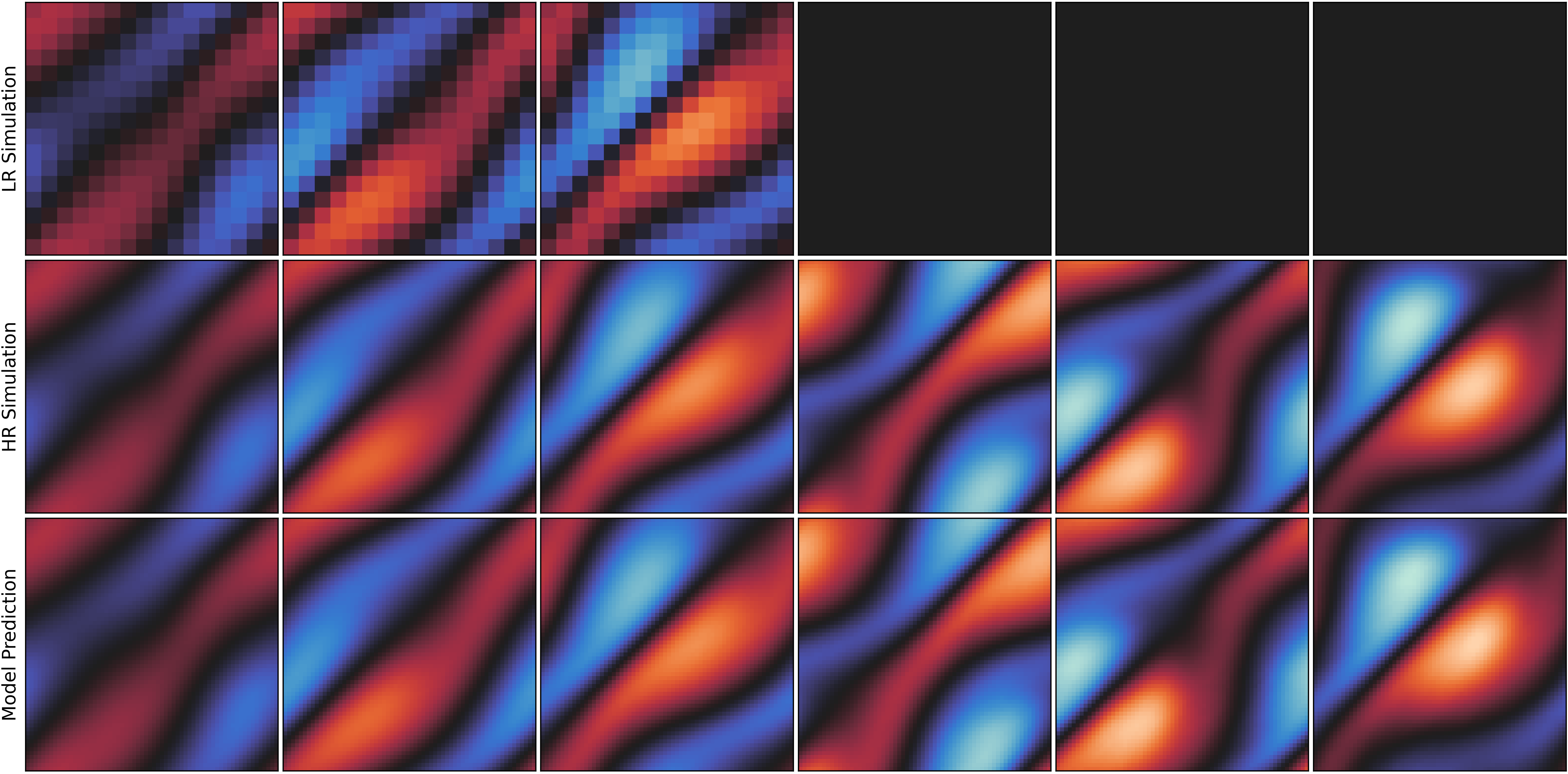} 
\vspace{-2.8mm}

\includegraphics[width=0.999\textwidth]{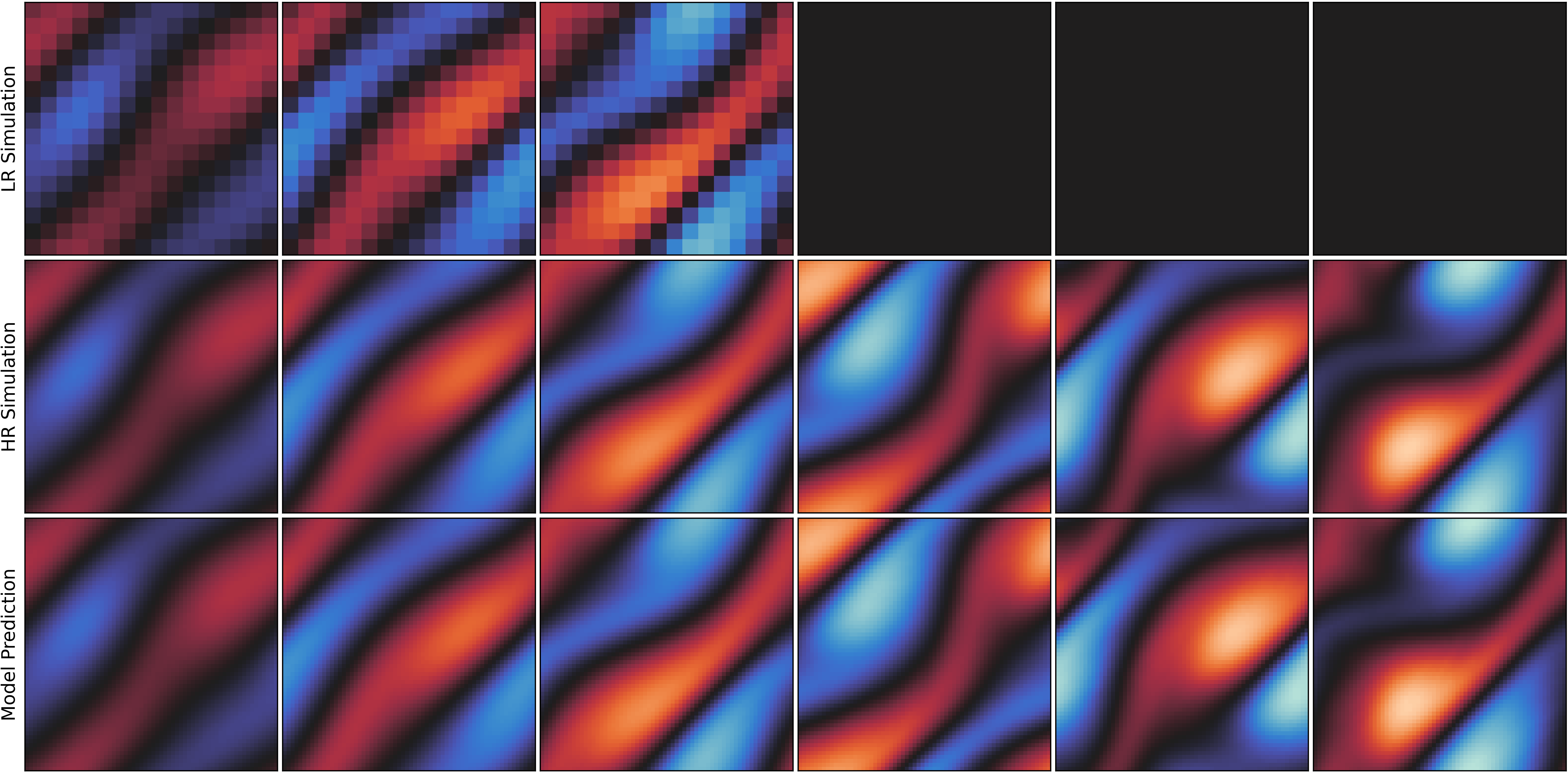} 
\vspace{-5.5mm}

\caption{Super-resolution results for 2 previously-unobserved sequences (in rows) from the 2D Kolmogorov flow dataset with fixed Reynolds number $Re = 20$, where only a partial low-resolution simulation is included as input of the SROpNet. \label{fig: Kolmogorov Re20_Partial} }  \vspace{2mm}
\end{figure}

%% file: TexFiles/Kolmogorov3.tex
\begin{figure}[!h]
	\vspace{-1.3mm}
	\centering
	\includegraphics[width=0.85\textwidth]{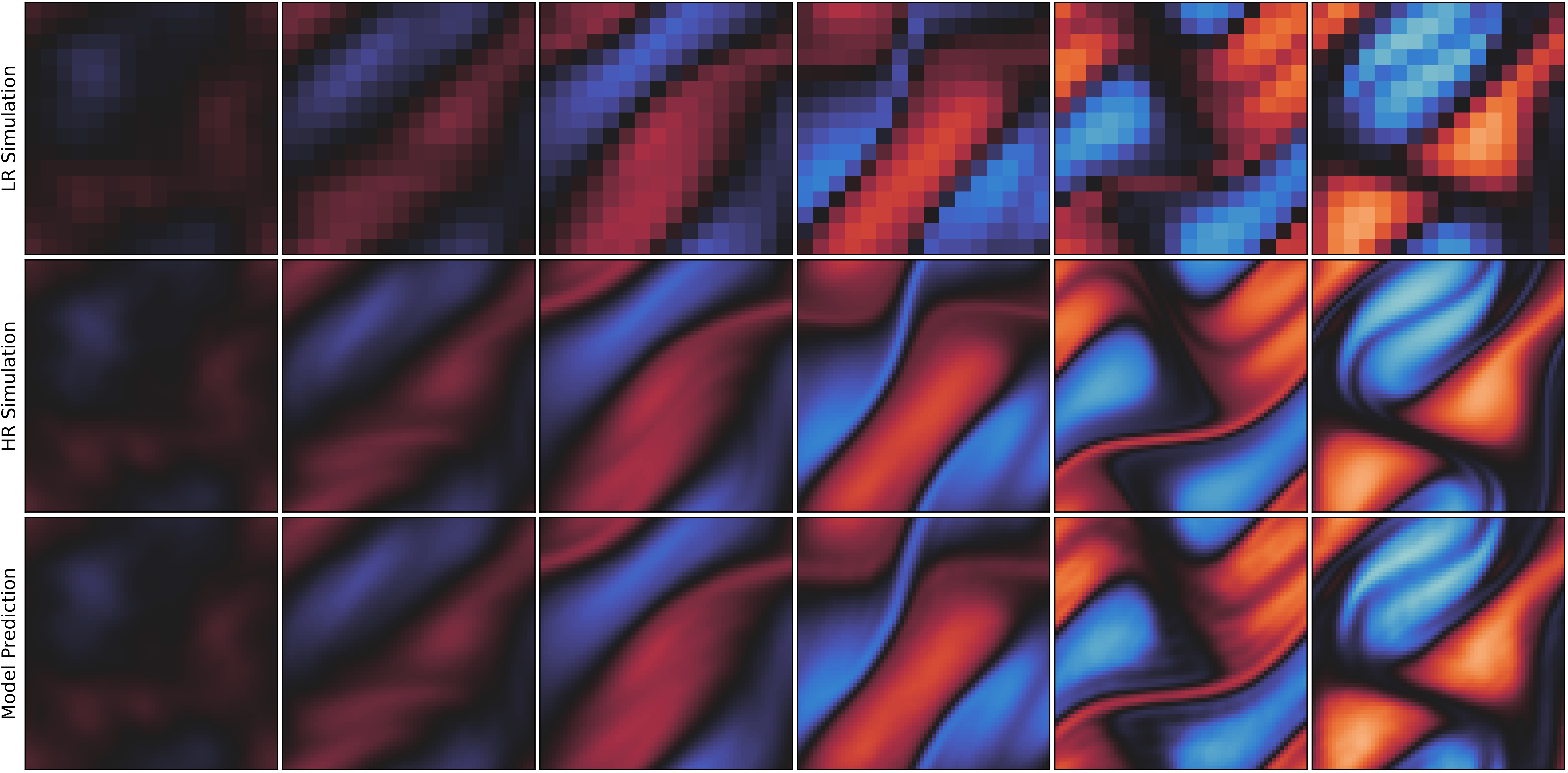} 
	\vspace{0.1mm}
	
	\includegraphics[width=0.85\textwidth]{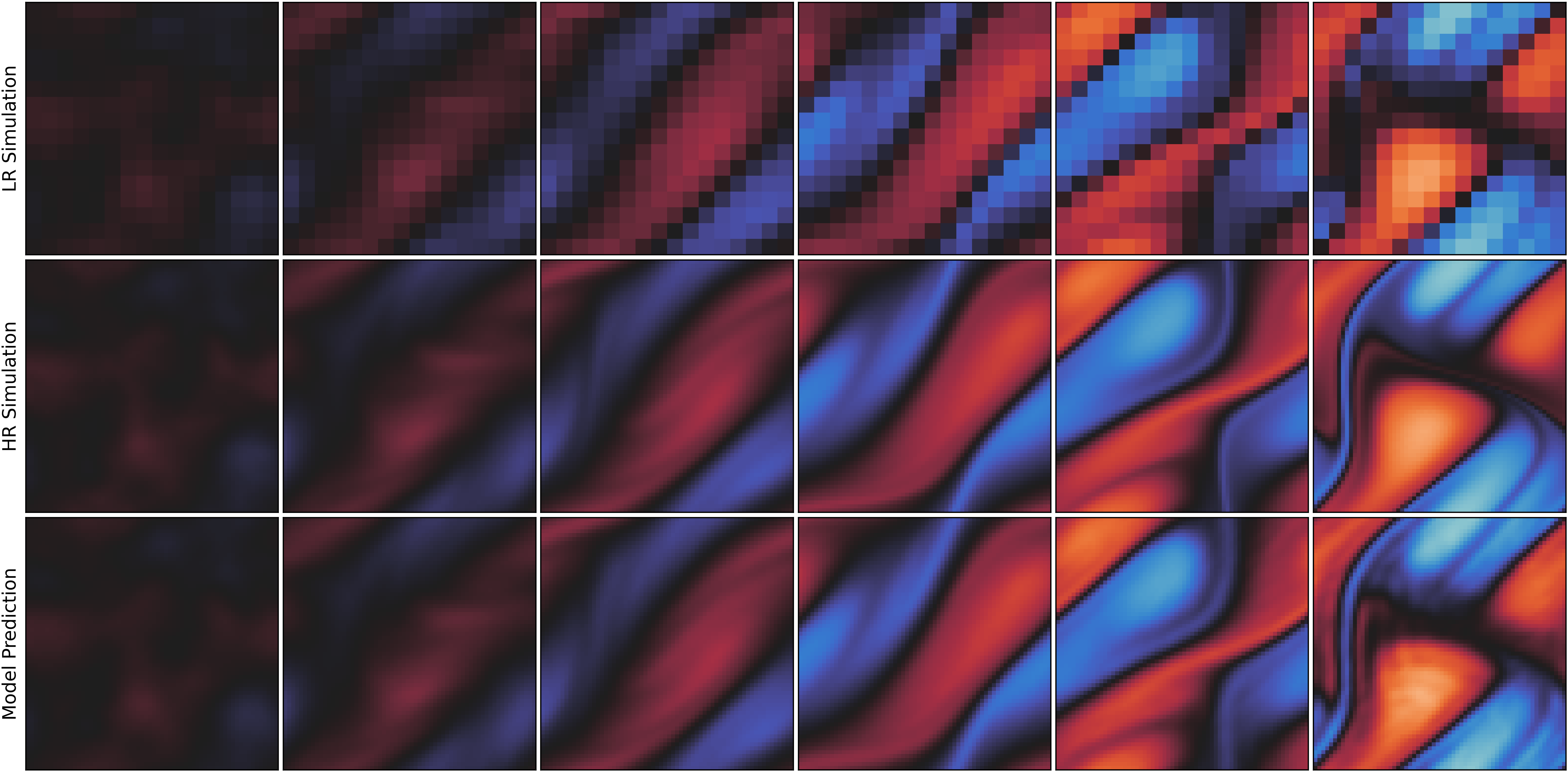} 
\vspace{0.1mm}
	
	\includegraphics[width=0.85\textwidth]{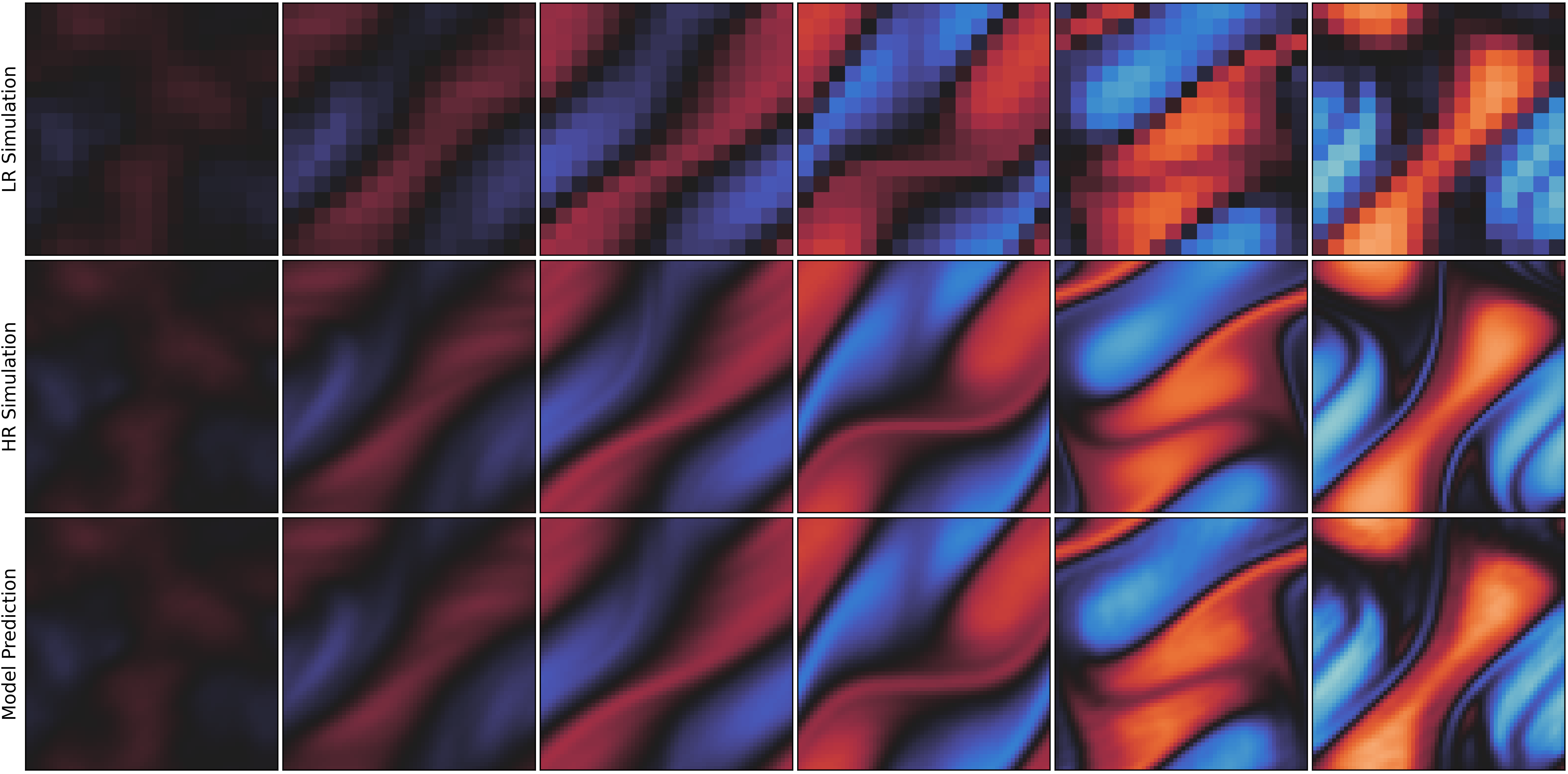} 
	\vspace{-2.4mm}
	
	\caption{Results for the 2D Kolmogorov flow dataset with mixed $Re \in [200,500]$. \label{fig: Kolmogorov MixedRe} }
\end{figure}